\pdfoutput=1
\documentclass[10pt,twocolumn,letterpaper]{article}

\usepackage[pagenumbers]{cvpr}

\usepackage[T1]{fontenc}
\usepackage{booktabs}
\usepackage{multirow}
\usepackage{array}
\usepackage{colortbl}
\usepackage{microtype}
\usepackage{enumitem}
\usepackage{amsmath}
\usepackage{amssymb}

\definecolor{cambrianred}{RGB}{249,235,232}
\definecolor{cambrianbg}{RGB}{242,246,252}
\definecolor{cambriansage}{RGB}{236,244,238}
\definecolor{cambriangray}{RGB}{240,240,242}
\definecolor{cambrianaccent}{RGB}{36,71,107}

\setlist[itemize]{leftmargin=1.15em, itemsep=0pt, topsep=1.5pt, parsep=0pt}
\setlist[enumerate]{leftmargin=1.3em, itemsep=0pt, topsep=1.5pt, parsep=0pt}
\usepackage{listings}
\usepackage{upquote}
\lstdefinestyle{vtaxprompt}{
    basicstyle=\ttfamily\tiny\linespread{0.92}\selectfont,
    breaklines=true,
    breakatwhitespace=true,
    breakindent=0pt,
    postbreak={},
    columns=fullflexible,
    keepspaces=true,
    upquote=true,
    aboveskip=1.5pt,
    belowskip=1.5pt,
    xleftmargin=2pt,
    xrightmargin=2pt,
    frame=single,
    framesep=1.5pt,
    framerule=0.25pt,
    rulecolor=\color{black!25},
}

\definecolor{linkblue}{rgb}{0.21,0.49,0.74}
\usepackage[pagebackref,breaklinks,colorlinks,allcolors=linkblue]{hyperref}
\hypersetup{pdftitle={Seeing Right, Writing Wrong: Where Short Chain-of-Thought Breaks Spatial VQA},bookmarksnumbered}

\newsavebox{\suppfitbox}
\newcommand{\fitto}[2]{\sbox{\suppfitbox}{#2}\ifdim\wd\suppfitbox>\dimexpr#1\relax\resizebox{#1}{!}{\usebox{\suppfitbox}}\else\usebox{\suppfitbox}\fi}
\extrafloats{100}
\renewcommand{\dbltopfraction}{0.82}
\renewcommand{\topfraction}{0.88}
\renewcommand{\bottomfraction}{0.45}
\renewcommand{\textfraction}{0.08}
\renewcommand{\dblfloatpagefraction}{0.78}
\renewcommand{\floatpagefraction}{0.78}
\title{Seeing Right, Writing Wrong:\\Where Short Chain-of-Thought Breaks Spatial VQA}

\author{Chenhao Zhang\\
Huazhong University of Science and Technology\\
{\tt\small ch\_zhang@hust.edu.cn}}
\hypersetup{pdfauthor={Chenhao Zhang}}

\begin{document}
\maketitle

\begin{abstract}
Asking a vision-language model (VLM) to think step by step often makes it worse at spatial questions, even though the image stays in context. We show that short chain-of-thought (CoT) turns a perception problem into a transcription problem. Once a VLM writes a spatial relation, its answer follows that sentence instead of the image, so whether CoT helps or hurts depends largely on whether the written relation is more accurate than answering directly. Studying $15$ open VLMs ($2$B--$38$B) and six frontier closed models, we rule out token count, turn format, and visual forgetting in open models and trace the failure through three stages. First, writing re-weights visual cues, weakening a ``lower is closer'' shortcut in all $13$ open models tested but also discarding valid cues, which produces false relations. Second, the written text takes over. With the model's own conclusion-removed rationale held fixed, swapping in an opposite-depth image moves the answer by $2$--$9$ points on rendered twins (vs.\ up to $77$ for direct answers) and by $3.0$--$3.4$ points on average on natural photographs (vs.\ $20.5$--$69.6$). Third, the concluding sentence can invert even a true relation, mainly on converse questions. Together, false relations and conclusion mis-mapping account for four in five CoT losses. Scale does not remove this bottleneck: \texttt{Qwen3-VL-32B} shows the same pattern. Frontier closed models are not hurt: they write relations as accurately as they answer, and three of six re-check the image after writing. The three queried with reasoning off lock in like open models, so re-grounding co-occurs with answer-turn reasoning; we identify answer-step re-grounding as the step open VLMs lack. In a single-model pilot, fine-tuning on fixed rationales with image-determined labels restores it on natural photographs. We open-sourced all our datasets and code at \href{https://ming-zch.github.io/seeing-right-writing-wrong}{https://ming-zch.github.io/seeing-right-writing-wrong}.
\end{abstract}

\section{Introduction}
\label{sec:intro}

\begin{figure*}[t]
\centering
\includegraphics[width=\textwidth]{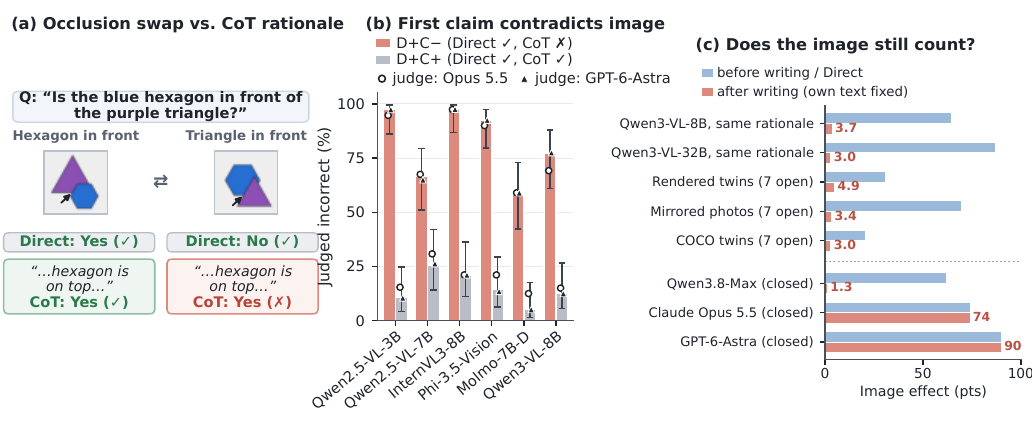}
\caption[Seeing right in direct readout while writing wrong spatial claims in short CoT]{\textbf{Seeing right, writing wrong.} \textbf{(a)} Occlusion twins for \texttt{Qwen3-VL-8B}, where \texttt{Direct} flips with the image while \texttt{CoT} calls the queried hexagon ``on top'' in both twins and answers Yes on both (paraphrased quotes; arrows mark the occlusion T-junction). \textbf{(b)} First two-object claims in \texttt{Synthetic Depth} rationales (all $500$ scenes; six models) judged false by two image-conditioned judges (consensus, Wilson CIs) on \texttt{D+C$-$} vs.\ \texttt{D+C$+$} items. \textbf{(c)} Twin image effect (in points, where $100$ means the answer follows the image) before vs.\ after a relation is written. Top: exploratory, selection-constrained comparison before vs.\ after the commit sentence $j^\ast$ within the same \texttt{D+C$-$} rationales (\cref{tab:supp_e57}). Middle: unselected \texttt{Direct} vs.\ the model's own conclusion-removed rationale $R_\mathrm{cut}$ held fixed (seven-model means; \cref{tab:profile,tab:supp_e60,tab:supp_e62}). Bottom: three closed models on the hard set (\texttt{Direct} at lowest reasoning vs.\ own conclusion-removed rationale held fixed; \cref{tab:supp_e61}).}
\label{fig:teaser}
\end{figure*}

Writing intermediate steps helps language models reason~\citep{wei2022cot}, yet on spatial questions about images, chain-of-thought (CoT) prompting and reasoning tuning often lower accuracy~\citep{kancheti2026cotspatial,yang2025thinking,liu2025mind}, and reinforcement-learning studies find that explicit thinking need not aid visual perception~\citep{li2025thinkornot,yu2025perceptionr1,jiang2025mmecot}. Even though the image remains in context throughout generation, three mechanisms could explain this drop: extra tokens or turn formatting may shift the answer distribution, generated prose may weaken access to the image~\citep{sun2025tvc,tian2026morethought,chen2026rememberr1}, or the semantic content of what the model writes may alter the decision.

We test these explanations on binary depth and spatial questions across $15$ open VLMs ($2$B--$38$B: seven core instruct models up to $8$B, three reasoning-tuned models, three $32$B--$38$B instruct models, and a \texttt{Qwen3-VL} $2$B/$4$B ladder) and six frontier closed models (\cref{sec:protocol}). Turn-closed filler of the rationale's length keeps all eight main-pool open models (the core seven and \texttt{Qwen3-VL-8B-Thinking}) within $2.3$ points of direct answering (\texttt{Direct}), and re-showing the image or question after the rationale recovers at most $0.8$ points each on the main pool (\cref{sec:behavioral}). Instead, the semantic content of the rationale drives the change. Much like verbal overshadowing in humans, where describing a face impairs its later recognition~\citep{schooler1990verbal,liu2025mind}, \emph{short CoT turns a perception problem into a transcription problem} (\cref{fig:teaser}). Once a VLM writes a spatial relation, it reads its final answer from that sentence rather than from the image, so CoT helps or hurts largely according to whether the written relation is more accurate than \texttt{Direct}, minus a conclusion-mapping cost. We trace this handoff with a \emph{forced readout}, the probability of the correct option at the prefix \texttt{``Answer: (''} appended after the question (\texttt{Direct}) or after any sentence of the rationale, and find three stages:
\begin{enumerate}
    \item \textbf{Stage 1: writing re-weights visual cues (\cref{sec:error_onset}).} Writing weakens a ``lower is closer'' height shortcut in all $13$ open models (\cref{fig:cues}a), which helps shortcut-reliant models on cue-balanced scenes, but it can also lose a valid cue that \texttt{Direct} reads. Of the $3{,}014$ items where \texttt{Direct} is right and \texttt{CoT} wrong (\texttt{D+C$-$}) on our three occlusion sets, $57\%$ end in a false written relation (\cref{fig:cues}b), typically an occlusion misreading that affirms the queried subject, even while a mid-layer probe at the relation word still decodes the true depth order.
    \item \textbf{Stage 2: the written text takes over (\cref{sec:error_onset,sec:answer_step}).} A coherent reasoner should follow its own text, so the informative test is whether the image still matters after a relation is written. In our primary unselected fixed-rationale swaps, holding each model's own conclusion-removed rationale ($R_\mathrm{cut}$) fixed reduces the twin image effect to $2.0$--$8.9$ points across the core seven models (vs.\ up to $76.5$ under \texttt{Direct}), lowers the mirror effect on natural photographs from $69.6$ to $3.4$, and lowers the occlusion effect on COCO object composites from $20.5$ to $3.0$. Conversely, editing the first written claim to the true relation beats \texttt{Direct} by $14.3$--$34.9$ points in five of six models. Along \texttt{D+C$-$} rationales, the forced readout stays right until a single commit sentence $j^\ast$, which usually states a spatial relation that contradicts the image ($88.0\%$ of audited \texttt{Synthetic Depth} cases). In an exploratory, selection-constrained prefix swap around that commit sentence, swapping in the twin moves the readout by $64.6$--$86.9$ points before it versus $3.0$--$3.7$ after it in \texttt{Qwen3-VL-8B} and \texttt{-32B} (an unselected \texttt{D+C$+$} control similarly falls from $47.1\to26.9$ and $70.8\to32.6$ after the first true relation, and to $7.8$ and $9.6$ two sentences later; \cref{tab:supp_e57}).
    \item \textbf{Stage 3: the conclusion can mis-map a true relation (\cref{sec:answer_step}).} Another $24\%$ of losses write a true relation and then answer the wrong option, mostly on converse questions ($29.5\%$ errors vs.\ $2.4\%$ on restatements).
\end{enumerate}
Together, a false relation or conclusion mis-mapping accounts for four in five \texttt{CoT} losses, where the model sees right under \texttt{Direct} but writes wrong in \texttt{CoT}. Separating written-relation accuracy from the answer step explains when CoT helps or hurts (\cref{sec:consequences}), and scaling model size alone does not remove this bottleneck (\texttt{Qwen3-VL-32B} exhibits the same pattern). Frontier closed models show that text lock-in is avoidable (\cref{sec:frontier}): when a swapped image contradicts their own correct written relation, \texttt{Claude Opus 5.5}, \texttt{GPT-6-Astra}, and \texttt{Gemini-4-Argon} follow the image (Claude and GPT reason at the answer turn, while Gemini reports no reasoning counts), whereas \texttt{Qwen3.8-Max}, \texttt{DeepSeek-V4.1-Flash}, and \texttt{HY4} (queried with reasoning off) lock onto their text like the open models. While label-free inference-time interventions do not beat \texttt{Direct} on average out of sample, a single-model pre-specified LoRA pilot indicates that answer-step re-grounding is learnable on natural photographs (\cref{sec:regrounding}). Our main contributions are:
\begin{itemize}[leftmargin=*,itemsep=1pt,topsep=2pt]
    \item \textbf{A controlled diagnosis} of why short CoT hurts spatial VQA across $15$ open and six closed VLMs, with controls in open models that rule out token count, turn format, and visual forgetting (\cref{sec:behavioral}).
    \item \textbf{A sentence-level account} of where rationales turn against the image, verified with forced readouts, claim edits, attention knockouts, and fixed-rationale image swaps on rendered scenes and natural photographs (\cref{sec:error_onset,sec:answer_step}).
    \item \textbf{An accounting decomposition} (written-relation accuracy versus the answer step) that explains when CoT helps or hurts across synthetic benchmarks and larger or reasoning-tuned open cohorts, predicts the out-of-sample \texttt{CoT}$-$\texttt{Direct} change on natural depth with a mean absolute error of $1.3$ points, and predicts the sign on held-out natural-object composites (\cref{sec:consequences}).
    \item \textbf{Evidence that lock-in is avoidable and re-grounding is learnable}, contrasting frontier closed models with open VLMs (\cref{sec:frontier}), evaluating label-free interventions, and showing in a single-model pre-specified LoRA pilot that fine-tuning on fixed rationales with image-determined labels restores answer-step re-grounding on natural photographs (\cref{sec:regrounding,tab:supp_e63_pilot}).
\end{itemize}

\section{Related Work}
\label{sec:related}

\paragraph{CoT and visual perception.}
Strong language skills do not ensure accurate perception~\citep{tong2024eyes,tong2024cambrian}: VLMs confuse spatial prepositions~\citep{kamath2023whatsup}, lose metric memory in video and 3D CoT~\citep{yang2025thinking,wang2026mindcube,yang2026cambrians}, and gain little or lose accuracy from thinking in RL and evaluation~\citep{yu2025perceptionr1,li2025thinkornot,jiang2025mmecot}. Across $13$ spatial benchmarks, CoT and reasoning tuning cut accuracy and raise language-prior reliance~\citep{kancheti2026cotspatial}. Beyond showing \emph{that} CoT hurts, we rule out format and visual forgetting, localize where a rationale turns against the image, and decompose the benchmark-level change into helpful shortcut dilution and harmful valid-cue loss, showing that \texttt{CoT}$-$\texttt{Direct} tracks written-relation accuracy minus \texttt{Direct} accuracy. CoT losses concentrate on tasks where verbalizing impairs humans~\citep{liu2025mind}; we locate this cost in the written relation. As benchmarks carry non-visual shortcuts~\citep{brown2025tst,geirhos2020shortcut}, we perturb cues in conflict as in depth psychophysics~\citep{landy1995cue} and shape-texture tests~\citep{geirhos2019texture}.

\paragraph{Faithfulness, snowballing, and grounding.}
Traces can rationalize answers~\citep{turpin2023unfaithful,arcuschin2025wild} or compute in filler~\citep{pfau2024dotbydot}, while VLM probes show a ``yes'' bias~\citep{li2023pope} and contrastive decoding counters priors via distorted images~\citep{leng2024vcd}. \citet{lanham2023measuring} test faithfulness through early answering and mistake injection. We refine early answering to identify the commit sentence $j^\ast$ where an initially correct forced readout turns wrong, and we edit perceptual claims whose ground truth is fixed by the rendered scene. Unlike sentence ranking by resampled rollouts~\citep{bogdan2025thoughtanchors}, post-answer rationalization in LLMs~\citep{zhang2024snowball}, or acceptance of earlier hallucinations~\citep{zhong2024mmhalsnowball}, we locate without resampling the sentence where a rationale turns against the image, before any answer token, and test image use after it via twin swaps. \citet{zhao2026robustness} inject misleading captions and wrong-think prefixes into RL-tuned VLMs; we show that unperturbed VLMs write their own false spatial claims and obey them.

\paragraph{Forgetting, probes, and explicit state.}
Against falling image attention in long chains, \texttt{TVC}~\citep{sun2025tvc} re-injects visual tokens and \texttt{VAPO}~\citep{tian2026morethought} and \texttt{Remember-R1}~\citep{chen2026rememberr1} reward grounding; in short CoT, re-showing the image does not recover accuracy. Mechanistic studies examine linear spatial IDs~\citep{kang2026spatialid}: \citet{fu2025hidden} show that VLMs encode visual features unused in direct answers, and \citet{wang2026knowing} ask whether spatial encodings reach the output. Complementing these findings, we show that a VLM can write a false spatial relation even while a probe at the relation token still decodes the true depth order, and we pinpoint where in the rationale visual information stops governing the answer. Our probe follows truthfulness probing~\citep{azaria2023internal,burns2023discovering,marks2024geometry}, where models encode the truth yet output an error~\citep{orgad2025llms}, and activation steering~\citep{li2023iti}. Explicit visual state (crops~\citep{wu2024vstar}, sketchpads~\citep{hu2024visual}, harnesses~\citep{han2026vista}) replaces prose; our structured state encodes the height shortcut rather than occlusion (\cref{sec:regrounding,tab:supp_state}).

\section{Experimental Setup}
\label{sec:protocol}

\begin{table*}[t]
\centering
\caption[Per-model profile across the core seven open instruct VLMs]{\textbf{Per-model profile across the core seven open instruct VLMs ($\le 8$B).} \textbf{Top:} \texttt{Direct} accuracy, lower-is-closer rate, and occlusion-reading status on the cue-factorial set (CIs in \cref{tab:supp_tab2_ci}). \textbf{Lower rows:} key metrics for Stage 1 false commit sentences on \texttt{Synthetic Depth} $431$ (two-judge consensus; \cref{sec:error_onset}), Stage 2 text lock-in (conclusion-removed rationale $R_\mathrm{cut}$; \cref{sec:error_onset,sec:answer_step}), and cue re-weighting plus net \texttt{CoT}$-$\texttt{Direct} accuracy changes (\cref{sec:consequences}; $^\ddagger$empty \texttt{BLINK} rationales; the identical \texttt{Qwen3-VL-8B} and \texttt{InternVL3.5-8B} depth entries are both $-15/324$). Symbols: -- not run; $^\ast$ $95\%$ CI excludes zero; red denotes primary estimators.}
\label{tab:profile}
\fontsize{7}{8.2}\selectfont
\setlength{\tabcolsep}{1.25pt}
\begin{tabular}{@{}lccccccc@{}}
\toprule
\textbf{Model} & \textbf{\shortstack{Qwen2.5-VL\\3B}} & \textbf{\shortstack{Qwen2.5-VL\\7B}} & \textbf{\shortstack{InternVL3\\8B}} & \textbf{\shortstack{Phi-3.5\\Vision}} & \textbf{\shortstack{Molmo\\7B-D}} & \textbf{\shortstack{Qwen3-VL\\8B}} & \textbf{\shortstack{InternVL3.5\\8B}} \\
\midrule
\rowcolor{cambriangray}\multicolumn{8}{@{}l@{}}{\textit{\textbf{What \texttt{Direct} relies on (cue-factorial set)}}} \\
Accuracy (\%) & $52.4$ & $53.4$ & $63.7$ & $59.5$ & $88.2$ & $78.9$ & $55.6$ \\
Follows lower-is-closer (\%) & $96.2$ & $95.6$ & $84.9$ & $62.6$ & $57.2$ & $62.2$ & $93.8$ \\
Reads occlusion & no & no & no & no & \textbf{yes} & \textbf{yes} & no \\
\midrule
\rowcolor{cambriangray}\multicolumn{8}{@{}l@{}}{\textit{\textbf{Stage 1: written relations are often false (commit sentences contradicting the image)}}} \\
\cellcolor{cambrianred}\texttt{Syn.\ Depth} 431: commit / matched (\%) & \cellcolor{cambrianred}$100$\,/\,$7$ & \cellcolor{cambrianred}$52$\,/\,$7$ & \cellcolor{cambrianred}$97$\,/\,$16$ & \cellcolor{cambrianred}$100$\,/\,$19$ & \cellcolor{cambrianred}$77$\,/\,$10$ & \cellcolor{cambrianred}$96$\,/\,$4$ & \cellcolor{cambrianred}$91$\,/\,$0$ \\
\midrule
\rowcolor{cambriangray}\multicolumn{8}{@{}l@{}}{\textit{\textbf{Stage 2: the answer follows the written text}}} \\
\bottomrule
\end{tabular}
\end{table*}

\begin{figure*}[p]
\centering
\includegraphics[width=0.98\textwidth]{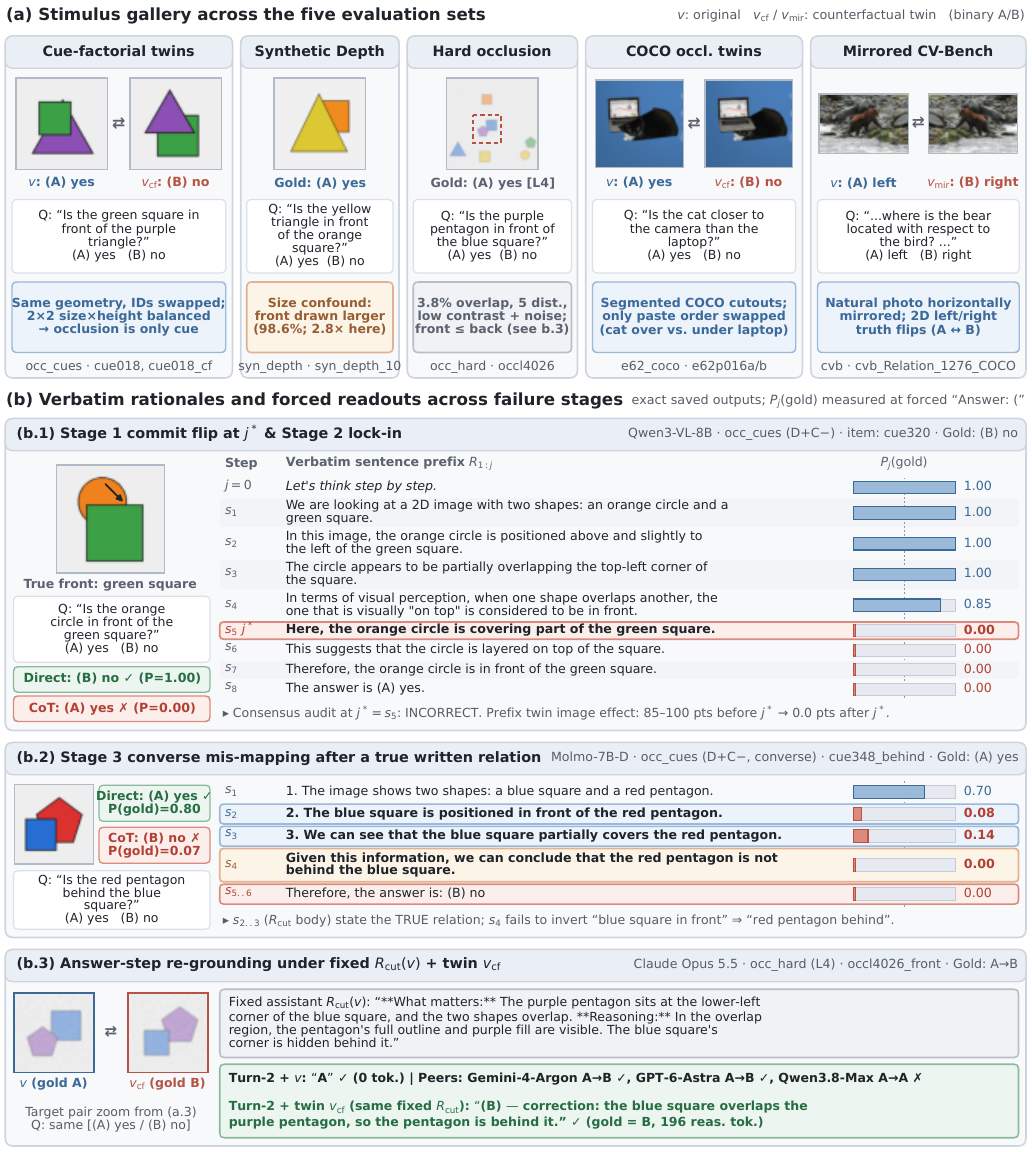}
\caption{\textbf{Stimulus gallery and verbatim rationales.} \textbf{(a)} Examples from five representative evaluation sets: cue-factorial twins (identical geometry, swapped identities, size/height balanced), size-confounded \texttt{Synthetic Depth}, closed-model hard occlusion, COCO occlusion twins (paste order swapped), and mirrored \texttt{CV-Bench} left/right photos. \textbf{(b)} Verbatim outputs with per-sentence forced readouts $P_j(\text{gold})$: \textbf{(b.1)} \texttt{Qwen3-VL-8B} Stage~1 false claim at commit sentence $j^\ast$ ($0.85\to0.00$) and Stage~2 text lock-in; \textbf{(b.2)} \texttt{Molmo-7B-D} Stage~3 conclusion mis-mapping after a true body relation; and \textbf{(b.3)} \texttt{Claude Opus 5.5} overriding fixed $R_\mathrm{cut}(v)$ on twin $v_\mathrm{cf}$, with $196$ hidden reasoning tokens on the twin call (peer outcomes show answer under $v\to$ under $v_\mathrm{cf}$). Panels (b.1)--(b.3) are single items; population estimates are in \cref{tab:supp_e57,tab:supp_e61}.}
\label{fig:supp_qual}
\end{figure*}

\paragraph{Model cohorts and forced readout.}
We evaluate $13$ open VLMs (plus \texttt{Qwen3-VL-2B/4B} on the twins) in \texttt{bfloat16} with greedy \texttt{vLLM}~\citep{kwon2023vllm} decoding across three cohorts: (i) the \emph{core seven instruct models} ($\le 8$B), \texttt{Qwen2.5-VL-3B}, \texttt{Qwen2.5-VL-7B}~\citep{bai2025qwen25vl}, \texttt{InternVL3-8B}~\citep{zhu2025internvl3}, \texttt{Phi-3.5-Vision}~\citep{abdin2024phi3}, \texttt{Molmo-7B-D}~\citep{deitke2024molmo}, \texttt{Qwen3-VL-8B}~\citep{bai2025qwen3vl}, and \texttt{InternVL3.5-8B}~\citep{wang2025internvl35}; (ii) \emph{three reasoning-tuned open models}, \texttt{Qwen3-VL-8B-Thinking}, \texttt{MiMo-VL-7B-RL}~\citep{xiaomi2025mimovl}, and \texttt{GLM-4.1V-9B-Thinking}~\citep{glmv2025glm41v}; and (iii) \emph{three $32$B--$38$B instruct models}, \texttt{Qwen3-VL-32B} (plus \texttt{Qwen3-VL-2B/4B} on the twins for a scale ladder), \texttt{Qwen2.5-VL-32B}, and \texttt{InternVL3.5-38B}. We also evaluate \emph{six frontier closed API models}: \texttt{Claude Opus 5.5}, \texttt{GPT-6-Astra}, \texttt{Gemini-4-Argon}, \texttt{Qwen3.8-Max}, \texttt{HY4}, and \texttt{DeepSeek-V4.1-Flash} (\cref{sec:frontier,sec:supp_closed}). Every prompt presents an image $v$, binary question $q$, and lettered options. We define the \emph{forced readout} as the higher-probability option letter at the appended assistant prefix \texttt{``Answer: (''}. \texttt{Direct} applies this forced readout immediately after the user turn; \texttt{CoT} first generates a greedy rationale $R$ after \texttt{``Let's think step by step.''} (at most $512$ tokens) and then applies the forced readout. We detail engine readouts and batch regeneration in \cref{sec:supp_impl,sec:supp_engine}.

\paragraph{Terms.}
\texttt{D+C$-$} denotes items that \texttt{Direct} answers correctly and \texttt{CoT} answers incorrectly (\texttt{D+C$+$} denotes both correct). The \emph{commit sentence} $j^\ast$ is the rationale sentence after which the forced readout's probability of the correct option falls below $0.5$ and stays there. The \emph{conclusion-removed rationale} $R_\mathrm{cut}$ strips the concluding answer sentence from $R$, and its last spatial relation is the \emph{body relation}. The \emph{twin image effect} is $P(\text{answer}=\text{gold}_v\mid v)-P(\text{answer}=\text{gold}_v\mid v_\mathrm{cf})$ in points, where the twin $v_\mathrm{cf}$ swaps which shape is in front ($100$ indicates that the answer follows the image, and $0$ that it ignores the image). Throughout, we evaluate the image effect on the discrete answer scale for rendered twins, mirrored photos, and closed models, and on the forced-readout probability scale for COCO occlusion twins and prefix trajectories. \emph{Pre-specified} refers to hypotheses and decision rules recorded in a timestamped internal file before running the experiment (not a public registry).

\paragraph{Stimuli.}
All items are binary; \cref{fig:supp_qual}a illustrates five representative evaluation sets, and \cref{fig:supp_qual}b shows verbatim rationales illustrating Stage~1 false claims, Stage~3 conclusion mis-mapping, and closed-model answer-step re-grounding. (1) \emph{Main pool} ($2{,}850$ items), comprising \texttt{CV-Bench}~\citep{tong2024cambrian} 2D Relation, 3D Depth, and 3D Distance, \texttt{Synthetic Relative Position}, and \texttt{Synthetic Depth} ($500$ two-shape occlusion scenes). In \texttt{Synthetic Depth}, the front shape is always drawn larger and keeps the larger visible area on $98.6\%$ of images, so \texttt{Direct}'s use of size and of occlusion cannot be separated. It therefore acts as a size-confounded stress test, and we analyze its $431$ visibly overlapping scenes. (2) \emph{Occlusion twins} ($800$ queries), consisting of $200$ scenes where the front shape is never drawn larger, paired with an identity-swapped twin of opposite depth order under ``in front of'' and ``behind''. (3) \emph{Cue-factorial set} ($1{,}600$ queries), a balanced crossing of drawn-area ratio, vertical height, twin identity swaps, and converse phrasings across $400$ base scenes, so that neither drawn size nor height predicts depth order and occlusion T-junctions provide the reliable cue (\cref{sec:supp_cues}). (4) \emph{Held-out natural pool} ($767$ items, including $324$ depth items), combining \texttt{BLINK}~\citep{fu2024blink} Spatial Relation and Relative Depth with five \texttt{3DSRBench}~\citep{ma2025threedsrbench} categories whose \texttt{Direct}-based predictions were frozen before the \texttt{CoT} runs (\cref{sec:consequences,sec:supp_heldout}). (5) A \emph{hard occlusion set} for closed models ($576$ prompts across $144$ base twin pairs; \cref{sec:frontier,sec:supp_closed}), together with mirrored \texttt{CV-Bench} left/right photos ($151$ items) and segmented COCO occlusion twins ($233$ pairs, $1{,}398$ prompts) for natural-image counterfactual swaps (\cref{sec:answer_step}).

\paragraph{Cue sensitivity and occlusion rule.}
Coding answers by whether the queried subject is placed in front removes yes/no polarity; we measure sensitivity via pooled $d'$ and pair-matched $d'_\mathrm{pair}$, the conditional-logit sensitivity across opposite-truth query or twin pairs that cancels per-image bias. On the cue-factorial set, a probit regression on occlusion, height, and size estimates each cue's probability-scale average marginal effect (AME, points) and probit weight, with cluster-bootstrap $95\%$ CIs. A model passes our occlusion-reading rule if its cue-set \texttt{Direct} accuracy CI lower bound exceeds $70\%$ and its occlusion weight CI lower bound exceeds $0$ (\cref{sec:supp_competence}). Two image-conditioned judges (\texttt{Claude Opus 5.5} and \texttt{GPT-6-Astra}) audit sentence truth against the image, and the same models act as text-only judges to extract which relation a rationale states (\cref{sec:supp_judge}).

\section{Analysis 1: Not Format or Forgetting}
\label{sec:behavioral}

\begin{table*}[t]
\centering
\caption[Behavioral controls on the binary spatial pool]{\textbf{Behavioral controls on the $2,850$-item binary spatial pool.} Accuracy (\%) and paired $\Delta$ (points); $^\ast$: $95\%$ image-cluster bootstrap CI excludes zero (CIs in \cref{tab:supp_tasks}). \texttt{Qwen3-VL-8B-Thinking} uses its native thinking trace $T$ (mean $307$ tokens) in the \texttt{CoT} rows and length-matched \texttt{Filler($|T|$)+turn}. \texttt{Molmo-7B-D} omits the multi-image run on this pool (--).}
\label{tab:behavioral_main}
\scriptsize
\setlength{\tabcolsep}{3.0pt}
\begin{tabular}{@{}lcccccccc@{}}
\toprule
\textbf{Condition} & \textbf{\shortstack{Qwen2.5-VL\\3B}} & \textbf{\shortstack{Qwen2.5-VL\\7B}} & \textbf{\shortstack{InternVL3\\8B}} & \textbf{\shortstack{Phi-3.5\\Vision}} & \textbf{\shortstack{Molmo\\7B-D}} & \textbf{\shortstack{Qwen3-VL\\8B}} & \textbf{\shortstack{Qwen3-VL\\8B-Think.}} & \textbf{\shortstack{InternVL3.5\\8B}} \\
\midrule
\texttt{Direct} & $80.3$ & $85.9$ & $89.6$ & $76.4$ & $74.5$ & $91.1$ & $92.8$ & $83.3$ \\
\rowcolor{cambrianred}\texttt{Short CoT} ($R$ / $T$) & $71.5$ & $82.1$ & $84.5$ & $67.2$ & $72.8$ & $87.6$ & $91.4$ & $85.8$ \\
\rowcolor{cambrianred}\quad $\Delta(\texttt{CoT}-\texttt{Direct})$ & $-8.8^\ast$ & $-3.9^\ast$ & $-5.1^\ast$ & $-9.2^\ast$ & $-1.7$ & $-3.5^\ast$ & $-1.4^\ast$ & $+2.5^\ast$ \\
\texttt{Filler($|R|$)} ($\Delta$ Dir.) & $77.8$ ($-2.5$) & $76.6$ ($-9.3$) & $87.0$ ($-2.7$) & $70.7$ ($-5.7$) & $72.6$ ($-1.9$) & $91.9$ ($+0.8$) & -- & $83.8$ ($+0.5$) \\
\rowcolor{cambriansage}\texttt{Filler($|R|$)+turn} ($\Delta$ Dir.) & $79.8$ ($-0.5$) & $86.6$ ($+0.7^\ast$) & $88.7$ ($-1.0^\ast$) & $74.1$ ($-2.3^\ast$) & $74.4$ ($-0.1$) & $91.6$ ($+0.5^\ast$) & $92.8$ ($0.0$) & $84.0$ ($+0.7^\ast$) \\
\quad $\Delta(\texttt{Fil+t}-\texttt{Fil})$ & $+2.0^\ast$ & $+10.0^\ast$ & $+1.7^\ast$ & $+3.4^\ast$ & $+1.9^\ast$ & $-0.3$ & -- & $+0.2$ \\
\texttt{CoT+turn} / \texttt{+restate} & $71.8$ / $72.3$ & $82.1$ / $82.6$ & $84.9$ / $84.9$ & $67.2$ / $68.0$ & $72.6$ / $73.3$ & $87.5$ / $87.6$ & $91.4$ / -- & $85.9$ / $86.2$ \\
\texttt{CoT+reimage} / \texttt{+both} & $72.0$ / $72.9$ & $82.2$ / $83.0$ & $84.5$ / $86.0$ & $68.0$ / $68.2$ & -- / -- & $87.6$ / $87.6$ & -- / -- & $85.9$ / $86.5$ \\
\bottomrule
\end{tabular}
\par\vspace{1pt}
\footnotesize{\raggedright --: condition not run. \texttt{+both}: \texttt{CoT+reimage+restate}. Task-level results: \cref{tab:supp_tasks}.\par}
\end{table*}

\paragraph{Format controls.}
Inserting neutral prose of rationale length $|R|$ in the same assistant turn before \texttt{``Answer: (''} costs up to $9.3$ points by shifting the option-letter prior (\cref{tab:behavioral_main}). Closing the assistant turn first (\texttt{Filler($|R|$)+turn}) removes this artifact and keeps all eight main-pool open models within $-2.3$ to $+0.7$ points of \texttt{Direct}. Re-showing the image or restating the question after $R$ recovers at most $0.8$ points each, and both together recover at most $1.5$ points. By contrast, \texttt{CoT} lowers main-pool accuracy significantly in six of the eight models ($-1.4$ to $-9.2$ points; \cref{tab:behavioral_main}); excluding the size-confounded \texttt{Synthetic Depth} stress test, three models lose and one gains (\cref{tab:supp_e58b}). The accuracy change therefore comes from what the rationale says, not from token length, turn structure, or lost access to the image.

\paragraph{What \texttt{Direct} relies on.}
The cue-factorial set isolates which visual cue each model's \texttt{Direct} readout uses (\cref{tab:profile}). Among the core seven instruct models, only \texttt{Molmo-7B-D} ($88.2\%$) and \texttt{Qwen3-VL-8B} ($78.9\%$) pass the occlusion-reading rule (scoring $55.2\%$ and $54.9\%$ on our harder occlusion set; \cref{sec:supp_competence}), whereas four models (\texttt{Qwen2.5-VL-3B/7B}, \texttt{InternVL3-8B}, and \texttt{InternVL3.5-8B}) follow ``lower is closer'' on $84.9\%$--$96.2\%$ of queries. Across the other open cohorts, \texttt{Qwen3-VL-32B} ($96.3\%$) and \texttt{Qwen3-VL-8B-Thinking} ($88.7\%$) strongly read occlusion, while \texttt{MiMo-VL-7B-RL} and \texttt{GLM-4.1V-9B-Thinking} pass at the rule's margin ($73.9\%$) and still name the lower shape on $74\%$--$76\%$ of queries. The remaining large instruct models, \texttt{InternVL3.5-38B} ($64.2\%$) and \texttt{Qwen2.5-VL-32B} ($61.7\%$), stay height-reliant (\cref{tab:cohorts_summary,tab:supp_e31,tab:supp_e43,tab:supp_e48}).

\section{Analysis 2: Written Relations Go Wrong}
\label{sec:error_onset}

\begin{figure}[t]
\centering
\includegraphics[width=0.9\columnwidth]{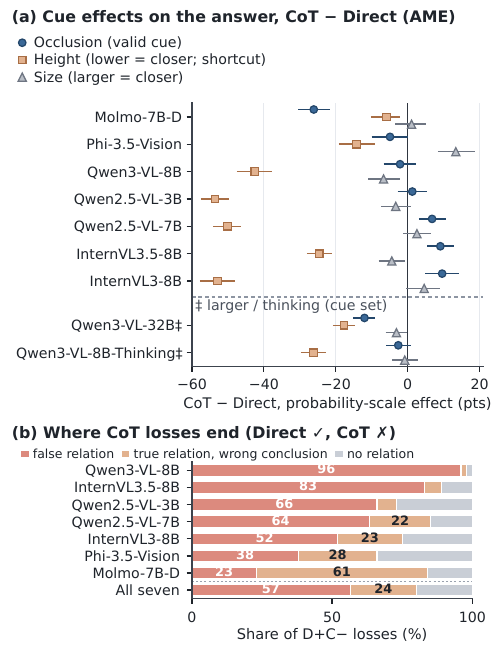}
\caption[Writing re-weights cues and ends most losses in a false statement]{\textbf{Writing re-weights cues and ends most losses in a false statement.} \textbf{(a)} \texttt{CoT}$-$\texttt{Direct} change in the average marginal effect (AME, points) of occlusion, vertical height (``lower is closer''), and size on the cue-factorial set ($95\%$ twin-pair bootstrap CIs; core seven models plus $^\ddagger$\texttt{Qwen3-VL-32B} and \texttt{Qwen3-VL-8B-Thinking}). Height falls in all nine plotted models (and all $13$ open models; \cref{tab:cohorts_summary,tab:supp_e31,tab:supp_e43,tab:supp_e48}). \textbf{(b)} Where the $3{,}014$ \texttt{D+C$-$} losses on the three occlusion sets end, by the last relation of the conclusion-removed rationale (two text-only judges; \cref{tab:supp_e54a}): a false relation (Stage 1), a true relation followed by a conclusion that answers the wrong option (Stage 3), or no relation.}
\label{fig:cues}
\end{figure}

\begin{figure*}[t]
\centering
\includegraphics[width=\textwidth]{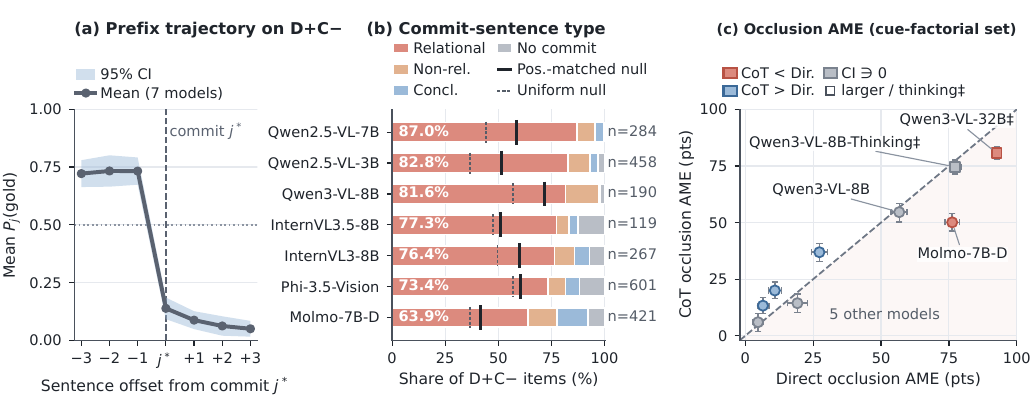}
\caption[Commit-sentence trajectories and cue-set occlusion effects]{\textbf{Commit-sentence trajectories and cue-set occlusion effects.} \textbf{(a)} Sentence-by-sentence forced-readout trajectory $P_j(\text{gold})$ on main-pool \texttt{D+C$-$} items aligned at the commit sentence $j^\ast$ (mean and $95\%$ band across the core seven models; $n=104$--$530$ items with a commit per model; original top-20 readout, so cell counts differ slightly from \cref{tab:behavioral_main}, \cref{sec:supp_engine}). The informative feature is the plateau near $0.73$ before $j^\ast$; the drop at $j^\ast$ holds by definition. \textbf{(b)} Commit-sentence types as shares of all \texttt{D+C$-$} items ($n$ at right; bar labels $63.9\%$--$87.0\%$ relational; among trajectories with a commit, $69.3\%$--$88.5\%$, \cref{tab:profile}). Solid black and dotted gray ticks mark the position-matched and uniform nulls. \textbf{(c)} Occlusion average marginal effect (AME, points) on the cue-factorial set under \texttt{Direct} vs.\ \texttt{CoT} ($95\%$ twin-pair bootstrap CIs; squares$^\ddagger$: \texttt{Qwen3-VL-32B} and \texttt{Qwen3-VL-8B-Thinking}). It falls significantly in the models that read occlusion and lose accuracy (\texttt{Molmo-7B-D} $-26.1$, \texttt{Qwen3-VL-32B} $-12.0$) and rises in three height-reliant models.}
\label{fig:error_onset}
\end{figure*}

\paragraph{Writing re-weights image cues.}
On the cue-factorial set, verbalization weakens the lower-is-closer height shortcut across all $13$ open models tested, reducing the height AME by up to $53.6$ points across the core seven models (\cref{fig:cues}a, \cref{tab:profile}) and similarly across the larger and reasoning-tuned cohorts (\cref{tab:cohorts_summary,tab:supp_e31,tab:supp_e43,tab:supp_e48}). Because this height prior is misleading when cues are balanced, weakening it helps height-reliant models gain accuracy and raises occlusion AME in three of them (\cref{fig:error_onset}c). Conversely, where \texttt{Direct} already reads a valid cue, \texttt{CoT} often degrades the final answer. The final-answer occlusion AME falls significantly in \texttt{Qwen3-VL-32B} ($-12.0$ points) and \texttt{Molmo-7B-D} ($-26.1$ points; \cref{fig:error_onset}c), and on the size-confounded \texttt{Synthetic Depth} stress test, \texttt{CoT} lowers accuracy by $8.4$--$25.8$ points in five of the core seven models (\cref{tab:profile}).

\begin{table*}[t]
\centering
\scriptsize
\setlength{\tabcolsep}{1.5pt}
\caption{\textbf{Cue effects of the written relation} (cue-factorial set). Average marginal effect (AME, points) of the height (lower is closer) and occlusion cues from the probit of \cref{tab:supp_cue_ame}, fitted to the \texttt{Direct} answer (D), to the last relation written in the conclusion-removed rationale $R_\mathrm{cut}$ (W; two text-only judges, consensus) and to the \texttt{CoT} answer (C), all on the items with a written relation (Cov.). Here W$-$D is the change made by writing and C$-$W is the change at the answer step. Twin-pair bootstrap $95\%$ CIs ($1{,}000$ draws, same draws and seed as \cref{tab:supp_cue_ame}). $^\dagger$No $R_\mathrm{cut}$ judge labels: rule parser on $R_\mathrm{cut}$; it reproduces the judges' height changes but not their occlusion changes (\eg, \texttt{Molmo-7B-D} W$-$D occlusion $-6.0$ vs.\ $+11.1$), so only the height columns are interpretable for these rows.}
\label{tab:supp_written_ame}
\fitto{\textwidth}{\begin{tabular}{lr rrrcc rrrcc}
\toprule
 & & \multicolumn{5}{c}{Height (lower is closer)} & \multicolumn{5}{c}{Occlusion} \\
\cmidrule(lr){3-7}\cmidrule(lr){8-12}
Model & Cov. & D & W & C & W$-$D & C$-$W & D & W & C & W$-$D & C$-$W \\
\midrule
Qwen2.5-VL-3B & 80 & 92.4 & 18.8 & 43.5 & $-$73.5 {\tiny[$-$78.6, $-$68.1]} & +24.7 {\tiny[+20.7, +29.0]} & 4.3 & 12.2 & 6.3 & +7.9 {\tiny[+2.9, +12.9]} & $-$5.9 {\tiny[$-$9.9, $-$1.8]} \\
Qwen2.5-VL-7B & 90 & 91.6 & 44.2 & 44.2 & $-$47.4 {\tiny[$-$51.8, $-$43.1]} & +0.0 {\tiny[$-$3.3, +3.4]} & 6.3 & 11.9 & 13.8 & +5.6 {\tiny[+1.5, +10.0]} & +1.9 {\tiny[$-$1.5, +5.3]} \\
InternVL3-8B & 83 & 69.2 & $-$4.6 & 16.1 & $-$73.8 {\tiny[$-$79.5, $-$68.3]} & +20.7 {\tiny[+16.7, +25.1]} & 28.0 & 53.8 & 41.2 & +25.8 {\tiny[+20.5, +31.3]} & $-$12.6 {\tiny[$-$16.7, $-$8.6]} \\
Phi-3.5-Vision & 77 & 22.2 & $-$8.0 & 3.6 & $-$30.2 {\tiny[$-$35.7, $-$25.0]} & +11.6 {\tiny[+7.5, +15.3]} & 19.4 & 23.9 & 16.6 & +4.4 {\tiny[$-$1.2, +9.9]} & $-$7.3 {\tiny[$-$11.2, $-$3.5]} \\
Molmo-7B-D & 88 & 14.6 & 0.5 & 9.1 & $-$14.1 {\tiny[$-$17.8, $-$10.8]} & +8.5 {\tiny[+4.5, +12.8]} & 75.5 & 86.6 & 51.7 & +11.1 {\tiny[+7.4, +14.7]} & $-$34.9 {\tiny[$-$39.2, $-$30.8]} \\
Qwen3-VL-8B & 99 & 26.0 & $-$19.4 & $-$16.2 & $-$45.4 {\tiny[$-$50.1, $-$41.0]} & +3.2 {\tiny[+1.8, +4.8]} & 56.5 & 59.2 & 55.2 & +2.7 {\tiny[$-$1.6, +7.1]} & $-$4.0 {\tiny[$-$5.5, $-$2.5]} \\
InternVL3.5-8B & 88 & 87.1 & 40.1 & 60.2 & $-$47.0 {\tiny[$-$51.7, $-$42.1]} & +20.1 {\tiny[+16.9, +23.5]} & 11.1 & 30.7 & 22.4 & +19.6 {\tiny[+14.7, +24.4]} & $-$8.3 {\tiny[$-$11.7, $-$4.9]} \\
Qwen3-VL-32B & 99 & 5.0 & $-$12.4 & $-$12.2 & $-$17.5 {\tiny[$-$20.5, $-$14.3]} & +0.2 {\tiny[+0.0, +0.7]} & 92.7 & 81.2 & 81.1 & $-$11.5 {\tiny[$-$14.6, $-$8.5]} & $-$0.1 {\tiny[$-$0.4, +0.0]} \\
Qwen3-VL-8B-Think. & 83 & 20.1 & $-$8.5 & $-$7.5 & $-$28.5 {\tiny[$-$32.1, $-$24.4]} & +1.0 {\tiny[$-$0.4, +2.5]} & 76.7 & 73.9 & 72.3 & $-$2.8 {\tiny[$-$6.7, +0.9]} & $-$1.6 {\tiny[$-$2.7, $-$0.4]} \\
\midrule
InternVL3.5-38B$^\dagger$ & 93 & 71.3 & 11.7 & 16.1 & $-$59.6 {\tiny[$-$64.3, $-$55.1]} & +4.4 {\tiny[+1.7, +7.2]} & 28.0 & 47.6 & 52.5 & +19.6 {\tiny[+14.8, +24.5]} & +4.9 {\tiny[+2.3, +7.7]} \\
Qwen2.5-VL-32B$^\dagger$ & 86 & 43.4 & $-$5.5 & 15.4 & $-$48.8 {\tiny[$-$54.2, $-$43.3]} & +20.9 {\tiny[+17.1, +25.0]} & 22.1 & 14.9 & 11.9 & $-$7.2 {\tiny[$-$12.3, $-$1.8]} & $-$3.0 {\tiny[$-$7.3, +1.0]} \\
MiMo-VL-7B-RL$^\dagger$ & 97 & 47.8 & 20.4 & 32.0 & $-$27.4 {\tiny[$-$31.7, $-$23.4]} & +11.5 {\tiny[+8.8, +14.2]} & 48.7 & 50.0 & 55.1 & +1.3 {\tiny[$-$2.8, +5.3]} & +5.0 {\tiny[+2.5, +7.7]} \\
GLM-4.1V-9B-Think.$^\dagger$ & 84 & 52.3 & 23.4 & 35.1 & $-$28.9 {\tiny[$-$33.4, $-$24.3]} & +11.7 {\tiny[+8.0, +15.2]} & 48.2 & 56.6 & 62.8 & +8.3 {\tiny[+4.0, +12.9]} & +6.3 {\tiny[+2.7, +9.8]} \\
\bottomrule
\end{tabular}}
\end{table*}

Refitting the cue-factorial probit directly to the last relation written before the conclusion ($W$, two-judge consensus; \cref{tab:supp_written_ame}) separates cue shifts during writing ($W-D$) from changes at the answer step ($C-W$). Height reliance drops already in $W$ across all nine judged models ($W-D$ from $-73.8$ to $-14.1$ points, every CI below $0$) and in the four parser-only models, whereas the answer step partly restores it ($C-W$ up to $+24.7$). For occlusion, the two models that lose accuracy on the cue set diverge in stage. \texttt{Qwen3-VL-32B} loses occlusion during writing ($W-D = -11.5$ $[-14.6, -8.5]$, $C-W = -0.1$), whereas \texttt{Molmo-7B-D}'s body relation uses occlusion more than \texttt{Direct} ($W-D = +11.1$ $[+7.4, +14.7]$) before losing it at the answer step ($C-W = -34.9$ $[-39.2, -30.8]$). Similarly, the occlusion gains of \texttt{InternVL3-8B} ($+25.8$) and \texttt{InternVL3.5-8B} ($+19.6$) are larger in $W$ than in the final answer $C$. A likelihood-ratio test also rejects uniform attenuation of all cue weights in all nine judged models ($p\le0.0013$; \cref{tab:supp_e54b}), confirming that writing discounts vertical position beyond any common scaling.

\paragraph{Commit-sentence onset and visual contradiction audit.}
To pinpoint where a rationale turns against a correct direct readout, we evaluate the forced readout $P_j(\text{gold})$ after each sentence prefix $s_{1:j}$ of a rationale $R=(s_1,\dots,s_M)$. We define the \emph{commit sentence} $j^\ast$ as the sentence where $P_j(\text{gold})$ drops below $0.5$ and remains below $0.5$ for all subsequent sentences. On main-pool \texttt{D+C$-$} items, $P_j(\text{gold})$ stays high prior to $j^\ast$ (mean $0.73$ two to three sentences before the drop; \cref{fig:error_onset}a). While the drop at $j^\ast$ holds by construction, the content of $j^\ast$ is specific: across the core seven models, $69.3\%$--$88.5\%$ of trajectories with a commit sentence flip at a sentence stating a spatial relation, $9.9$--$32.2$ points above a position-matched null (\cref{fig:error_onset}b). Two image-conditioned judges (\texttt{Claude Opus 5.5}, \texttt{GPT-6-Astra}; agreeing with a human check by one annotator, the author, on $23/24$ judgeable sentences, $\kappa=0.91$) mark $88.0\%$ of judgeable commit sentences on \texttt{Synthetic Depth} $431$ as contradicting the image, compared with $9.4\%$ of position-matched relational sentences from \texttt{D+C$+$} rationales (\cref{tab:profile}; first claims show the same gap in \cref{fig:teaser}b). On the cue-factorial set, commit sentences contradict the image in $91.1\%$ of audited \texttt{Qwen3-VL-8B} cases (vs.\ $8.4\%$ of matched \texttt{D+C$+$} comparators) and in $98.4\%$ (vs.\ $1.9\%$) of \texttt{Qwen3-VL-32B} cases (\cref{tab:supp_e57}). By contrast, commit sentences contradict the image in $37.4\%$ of \texttt{Molmo-7B-D} cases, where errors arise later at the conclusion step (\cref{sec:supp_a2}). Running the full pre-specified sentence-level pipeline on our strongest open occlusion reader, \texttt{Qwen3-VL-32B} (\texttt{Direct} $96.3\%$ on the cue set; \cref{tab:supp_e57}), confirms that model scale does not remove this bottleneck. Every \texttt{D+C$-$} trajectory commits at a single sentence ($150/150$ on the cue set, $86.0\%$ relational, $+8.2$ points over the position-matched null, $p=0.0042$), and all $145/145$ relation-bearing cue-set losses end in a false relation citing the overlap ($93.8\%$ affirming the queried subject). Earlier false sentences can precede $j^\ast$ without flipping the readout, so $j^\ast$ marks where the false relation begins to drive the answer rather than always the first false sentence (\cref{sec:supp_a2,tab:supp_jstar_condition}).

\begin{table*}[t]
\centering
\caption{\textbf{Decomposition of \texttt{CoT} losses (\texttt{D+C$-$}) and rescues (\texttt{D$-$C+})} by the last written relation of the conclusion-removed rationale (two-judge consensus), pooled over the cue set, twins, and \texttt{Synthetic Depth} $431$ (per-set rows in the released outputs). (a) False relation, split by stated cue (affirmation without a visual reason, aff.; vertical position, height; size; misread overlap, occl.; other); (b) true relation with a wrong answer, split into mapping (the question is not the written proposition in its own words), reversal, and other; (c) no relation. Shares are in \% of \texttt{D+C$-$} with $95\%$ CIs from a twin-scene bootstrap ($400$ scenes of $4$ prompts on the cue set, $200$ scenes on the twins, and image clusters on \texttt{Synthetic Depth}); the pooled row resamples each scene jointly for all seven models (pooled integer shares $57+24+20=101\%$ from rounding $56.6+23.8+19.6$; in $696/717$ of (b) items the wrong answer is already stated in the removed conclusion). Mirror columns show \texttt{D$-$C+} with a true relation (a'), a false relation but a right answer (b'), and no relation (c').}
\label{tab:supp_e54a}
\scriptsize
\setlength{\tabcolsep}{2pt}
\fitto{\textwidth}{\begin{tabular}{lr rrrrrr rrrr r rrrr}
\toprule
 & & \multicolumn{6}{c}{(a) false relation} & \multicolumn{4}{c}{(b) true relation, wrong answer} & (c) & \multicolumn{4}{c}{D$-$C+ mirror} \\
\cmidrule(lr){3-8}\cmidrule(lr){9-12}\cmidrule(lr){14-17}
Model (pooled over sets) & $n$ & all & aff. & height & size & occl. & other & all & map. & rev. & other & none & $n$ & (a') & (b') & (c') \\
\midrule
Qwen2.5-VL-3B & 470 & 66 [61, 70] & 20 & 10 & 1 & 13 & 21 & 7 [5, 9] & 5 & 2 & 0 & 27 & 376 & 67 & 7 & 26 \\
Qwen2.5-VL-7B & 365 & 64 [59, 69] & 1 & 17 & 0 & 41 & 5 & 22 [18, 26] & 19 & 3 & 0 & 15 & 417 & 64 & 24 & 12 \\
InternVL3-8B & 449 & 52 [48, 56] & 1 & 2 & 0 & 47 & 2 & 23 [20, 27] & 18 & 5 & 0 & 25 & 453 & 81 & 4 & 15 \\
Phi-3.5-V & 539 & 38 [34, 42] & 9 & 6 & 1 & 12 & 10 & 28 [25, 32] & 25 & 3 & 0 & 34 & 446 & 55 & 13 & 32 \\
Molmo-7B-D & 541 & 23 [20, 27] & 2 & 2 & 0 & 15 & 4 & 61 [57, 65] & 46 & 15 & 0 & 16 & 159 & 81 & 4 & 15 \\
Qwen3-VL-8B & 473 & 96 [94, 98] & 0 & 0 & 0 & 96 & 0 & 2 [1, 3] & 2 & 0 & 0 & 2 & 345 & 99 & 1 & 1 \\
InternVL3.5-8B & 177 & 83 [77, 89] & 2 & 4 & 0 & 67 & 10 & 6 [3, 9] & 4 & 2 & 0 & 11 & 306 & 94 & 2 & 4 \\
\midrule
All seven (all sets) & 3014 & 57 [55, 59] & 5 & 6 & 1 & 38 & 7 & 24 [22, 25] & 19 & 5 & 0 & 20 & 2502 & 75 & 9 & 16 \\
\bottomrule
\end{tabular}}
\end{table*}

\paragraph{False body relations and encoded visual truth.}
To separate perceptual transcription errors from downstream conclusion errors across the core seven models, two text-only judges audited the last spatial relation before the concluding sentence across all $3{,}014$ \texttt{D+C$-$} losses on the cue-factorial set, occlusion twins, and \texttt{Synthetic Depth} $431$ (reason-basis agreement $96.4\%$, $\kappa=0.92$; grammatical form $94.4\%$, $\kappa=0.94$; conclusion answer $100\%$; \cref{fig:cues}b, \cref{tab:supp_e54a}). Overall, $56.6\%$ $[54.7, 58.6]$ of losses end in a false body relation, $23.8\%$ $[22.2, 25.4]$ write a true body relation whose conclusion mis-maps it to the question ($19\%$ question--relation mapping errors and $5\%$ conclusion reversals; under $1\%$ pair a true relation with a readout that departs from the text), and $19.6\%$ state no explicit relation. Rescues (\texttt{D$-$C$+$}, $n=2{,}502$) mirror this pattern, with $75.3\%$ ending in a true body relation. Among losses with a relation, Stage-1 false relations dominate in six of the seven core models ($57.3\%$ in \texttt{Phi-3.5-Vision} to $98.1\%$ in \texttt{Qwen3-VL-8B}) and account for $145/145$ of relation-bearing cue-set losses in \texttt{Qwen3-VL-32B} (\cref{tab:supp_e57}), whereas Stage-3 conclusion mis-mapping dominates in \texttt{Molmo-7B-D} ($72.4\%$ of relation-bearing losses, $61\%$ of all losses). Among false relations, the text most often misreads the occlusion overlap ($37.9\%$ of all losses), followed by vertical position ($5.6\%$), affirmation without a visual reason ($5.4\%$), and size ($0.6\%$). Across these categories, $69.5\%$ of false relations affirm the queried subject. Body relations are not uniformly worse than \texttt{Direct}. On the cue-balanced set, once conclusions are removed, body relations are at least as accurate as \texttt{Direct} in all seven core models, so losses there arise at the conclusion (\cref{sec:answer_step}), whereas the transcription deficit concentrates where \texttt{Direct} reads a strong correlated or natural cue (\cref{sec:consequences}). These transcription errors occur even while mid-layer representations retain the visual truth. An $\ell_2$ logistic probe on the layer-$20$ residual stream at the first claim's relation token, fit within strata that write the same claim, still decodes the true depth order with $0.88$--$0.93$ balanced accuracy across the three core models probed ($0.82$ at layer $36$ in \texttt{Qwen3-VL-32B}; gray-image and twin-swap controls in \cref{sec:supp_a2,tab:supp_probe_align}), though linear steering along the probe direction gives no net accuracy gain (\cref{tab:supp_steering_nonoracle}).

\section{Analysis 3: The Answer Follows the Text}
\label{sec:answer_step}

\begin{figure}[t]
\centering
\includegraphics[width=0.92\columnwidth]{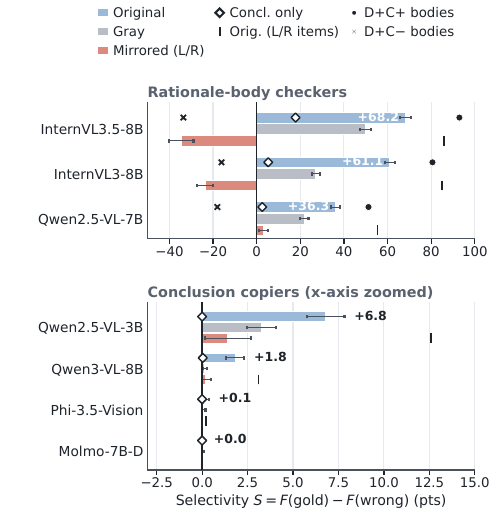}
\caption{\textbf{Conclusion-swap selectivity $S=F(\text{gold})-F(\text{wrong})$ (points) after truncated rationales $R_{\text{cut}}$ across the core seven models.} Here $F(\text{gold})$ and $F(\text{wrong})$ denote how often the answer equals the option stated by an appended concluding sentence when that option is the gold versus the wrong one ($\%$); copiers have $S\approx0$. Original (blue) and gray-image (gray) bars cover all main-pool items with a parsable rationale; mirrored bars (red) cover left/right items whose truth flips under horizontal mirroring (vertical ticks mark $S_{\text{orig}}$ on the same subset). Diamonds mark conclusion-only prompts (\texttt{concl}), and dots / crosses show $S$ for $R_{\text{cut}}$ from \texttt{D+C$+$} / \texttt{D+C$-$} items. The bottom panel zooms the x-axis for the conclusion copiers.}
\label{fig:swaps}
\end{figure}

\begin{figure*}[t]
\centering
\includegraphics[width=0.92\textwidth]{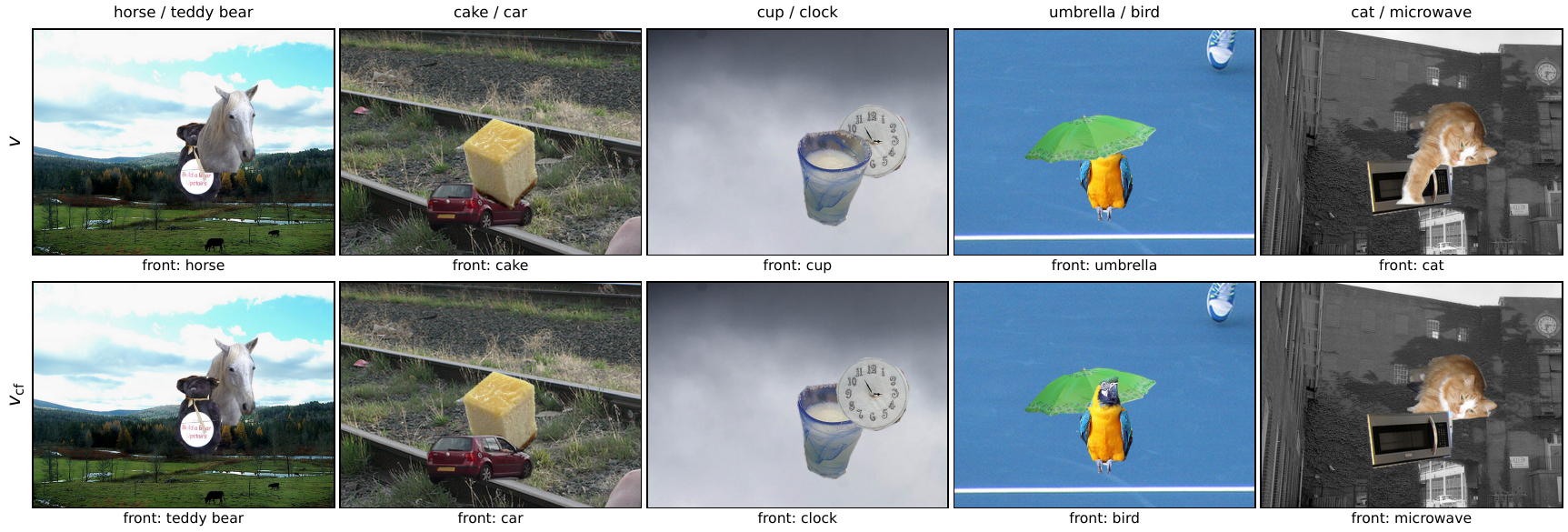}
\caption{\textbf{Natural-object occlusion twins} (COCO composites), where each pair differs only in which object is pasted on top.}
\label{fig:coco_examples}
\end{figure*}

\paragraph{Stage 2: Fixed-rationale twin swaps and truncation sensitivity.}
A coherent reasoner should follow its own text, so following $R_\mathrm{cut}$ is not itself diagnostic; the informative test is whether the image still matters once a relation is written. Our primary, unselected test holds each model's own conclusion-removed rationale ($R_\mathrm{cut}$, with its concluding answer sentence removed) fixed on the $800$ occlusion-twin queries and compares the forced readout under the original image $v$ versus its identity-swapped twin $v_\mathrm{cf}$. Under \texttt{Direct}, the twin image effect ranges from $3.0$ to $76.5$ points across the core seven models ($92.8$ in \texttt{Qwen3-VL-32B}). Under the fixed truncated rationale $R_\mathrm{cut}$, it drops to $2.0$--$8.9$ points across the core seven models (a $7$--$9\times$ reduction in the two core models that read occlusion) and to $0.1$ points in \texttt{Qwen3-VL-32B} (\cref{tab:profile,tab:supp_e57}), while three frontier closed models still follow a contradicting image under their own fixed text (\cref{sec:frontier}). On \texttt{D+C$+$} rationales without error selection, writing the first true relation already drops the prefix twin image effect from $47.1$ $[44.1, 50.1]$ to $26.9$ $[24.2, 29.7]$ in \texttt{Qwen3-VL-8B} ($n=986$). It likewise drops the effect from $70.8$ $[68.2, 73.2]$ to $32.6$ $[30.5, 34.7]$ in \texttt{Qwen3-VL-32B} ($n=1{,}345$), and to $7.8$ and $9.6$ two sentences later (\cref{tab:supp_e57}). Within \texttt{D+C$-$} rationales (exploratory and selection-constrained because $P_j(\text{gold}_v\mid v)<0.5$ after $j^\ast$ by definition), swapping in the twin moves the prefix readout by $64.6$ points before $j^\ast$ versus $3.7$ after it in \texttt{Qwen3-VL-8B} ($86.9\to3.0$ in \texttt{Qwen3-VL-32B}; \cref{tab:supp_e57}).

To check whether unlettered concluding clauses left in $R_\mathrm{cut}$ account for this lock-in, we re-read the $800$ occlusion-twin queries under two stricter deterministic conclusion-removal rules (\cref{sec:supp_engine,tab:supp_x2_trunc}): a \emph{uniform} rule (stripping answer headers and trailing concluding sentences) and a \emph{strict} rule (iteratively removing trailing sentences with mid-sentence conclusion clauses). Across six of the eight models, the residual image effect stays within about $5$ points of $R_\mathrm{cut}$; it rises most in models with multi-sentence conclusions, \texttt{Qwen3-VL-32B} ($0.1\to14.1$ uniform, $24.2$ strict) and \texttt{Molmo-7B-D} ($8.9\to15.5$, $15.9$), while staying far below \texttt{Direct} ($26.8$--$92.8$ points) in every model whose \texttt{Direct} answers use the image.

\begin{table*}[t]
\centering
\caption{\textbf{Fixed rationale, mirrored natural image} (\texttt{CV-Bench} 2D Relation left/right, $n=151$). Image effect $P(\text{answer}=\text{gold}_v\mid v)-P(\text{answer}=\text{gold}_v\mid \text{mirror}(v))$ in points ($100$ follows the image; $0$ ignores it) under three fixed contexts: \texttt{Direct}, the model's own conclusion-removed rationale ($R_\mathrm{cut}$), and the full rationale $R$. Brackets show image bootstrap CIs (the $151$ items come from $151$ distinct photos, with one resample shared across all models, contexts, and the mean). The last columns break down the $R_\mathrm{cut}$ effect by the truth of its last written relation (two text-only judges).}
\label{tab:supp_e60}
\scriptsize
\setlength{\tabcolsep}{3pt}
\fitto{\textwidth}{\begin{tabular}{lccccccc}
\toprule
 & Direct acc. & \multicolumn{3}{c}{Image effect $E$ (points)} & $E_{R_\mathrm{cut}}/E_\mathrm{Dir}$ & \multicolumn{2}{c}{$E_{R_\mathrm{cut}}$ by written relation} \\
\cmidrule(lr){3-5} \cmidrule(lr){7-8}
Model & ($v$ / mirror) & Direct & $R_\mathrm{cut}$ & full $R$ & & true ($n$) & false ($n$) \\
\midrule
Qwen2.5-VL-3B & 74.8 / 73.5 & $+48.3$ {\scriptsize$[+38.4, +57.6]$} & $+3.3$ {\scriptsize$[+0.0, +7.3]$} & $+1.3$ & 0.07 & $+1.0$ (104) & $+0.0$ (33) \\
Qwen2.5-VL-7B & 88.1 / 89.4 & $+77.5$ {\scriptsize$[+68.2, +85.4]$} & $+0.7$ {\scriptsize$[+0.0, +2.0]$} & $+0.0$ & 0.01 & $+0.8$ (131) & $+0.0$ (15) \\
InternVL3-8B & 93.4 / 89.4 & $+82.8$ {\scriptsize$[+74.8, +90.1]$} & $+3.3$ {\scriptsize$[+0.7, +6.6]$} & $+19.9$ & 0.04 & $+2.9$ (138) & $+0.0$ (10) \\
Phi-3.5-Vision & 75.5 / 80.1 & $+55.6$ {\scriptsize$[+45.7, +65.6]$} & $+1.3$ {\scriptsize$[-1.3, +4.0]$} & $+0.7$ & 0.02 & $+0.0$ (106) & $+0.0$ (33) \\
Molmo-7B-D & 78.2 / 77.5 & $+55.6$ {\scriptsize$[+45.7, +64.9]$} & $+3.3$ {\scriptsize$[+0.7, +6.6]$} & $+0.0$ & 0.06 & $+1.2$ (84) & $+0.0$ (24) \\
Qwen3-VL-8B & 91.4 / 94.7 & $+86.1$ {\scriptsize$[+79.5, +92.1]$} & $+1.3$ {\scriptsize$[+0.0, +3.3]$} & $+1.3$ & 0.02 & $+0.0$ (141) & $+0.0$ (6) \\
InternVL3.5-8B & 90.1 / 91.4 & $+81.5$ {\scriptsize$[+73.5, +88.7]$} & $+10.6$ {\scriptsize$[+6.0, +16.6]$} & $+20.5$ & 0.13 & $+9.8$ (143) & $+0.0$ (4) \\
\midrule
Mean of 7 & & $+69.6$ {\scriptsize$[+62.3, +75.8]$} & $+3.4$ {\scriptsize$[+2.2, +4.8]$} & $+6.2$ {\scriptsize$[+4.5, +8.0]$} & & $+2.2$ & $+0.0$ \\
\bottomrule
\end{tabular}}
\end{table*}

\begin{table*}[t]
\centering
\caption{\textbf{Natural-object occlusion twins from COCO.} Each twin pair swaps only the paste order ($233$ pairs, $1{,}398$ prompts across closer, farther, and which-is-closer). Both-correct denotes both twins answered correctly (image-blind answering gives $0$). We report the image effect $E$ with the context fixed to \texttt{Direct} or the model's own $R_\mathrm{cut}$, along with the written-relation accuracy of $R_\mathrm{cut}$ vs.\ \texttt{Direct} on the same items (parentheses give the coverage \% of $R_\mathrm{cut}$ with a stated relation) and the share of twin readouts that follow it. \texttt{Qwen3-VL-32B} is excluded from the mean. CIs use a twin-pair bootstrap.}
\label{tab:supp_e62}
\scriptsize
\setlength{\tabcolsep}{2pt}
\fitto{\textwidth}{\begin{tabular}{lcccccccc}
\toprule
 & \multicolumn{3}{c}{Accuracy (\%)} & \multicolumn{3}{c}{Image effect $E$ (points)} & \multicolumn{2}{c}{Written relation of R$_\mathrm{cut}$} \\
\cmidrule(lr){2-4}\cmidrule(lr){5-7}\cmidrule(lr){8-9}
Model & Direct & both-correct & CoT$-$Direct & Direct & R$_\mathrm{cut}$ fixed & ratio & acc.\ / Direct (cov.) & follows, $v_\mathrm{cf}$ \\
\midrule
Qwen2.5-VL-3B & $58.9$\,{\tiny$[57.3, 60.7]$} & $21.2$\,{\tiny$[18.2, 24.5]$} & $-3.0$\,{\tiny$[-5.7, -0.4]$} & $+10.3$\,{\tiny$[+9.3, +11.2]$} & $+1.3$\,{\tiny$[+0.9, +1.6]$} & 0.12 & 55.6 / 62.7 (94) & 91.4 \\
Qwen2.5-VL-7B & $65.5$\,{\tiny$[63.5, 67.3]$} & $31.5$\,{\tiny$[27.6, 35.2]$} & $-2.9$\,{\tiny$[-5.5, -0.4]$} & $+23.1$\,{\tiny$[+21.1, +25.0]$} & $+4.6$\,{\tiny$[+3.9, +5.4]$} & 0.20 & 65.1 / 70.3 (96) & 88.8 \\
InternVL3-8B & $69.0$\,{\tiny$[67.1, 71.0]$} & $38.3$\,{\tiny$[34.5, 42.2]$} & $-4.9$\,{\tiny$[-7.2, -2.6]$} & $+25.5$\,{\tiny$[+23.7, +27.3]$} & $+2.7$\,{\tiny$[+2.1, +3.2]$} & 0.11 & 68.7 / 74.6 (93) & 92.2 \\
Phi-3.5-Vision & $56.7$\,{\tiny$[55.2, 58.2]$} & $18.2$\,{\tiny$[15.6, 20.9]$} & $-3.5$\,{\tiny$[-5.7, -1.3]$} & $+10.7$\,{\tiny$[+9.5, +12.0]$} & $+0.4$\,{\tiny$[+0.0, +0.8]$} & 0.04 & 55.4 / 60.8 (92) & 84.0 \\
Molmo-7B-D & $59.6$\,{\tiny$[58.1, 61.1]$} & $20.6$\,{\tiny$[17.7, 23.5]$} & $+1.0$\,{\tiny$[-1.6, +3.7]$} & $+9.9$\,{\tiny$[+9.0, +10.8]$} & $+1.4$\,{\tiny$[+0.9, +2.0]$} & 0.14 & 70.1 / 63.7 (77) & 84.2 \\
Qwen3-VL-8B & $72.5$\,{\tiny$[70.3, 74.7]$} & $45.2$\,{\tiny$[40.9, 49.6]$} & $+1.4$\,{\tiny$[-1.1, +3.9]$} & $+40.7$\,{\tiny$[+37.4, +43.9]$} & $+2.9$\,{\tiny$[+2.2, +3.7]$} & 0.07 & 81.0 / 79.8 (91) & 95.4 \\
\bottomrule
\end{tabular}}
\end{table*}

\paragraph{Fixed-rationale image swaps on natural photographs.}
Two pre-specified tests verify that the same text lock-in holds outside rendered shapes. First, on $151$ \texttt{CV-Bench} 2D Relation left/right photographs without legible text (\cref{tab:supp_e60}), horizontally mirroring the image flips the ground-truth answer while keeping the scene natural. Under \texttt{Direct}, mirroring changes the answer by $48.3$--$86.1$ points (seven-model mean $+69.6$ $[+62.3, +75.8]$). With each model's own $R_\mathrm{cut}$ fixed, the mirror effect drops to $+0.7$--$+10.6$ points (seven-model mean $+3.4$ $[+2.2, +4.8]$, at most $0.13$ of \texttt{Direct} in every model), and none of the $125$ false written relations is corrected by the image (when the written relation is false for $v$, the model never overrides it under $v$ and follows it under $\text{mirror}(v)$ as well, yielding $0-0=+0.0$ in every model). This pattern holds across all $383$ left/right items without the text filter. Keeping the full rationale $R$ yields a mean of $+6.2$, driven by \texttt{InternVL3-8B} ($+19.9$) and \texttt{InternVL3.5-8B} ($+20.5$), which re-check an appended conclusion.

Second, we construct $233$ natural-object occlusion twins ($1{,}398$ prompts; examples in \cref{fig:coco_examples}, results in \cref{tab:supp_e62}) by pasting two segmented COCO~\citep{lin2014coco} objects ($42$ categories) onto a natural background with $15\%$--$40\%$ overlap and swapping only the paste order (with the front object higher, larger, and on the left in exactly half of the pairs). Under \texttt{Direct}, five models follow paste-order occlusion strongly (image effect $+23.0$ to $+45.6$ points, including \texttt{Qwen3-VL-32B}; \texttt{Molmo-7B-D}, which passes the cue-set rule, reaches $+9.9$). With each model's own $R_\mathrm{cut}$ fixed, the image effect falls to $+0.4$--$+7.4$ points across those five models (seven-model mean $+20.5\to+3.0$; $+5.5$ $[+5.1, +5.9]$ under the strict truncation rule of \cref{sec:supp_engine}), and $84.0\%$--$97.7\%$ of $R_\mathrm{cut}$ readouts under the twin $v_\mathrm{cf}$ still follow the written relation. On these composites, net \texttt{CoT}$-$\texttt{Direct} ranges from $-4.9$ to $+5.5$ points, and its sign matches whether the written relation (two text-only judges, $\kappa=0.97$, $360$-item sample) is more or less accurate than \texttt{Direct} on the same items in all eight models tested.

\begin{table*}[t]
\centering
\caption{\textbf{Unconditional claim edit} over all Direct/CoT cells: accuracy (\%) / $d'$, and differences to \texttt{Direct} (points, $95\%$ CIs).}
\label{tab:supp_claim_uncond}
\scriptsize
\setlength{\tabcolsep}{2pt}
\fitto{\textwidth}{\begin{tabular}{lcccccccc}
\toprule
Model & $n$ (D+C+/D+C$-$/D$-$C+/D$-$C$-$) & Direct & CoT & Original & Neutral & Correct & Neutral$-$Dir. & Correct$-$Dir. \\
 & & \multicolumn{5}{c}{accuracy (\%) / $d'$} & \multicolumn{2}{c}{pts [95\% CI]} \\
\midrule
Qwen2.5-VL-3B & 389/218/87/181 & 69.4 / 1.05 & 54.4 / 0.26 & 54.3 / 0.27 & 57.9 / 0.56 & 85.3 / 2.12 & $-11.4$\,{\scriptsize[$-$15.3,$-$7.7]} & $+15.9$\,{\scriptsize[+12.1,+19.6]} \\
Qwen2.5-VL-7B & 470/112/104/170 & 68.0 / 0.94 & 67.1 / 0.94 & 67.2 / 0.95 & 65.7 / 0.83 & 95.2 / 4.35 & $-2.3$\,{\scriptsize[$-$5.4,+0.9]} & $+27.2$\,{\scriptsize[+24.0,+30.5]} \\
InternVL3-8B & 484/194/102/94 & 77.6 / 1.51 & 67.0 / 0.97 & 66.1 / 0.91 & 65.0 / 0.84 & 97.0 / 4.60 & $-12.6$\,{\scriptsize[$-$16.7,$-$8.6]} & $+19.5$\,{\scriptsize[+16.3,+22.5]} \\
Phi-3.5-V & 259/130/92/117 & 65.1 / 0.75 & 58.7 / 0.39 & 58.4 / 0.37 & 58.2 / 0.32 & 100.0 / 5.86 & $-6.9$\,{\scriptsize[$-$11.1,$-$2.6]} & $+34.9$\,{\scriptsize[+31.4,+38.7]} \\
Molmo-7B-D & 607/175/45/30 & 91.2 / 2.75 & 76.1 / 1.49 & 75.8 / 1.55 & 74.9 / 1.41 & 79.0 / 2.87 & $-16.3$\,{\scriptsize[$-$19.5,$-$13.2]} & $-12.3$\,{\scriptsize[$-$15.6,$-$9.2]} \\
Qwen3-VL-8B & 564/183/44/97 & 84.1 / 2.28 & 68.5 / 1.11 & 68.4 / 1.11 & 70.4 / 1.13 & 98.4 / 4.36 & $-13.7$\,{\scriptsize[$-$17.0,$-$10.6]} & $+14.3$\,{\scriptsize[+11.9,+16.8]} \\
\bottomrule
\end{tabular}}
\end{table*}

\paragraph{Causal first-claim edits.}
Conversely, editing the first written spatial claim causally moves the final answer (\cref{tab:supp_claim_uncond}). At the first sentence of $R$ naming both shapes with an occlusion or front verb on \texttt{Synthetic Depth} and occlusion twins (``in front'' phrasing), we either keep the sentence (\emph{Original}), replace it with a neutral sentence (\emph{Neutral}, deleting the claim), or replace it with the true relation (\emph{Correct}), and let the model continue greedily before the forced readout. Over all four \texttt{Direct}/\texttt{CoT} cells on items with a parsed claim, deleting the claim via \emph{Neutral} does not recover accuracy, staying within $3.5$ points of \texttt{CoT} and $2.3$--$16.3$ points below \texttt{Direct} because the continuation introduces new errors at a similar rate. By contrast, inserting the oracle \emph{Correct} claim beats \texttt{Direct} by $+14.3$ to $+34.9$ points in five of six core models tested, while lowering \texttt{Molmo-7B-D} by $12.3$ points (which depends on the ``in front of'' edit phrasing; \cref{sec:supp_claim_edit}). On the cue set in \texttt{Qwen3-VL-32B} (\texttt{Direct} $96.4\%$ on items with a parsed claim), true or wrong first-claim edits move accuracy to $99.2\%$ or $17.6\%$ (\cref{tab:supp_e57}). Making the same edits inside the full stored $R$ without regeneration yields $0.0\%$--$16.5\%$ accuracy on \texttt{D+C$-$} items, showing that the edited claim acts through the continuation generated after it (commit-sentence and phrasing-form edits are detailed in \cref{sec:supp_claim_edit}).

\paragraph{Conclusion-blocked knockouts and text checking.}
Attention-edge knockouts at the forced answer prompt confirm where the answer reads from. On $R_\mathrm{cut}$ in all seven core models, removing the image from every text position flips $2.1\%$--$9.0\%$ of answers, against $26.6\%$--$53.9\%$ under \texttt{Direct}, whereas blocking attention to the last written relation flips $10\%$--$24\%$, $1.9$ to $11$ times more (\cref{tab:supp_e59}). Appending \texttt{``Therefore, the answer is (Y) <option>.''} after $R_{\text{cut}}$ and measuring selectivity $S=F(\text{gold})-F(\text{wrong})$ (where $F(\text{gold})$ and $F(\text{wrong})$ denote how often the answer equals the option stated by the appended conclusion when that option is the gold or the wrong one; \cref{fig:swaps}) separates \emph{conclusion copiers} (\texttt{Molmo-7B-D}, \texttt{Phi-3.5-Vision}, \texttt{Qwen3-VL-8B}, \texttt{Qwen2.5-VL-3B}; $S_{\text{orig}}\le6.8$) from \emph{rationale-body checkers} (\texttt{InternVL3.5-8B}, \texttt{InternVL3-8B}, \texttt{Qwen2.5-VL-7B}; $S_{\text{orig}}=36.3$--$68.2$). Splitting $S$ by the source cell of $R_\mathrm{cut}$ (\cref{tab:supp_recheck}) reveals that checkers accept conclusions matching the body rather than independently re-grounding in the image: $S$ is strongly positive after bodies from \texttt{D+C$+$} items ($+51.3$ to $+92.9$) and negative after bodies from \texttt{D+C$-$} items ($-16.1$ to $-33.6$). Checkers do use the image to reject an appended conclusion that contradicts it: on 2D left/right items, removing the image lowers $S$ from $86.7$ to $36.9$ in \texttt{InternVL3-8B} (\cref{tab:supp_e59}). Yet when their own full rationale is held fixed, swapping in the twin on wrong-claim items moves the answer by $-1.0$ to $+0.3$ points in five models, $+5.2$ $[-1.7, +12.1]$ in \texttt{InternVL3.5-8B}, and $+10.2$ $[+6.3, +14.7]$ in \texttt{InternVL3-8B} (\cref{tab:supp_recheck}), and under their own $R_\mathrm{cut}$ removing the image changes at most $1.2\%$ of 2D left/right answers (\cref{tab:supp_e59}).

\paragraph{Stage 3: Conclusion mis-mapping on converse questions.}
In the $23.8\%$ of \texttt{D+C$-$} losses whose body states a true relation (\cref{tab:supp_e54a}), conclusion mis-mapping occurs mainly on converse phrasings. After a true body relation, conclusions err on $29.5\%$ of converse questions (\eg, answering ``Is K behind F?'' after writing ``F is in front of K'') versus $2.4\%$ of direct restatements, spanning \texttt{Molmo-7B-D} ($31.8\%$), \texttt{InternVL3-8B} ($46.5\%$), \texttt{Phi-3.5-Vision} ($52.4\%$), and \texttt{Qwen2.5-VL-7B} ($24.6\%$). Conclusion mis-mapping accounts for $61\%$ of losses in \texttt{Molmo-7B-D} ($46\%$ mapping and $15\%$ reversal). After an edited true claim ``F is in front of K'', \texttt{Molmo-7B-D} answers accurately on identical restatements ($99.0\%$) and negations ($96.3\%$), but drops to $17.1\%$ $[11.5, 22.6]$ on the converse ``Is K behind F?'' (\cref{tab:supp_e54a,sec:supp_claim_edit}).

\section{Analysis 4: When Short CoT Helps or Hurts}
\label{sec:consequences}

\begin{figure}[t]
\centering
\includegraphics[width=0.86\columnwidth]{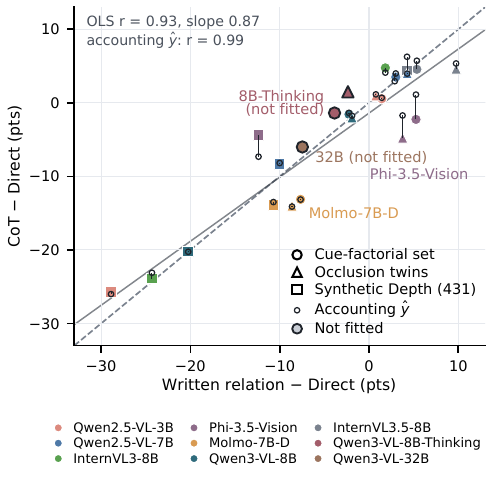}
\caption[Net effect of CoT follows the written relation]{\textbf{Reading the net effect of \texttt{CoT} through written-relation accuracy.} Each point represents one model on one synthetic occlusion set. The horizontal axis $x$ is the accuracy of the last written relation (rule-parsed, conclusion kept; conclusion-removed analysis in \cref{tab:supp_e45}) minus \texttt{Direct} accuracy on the same items, and the vertical axis $y$ is \texttt{CoT}$-$\texttt{Direct} accuracy over all items (points). The solid line shows the linear fit over the $21$ core-model points, the dashed line marks $y=x$, and hollow circles mark the accounting value $\hat y$ from answer--relation agreement rates. Black-edged markers denote \texttt{Qwen3-VL-32B} (cue set) and \texttt{Qwen3-VL-8B-Thinking} (twins), which are not included in the fit.}
\label{fig:two_channel}
\end{figure}

\paragraph{Decomposing the net CoT effect.}
Because the final answer follows the written relation (Stage 2), we decompose the net change on items with a parsed relation as $\texttt{CoT}-\texttt{Direct} = x + A$. Here $x = W - D$ is written-relation accuracy minus \texttt{Direct} accuracy (Stage 1), and $A=(1-a_F)(1-W)-(1-a_T)W$ captures mismatches between the written relation and the final answer from agreement rates $a_T, a_F$ (Stage 3, plus unparsed items). Because both axes subtract $D$, even a within-set shuffle of $W$ gives $r=0.72$ (vs.\ $0.93$ in \cref{fig:two_channel}), so the decomposition's value lies in separating Stage 1 ($x$) from Stage 3 ($A$) once conclusions are removed.

\begin{table*}[t]
\centering
\caption{\textbf{Two-channel analysis under one conclusion-removal protocol.} All points use the uniform conclusion-removal rule (\cref{sec:supp_engine}) and the same two text-only judges; columns and fit statistics are defined in the footer. Exact accounting: on covered items \texttt{CoT}$|$c$\,-\,$\texttt{Dir.}$|$c $=x+A$, so $y=c\,(x+A)+(1-c)\,\Delta_u$ with $c=$ Cov.\ and $\Delta_u=$ \texttt{CoT}$-$\texttt{Dir.} on items without a written relation (residual $<0.001$ at every point). Without the second term, $|y-(x+A)|\le1.5$ at $30$ of the $36$ open-model points of \cref{tab:supp_e45,tab:supp_e35}; the misses are low-coverage cells, \eg, \texttt{Phi-3.5-Vision} \texttt{Syn.\ Depth} ($c=0.57$, $\Delta_u=-12.8$: $-6.5$), \texttt{Molmo-7B-D} \texttt{CV-Bench} ($-5.2$), \texttt{Qwen2.5-VL-7B} \texttt{BLINK} ($-4.2$).}
\label{tab:supp_e45}
\scriptsize
\setlength{\tabcolsep}{2pt}
\fitto{\textwidth}{\begin{tabular}{llrrrrrrrrrrr}
\toprule
Model & Set & $n$ & Cov. & $W$ & \texttt{Dir.} & \texttt{Dir.}$|$c & \texttt{CoT} & Agr.\ T/F & $x$ & $y$ & $A$ & \texttt{D+C$-$}/\texttt{D$-$C+} \\
\midrule
Qwen2.5-VL-3B & Cue set & 1600 & 80.1 & 56.3 & 52.4 & 51.9 & 53.0 & 81/84 & $+4.4$ & $+0.6$ & $-3.4$ & 226/235 \\
Qwen2.5-VL-3B & Twins & 800 & 80.9 & 53.6 & 51.5 & 51.2 & 52.5 & 82/82 & $+2.5$ & $+1.0$ & $-1.5$ & 117/125 \\
Qwen2.5-VL-3B & Syn.\ Depth 431 & 431 & 83.8 & 59.0 & 84.0 & 83.7 & 58.2 & 98/93 & $-24.7$ & $-25.8$ & $+1.7$ & 127/16 \\
Qwen2.5-VL-7B & Cue set & 1600 & 89.9 & 55.7 & 53.4 & 52.7 & 56.9 & 88/82 & $+3.0$ & $+3.5$ & $+0.9$ & 197/253 \\
Qwen2.5-VL-7B & Twins & 800 & 89.4 & 60.1 & 54.0 & 53.4 & 58.0 & 85/85 & $+6.7$ & $+4.0$ & $-2.9$ & 95/127 \\
Qwen2.5-VL-7B & Syn.\ Depth 431 & 431 & 91.2 & 70.7 & 78.7 & 77.6 & 70.3 & 91/78 & $-6.9$ & $-8.4$ & $-0.3$ & 73/37 \\
InternVL3-8B & Cue set & 1600 & 83.2 & 76.9 & 63.7 & 64.5 & 68.4 & 82/66 & $+12.4$ & $+4.8$ & $-6.3$ & 216/292 \\
\end{tabular}}
\end{table*}

\cref{tab:supp_e45} applies one uniform conclusion-removal protocol (\cref{sec:supp_engine}) and the same two text-only judges ($4$-way agreement $95.7\%$, $\kappa=0.92$) across all synthetic open-model and closed-model points. Under this uniform protocol, the $21$ synthetic open-model points have OLS slope $0.75$ $[0.55, 0.90]$ and intercept $-4.6$ $[-8.2, -2.1]$ (items and model clusters resampled; $r=0.89$). When concluding sentences are removed, the remaining body relation on the cue-controlled set is at least as accurate as \texttt{Direct} in all seven core models ($x \ge +0.8$), whereas Stage-3 conclusion mis-mapping creates a negative answer-step term $A$ pooled over the three synthetic sets reaching $-15.6$ $[-17.2, -14.1]$ points in \texttt{Molmo-7B-D}, $-5.5$ in \texttt{InternVL3-8B}, $-4.1$ in \texttt{Phi-3.5-Vision}, $-4.0$ in \texttt{InternVL3.5-8B}, and $-2.1$ to $-0.4$ in the three Qwen models. This Stage-3 loss concentrates on the cue set and twins, whereas $A \ge -0.8$ on \texttt{Synthetic Depth} $431$ except in \texttt{Molmo-7B-D} ($-6.0$), and closed models show $|A| \le 0.1$ in every cell.

\begin{table*}[t]
\centering
\caption{\textbf{Natural depth questions: two-channel test out of sample.} Cov.\ is the share of items with a consensus depth claim after conclusion removal; \texttt{Dir.}$|$p and $y|$p are evaluated on those items; $W$ is written-relation accuracy; Agr.\ T/F is the rate at which the answer agrees with a true or false written relation (\%); $\hat y$ is the synthetic linear prediction; $^\dagger$not counted in the fit evaluation.}
\label{tab:supp_e35}
\scriptsize
\setlength{\tabcolsep}{2pt}
\fitto{\textwidth}{\begin{tabular}{llrrrrrrrrrrrr}
\toprule
Model & Data & Cov. & \texttt{Dir.} & \texttt{Dir.}$|$p & \texttt{CoT} & $W$ & Agr.\ T/F & $x$ & $y$ & $y|$p & $\hat y$ & $y-\hat y$ & \texttt{D+C$-$}/\texttt{D$-$C+} \\
\midrule
Qwen2.5-VL-3B & CV-Bench 3D Depth & 96.0 & 78.8 & 79.3 & 70.8 & 71.5 & 98/98 & $-7.8$ & $-8.0$ & $-8.5$ & $-8.2$ & $+0.2$ & 94/46 \\
Qwen2.5-VL-7B & CV-Bench 3D Depth & 98.5 & 85.0 & 85.3 & 82.7 & 82.7 & 100/99 & $-2.5$ & $-2.3$ & $-2.4$ & $-3.6$ & $+1.3$ & 56/42 \\
InternVL3-8B & CV-Bench 3D Depth & 99.7 & 85.5 & 85.6 & 86.2 & 86.0 & 99/94 & $+0.3$ & $+0.7$ & $+0.5$ & $-1.1$ & $+1.8$ & 28/32 \\
Phi-3.5-V & CV-Bench 3D Depth & 99.5 & 76.3 & 76.4 & 53.3 & 53.8 & 99/100 & $-22.6$ & $-23.0$ & $-22.8$ & $-21.1$ & $-1.9$ & 210/72 \\
Molmo-7B-D & CV-Bench 3D Depth & 55.2 & 68.0 & 67.7 & 71.0 & 78.5 & 96/97 & $+10.9$ & $+3.0$ & $+8.2$ & $+8.1$ & $-5.1$ & 74/92 \\
Qwen3-VL-8B & CV-Bench 3D Depth & 99.3 & 95.2 & 95.1 & 92.3 & 93.0 & 99/100 & $-2.2$ & $-2.8$ & $-2.9$ & $-3.3$ & $+0.5$ & 26/9 \\
InternVL3.5-8B & CV-Bench 3D Depth & 98.3 & 87.2 & 88.0 & 87.2 & 86.1 & 98/79 & $-1.9$ & $+0.0$ & $-0.5$ & $-3.0$ & $+3.0$ & 24/24 \\
Qwen3-VL-8B-Think.$^\dagger$ & CV-Bench 3D Depth & 93.5 & 94.0 & 94.3 & 94.5 & 94.8 & 100/100 & $+0.5$ & $+0.5$ & $+0.5$ & $-0.9$ & $+1.4$ & 14/17 \\
\midrule
Qwen2.5-VL-3B & BLINK Rel. Depth & 64.5 & 66.1 & 65.0 & 55.6 & 53.8 & 100/100 & $-11.2$ & $-10.5$ & $-11.2$ & $-11.2$ & $+0.7$ & 27/14 \\
Qwen2.5-VL-7B & BLINK Rel. Depth & 91.1 & 82.3 & 82.3 & 71.0 & 72.6 & 100/90 & $-9.7$ & $-11.3$ & $-7.1$ & $-9.9$ & $-1.4$ & 22/8 \\
InternVL3-8B & BLINK Rel. Depth & 87.9 & 80.6 & 84.4 & 76.6 & 80.7 & 99/100 & $-3.7$ & $-4.0$ & $-4.6$ & $-4.6$ & $+0.6$ & 11/6 \\
Phi-3.5-V & BLINK Rel. Depth & 0.0 & 74.2 & -- & 68.5 & \multicolumn{3}{c}{not parseable (empty rationales)} & $-5.6$ & -- & -- & -- & 8/1 \\
Molmo-7B-D & BLINK Rel. Depth & 41.9 & 66.1 & 63.5 & 55.6 & 53.8 & 93/92 & $-9.6$ & $-10.5$ & $-9.6$ & $-9.8$ & $-0.7$ & 34/21 \\
Qwen3-VL-8B & BLINK Rel. Depth & 92.7 & 87.9 & 89.6 & 83.1 & 86.1 & 100/100 & $-3.5$ & $-4.8$ & $-3.5$ & $-4.4$ & $-0.4$ & 11/5 \\
InternVL3.5-8B & BLINK Rel. Depth & 91.9 & 78.2 & 78.9 & 75.0 & 76.3 & 100/96 & $-2.6$ & $-3.2$ & $-1.8$ & $-3.7$ & $+0.5$ & 16/12 \\
Qwen3-VL-8B-Think.$^\dagger$ & BLINK Rel. Depth & 79.0 & 83.9 & 84.7 & 76.6 & 77.6 & 96/100 & $-7.1$ & $-7.3$ & $-10.2$ & $-7.6$ & $+0.4$ & 14/5 \\
\bottomrule
\multicolumn{14}{p{0.97\linewidth}}{\scriptsize Written relation $W$: the option that the last depth claim of the CoT rationale names as closer to the camera, after removing the answer and concluding sentences (uniform rule, \cref{sec:supp_engine}); text-only consensus of Claude Opus 5.5 and GPT-6-Astra (agreement 99.0\%, $\kappa=0.98$); Cov.: share of items with a consensus claim. $x = W - $\texttt{Dir.}$|$p (parsed items); $y=$ \texttt{CoT}$-$\texttt{Dir.} (all items); $y|$p on parsed items. $\hat y = -1.40 + 0.871\,x$: synthetic fit of \cref{tab:supp_two_channel} (21 points), not refit. This fit uses rule-parsed $x$ on conclusion-kept synthetic text, whereas $x$ here is judge-extracted after conclusion removal. The protocol-matched synthetic fit of \cref{tab:supp_e45} ($-4.55+0.745\,x$) predicts the same $13$ points worse: MAE $2.8$ (bias $+2.4$; $+1.4$ $[+0.5, +1.9]$ over the fit used here, image bootstrap), sign $11/13$. $y=x$ predicts best ($1.3$) because natural points carry no answer-step loss ($A\approx0$). Closed points: sign $4/13$ (this fit), $2/13$ (matched), $13/13$ ($y=x$). Out of sample (13 instruct points): MAE 1.4 pts vs.\ 5.0 (predict mean synthetic $y$; difference [$-$5.2, $-$1.7]), 6.5 (predict 0; [$-$6.4, $-$2.6]) and 1.3 ($y=x$; [$-$0.3, +0.3]); $r(\hat y, y) = 0.96$ [0.78, 0.97]; sign agreement 11/13 (misses: InternVL3-8B and InternVL3.5-8B on CV-Bench, $|y| \le 0.7$); mean $y-\hat y$ $-0.1$ [$-$1.3, +1.2]. Over the $13$ instruct points, $x<0$ in $11$ (the body relation is less accurate than \texttt{Direct}), and the answer-step term $A$ of \cref{tab:supp_e45} is within $\pm1.4$ points in $11$ ($-2.7$ \texttt{Molmo-7B-D} CV-Bench, $+2.7$ \texttt{Qwen2.5-VL-7B} BLINK). Brackets: image bootstrap (2,000), MAE differences paired. Agr.\ T/F: CoT answer agrees with the written relation when it is true / false (\%). $^\dagger$Native thinking trace, not in the out-of-sample statistics.} \\
\end{tabular}}
\end{table*}

On natural depth questions (\texttt{CV-Bench} 3D Depth, $600$ items, and \texttt{BLINK} Relative Depth, $124$ items; conclusions removed, two text-only judges agreeing on $99.0\%$, $\kappa=0.98$ over $5{,}607$ pairs; \cref{tab:supp_e35}), $A$ is near zero (within $\pm1.4$ points in $11$ of $13$ instruct cells), and answers follow the body relation on $93\%$--$100\%$ of true relations and at least $90\%$ of false relations in $12$ of $13$ cells. The written body relation is less accurate than \texttt{Direct} in $11$ of $13$ instruct model--benchmark pairs, and the unfitted synthetic line of \cref{fig:two_channel} predicts \texttt{CoT}$-$\texttt{Direct} with a mean absolute error of $1.39$ points ($r=0.96$, sign agreement $11/13$), while $y=x$ alone gives $1.31$ points (the protocol-matched uniform synthetic line gives $2.8$ because natural points carry no Stage-3 intercept loss). The one large miss is \texttt{Molmo-7B-D} on \texttt{CV-Bench} ($-5.1$), where $45\%$ of rationales state no depth claim. On the COCO occlusion composites (\cref{tab:supp_e62}), the sign of $x$ likewise matches the sign of \texttt{CoT}$-$\texttt{Direct} in all eight open models tested: the four that write less accurate relations than \texttt{Direct} lose $2.9$--$4.9$ points and the four that write more accurate ones change by $+1.0$ to $+5.5$.

\begin{table*}[t]
\centering
\caption{\textbf{Held-out accuracy by task (\texttt{Direct}/\texttt{CoT}, \%).} \texttt{BLINK} val Spatial Relation ($143$) and Relative Depth ($124$); \texttt{3DSRBench} COCO images, five categories with $100$ items each (one item per image, seed $0$). Predictions and the gate threshold were frozen before the CoT, vote, and description runs. \texttt{Direct} on \texttt{3DSR Left} (\texttt{orientation\_on\_the\_left}) is below chance in every model.}
\label{tab:supp_heldout}
\scriptsize
\fitto{\textwidth}{\begin{tabular}{lcccccccc}
\toprule
Model & BLINK Rel. & BLINK Depth & 3DSR Closer & 3DSR Above & 3DSR Higher & 3DSR Front & 3DSR Left & Pooled \\
\midrule
Qwen2.5-VL-3B & 82.5/70.6 & 66.1/55.6 & 67.0/69.0 & 49.0/57.0 & 47.0/55.0 & 59.0/63.0 & 32.0/39.0 & 59.2/59.1 \\
Qwen2.5-VL-7B & 87.4/75.5 & 82.3/71.0 & 74.0/73.0 & 59.0/65.0 & 51.0/51.0 & 64.0/63.0 & 28.0/31.0 & 65.6/62.5 \\
InternVL3-8B & 88.8/84.6 & 80.6/76.6 & 82.0/78.0 & 60.0/67.0 & 52.0/69.0 & 67.0/64.0 & 29.0/33.0 & 67.4/68.7 \\
Phi-3.5-V & 67.8/62.2 & 74.2/68.5 & 68.0/72.0 & 50.0/49.0 & 48.0/56.0 & 58.0/61.0 & 39.0/42.0 & 58.9/59.2 \\
Molmo-7B-D & 69.2/69.9 & 66.1/55.6 & 62.0/63.0 & 48.0/54.0 & 54.0/57.0 & 60.0/54.0 & 41.0/49.0 & 58.1/58.1 \\
Qwen3-VL-8B & 88.1/85.3 & 87.9/83.1 & 88.0/82.0 & 66.0/69.0 & 53.0/65.0 & 68.0/65.0 & 32.0/32.0 & 70.7/70.1 \\
InternVL3.5-8B & 85.3/85.3 & 78.2/75.0 & 78.0/73.0 & 54.0/62.0 & 54.0/60.0 & 71.0/65.0 & 30.0/33.0 & 66.0/66.2 \\
\bottomrule
\end{tabular}}
\end{table*}

\paragraph{Shortcuts, stress tests, and natural benchmarks.}
Across datasets, the net effect of prompted short CoT depends on whether Stage-1 cue re-weighting helps or hurts more than Stage-3 mapping costs. On the cue-factorial set, where ``lower is closer'' is wrong on half the items, weakening that shortcut helps the height-reliant core models gain $+0.6$ to $+4.8$ points (\cref{tab:profile}). On the size-confounded \texttt{Synthetic Depth} stress test, where \texttt{Direct} achieves high accuracy via the correlated size cue, verbalization dilutes that valid cue and drops accuracy by $8.4$--$25.8$ points in five of the seven core models. Excluding this stress test, three of the eight main-pool open models show a significant loss and one shows a significant gain (\cref{tab:supp_e58b}). On standard natural-image benchmarks, however, shortcut relief and transcription errors largely cancel out: although individual answers churn heavily across items (\cref{tab:supp_churn}b), net \texttt{CoT}$-$\texttt{Direct} accuracy on the frozen $767$-item held-out pool (\texttt{BLINK} and \texttt{3DSRBench}; \cref{tab:supp_heldout}) lies between $-3.1$ and $+1.3$ points with no per-model CI excluding zero (\cref{tab:profile}). Task-level breakdowns in \cref{tab:supp_heldout} explain why the pooled average is near zero. Across models, \texttt{CoT} lowers accuracy on \texttt{BLINK Relative Depth} ($-7.2$ $[-10.4, -4.0]$ over eight open models; $-5.5$ $[-9.0, -2.1]$ vs.\ \texttt{Filler+turn} over the core seven) and on \texttt{BLINK Spatial Relation} ($-4.4$ $[-7.4, -1.2]$). Accuracy also drops on the post hoc $324$-item depth-valid subset ($-3.7$ $[-6.3, -1.0]$ over the core seven, or $-4.3$ without \texttt{Phi-3.5-Vision}'s empty \texttt{BLINK} rationales, whose empty-prefix format effect alone costs $-4.8$ $[-9.7, -0.8]$). Conversely, \texttt{CoT} improves accuracy on \texttt{3DSRBench} height ($+8.6$ $[+4.1, +13.4]$), above ($+5.4$ $[+0.5, +10.1]$), and left ($+3.5$ $[+0.5, +6.6]$, where \texttt{Direct} is below chance at $28\%$--$41\%$; \cref{sec:supp_heldout}). On the multi-image \texttt{VSI-Bench}~\citep{yang2025thinking} and \texttt{MindCube-Tiny}~\citep{wang2026mindcube} benchmarks (\cref{tab:supp_e51}), the six-model net effect is likewise null ($-0.6$ and $+0.1$) while answers still follow false written claims ($87.6\%$--$95.9\%$) and $y$ tracks $x$ ($r=0.84$).

\subsection{Larger and reasoning-tuned open cohorts}
\label{sec:open_cohorts}

\begin{table*}[t]
\centering
\caption{\textbf{Summary of larger instruct ($32$B--$38$B) and reasoning-tuned open models on the cue-factorial set ($1{,}600$ queries).} Cells are compiled from the full cohort tables in \cref{tab:supp_e31,tab:supp_e43,tab:supp_e48} (plus \cref{tab:supp_e57,tab:supp_x2_trunc} for fixed-$R_\mathrm{cut}$ twin image effect $E$). Lower: share of \texttt{Direct} answers naming the lower shape as closer. Height / Occl.\ AME $\Delta$: probability-scale average marginal effect shift (writing arm$-$\texttt{Direct}, points). $x$: written-relation accuracy minus \texttt{Direct} on covered items (conclusion kept; rule-parsed for the first four rows, two-judge consensus for \texttt{MiMo-VL-7B-RL} and \texttt{GLM-4.1V-9B-Thinking}; under uniform conclusion removal in \cref{tab:supp_e45}, $x=-5.7$ for \texttt{Qwen3-VL-32B} and $-1.4$ for \texttt{Qwen3-VL-8B-Thinking}); $y$: writing arm$-$\texttt{Direct} accuracy on all items. $^\ast$: $95\%$ twin-pair bootstrap CI excludes zero; --: not run.}
\label{tab:cohorts_summary}
\scriptsize
\setlength{\tabcolsep}{3.5pt}
\fitto{\textwidth}{\begin{tabular}{@{}llcccccccc@{}}
\toprule
\textbf{Model} & \textbf{Arm} & \textbf{\texttt{Direct} (\%)} & \textbf{Reads occl.} & \textbf{Lower (\%)} & \textbf{Height AME $\Delta$} & \textbf{Occl.\ AME $\Delta$} & \textbf{$x$ (pts)} & \textbf{$y$ (pts)} & \textbf{Twins $R_\mathrm{cut}$ $E$} \\
\midrule
Qwen3-VL-32B & short CoT & $96.3$ & \textbf{yes} & $52.7$ & $-17.7^\ast$ & $-12.0^\ast$ & $-7.5$ & $-6.0^\ast$ & $0.1$ \\
InternVL3.5-38B & short CoT & $64.2$ & no & $84.9$ & $-53.0^\ast$ & $+23.1^\ast$ & $+9.9$ & $+11.6^\ast$ & -- \\
Qwen2.5-VL-32B & short CoT & $61.7$ & no & $72.3$ & $-30.6^\ast$ & $-10.5^\ast$ & $-3.4$ & $-5.6^\ast$ & -- \\
\midrule
Qwen3-VL-8B-Thinking & think & $88.7$ & \textbf{yes} & $59.4$ & $-26.2^\ast$ & $-2.6$ & $-3.9$ & $-1.4$ & -- \\
MiMo-VL-7B-RL & think & $73.9$ & \textbf{yes} & $74.3$ & $-24.9^\ast$ & $+11.2^\ast$ & $+6.5^\ast$ & $+5.7^\ast$ & -- \\
MiMo-VL-7B-RL & short CoT & $73.9$ & \textbf{yes} & $74.3$ & $-15.8^\ast$ & $+6.2^\ast$ & $+3.8^\ast$ & $+3.2^\ast$ & -- \\
GLM-4.1V-9B-Thinking & think & $73.9$ & \textbf{yes} & $75.9$ & $-19.7^\ast$ & $+16.7^\ast$ & $+8.4^\ast$ & $+8.2^\ast$ & -- \\
GLM-4.1V-9B-Thinking & short CoT & $73.9$ & \textbf{yes} & $75.9$ & $-18.5^\ast$ & $+14.2^\ast$ & $+7.5^\ast$ & $+7.1^\ast$ & -- \\
\bottomrule
\end{tabular}}
\end{table*}

Scale alone does not remove the transcription bottleneck (\cref{tab:cohorts_summary}; full cohort tables in \cref{tab:supp_r29_ladder,tab:supp_e31,tab:supp_e43,tab:supp_e48}). On the \texttt{Qwen3-VL} Instruct scale ladder from $2$B to $32$B (\cref{tab:supp_r29_ladder}), pooled twin $d'$ falls under \texttt{CoT} at every size (from $-0.44$ $[-0.89, -0.10]$ at $2$B to $-1.96$ $[-2.24, -1.66]$ at $32$B). Pair-matched $d'_\mathrm{pair}$ likewise falls at $2$B and $4$B ($-1.12$, $-1.23$) and hits its ceiling bound at $32$B, while accuracy drops at $4$B ($-8.2$) and $32$B ($-6.8$). On the cue-factorial set (\cref{tab:cohorts_summary,tab:supp_e31}), both \texttt{Qwen3-VL-32B} (\texttt{Direct} $96.3\%$) and \texttt{Qwen3-VL-8B-Thinking} (\texttt{Direct} $88.7\%$; native thinking trace mean $193$ tokens) pass the occlusion rule. In \texttt{Qwen3-VL-32B}, the probability-scale occlusion AME drops by $12.0$ points ($92.6\to80.6$, accuracy $-6.0$), while in \texttt{Qwen3-VL-8B-Thinking} it moves by $-2.6$ $[-6.0, +0.8]$ (its question-blind description arm costs $10.1$ points). On the conclusion-kept two-channel axes of \cref{fig:two_channel}, their cue-set coordinates are $(-7.5, -6.0)$ and $(-3.9, -1.4)$ (\cref{fig:two_channel} plots the first and, for \texttt{Qwen3-VL-8B-Thinking}, its twins point $(-2.4, +1.5)$ from \cref{tab:supp_two_channel}; neither model is in the fit, and both also follow the uniform conclusion-removed accounting in \cref{tab:supp_e45}).

The other larger instruct models follow the same accounting (\cref{tab:cohorts_summary,tab:supp_e43}). Neither \texttt{InternVL3.5-38B} ($64.2\%$ $[61.5, 67.0]$) nor \texttt{Qwen2.5-VL-32B} ($61.7\%$ $[59.4, 64.1]$) passes the occlusion rule, and both follow lower-is-closer under \texttt{Direct} ($84.9\%$ and $72.3\%$). Under \texttt{CoT}, the height AME falls in both ($-53.0$ and $-30.6$ points), as in all $13$ open models tested. \texttt{InternVL3.5-38B} writes relations more accurate than \texttt{Direct} ($x=+9.9$) and gains $11.6$ points (occlusion AME $+23.1$, $d'_\mathrm{pair}$ $+0.90$), whereas \texttt{Qwen2.5-VL-32B} writes less accurate relations ($x=-3.4$) and loses $5.6$ points. Both models lie close to $y=x$ ($y-x=+1.6$ and $-2.2$).

Reasoning-tuned open models from other families (\texttt{MiMo-VL-7B-RL} and \texttt{GLM-4.1V-9B-Thinking}; \cref{tab:cohorts_summary,tab:supp_e48}) test both native thinking traces and short \texttt{CoT} after an empty think block. Both pass the occlusion rule at its margin ($73.9\%$ $[71.5, 76.4]$ and $[71.3, 76.4]$) while still leaning on height ($74.3\%$ and $75.9\%$ lower-is-closer, vs.\ $59.4\%$ in \texttt{Qwen3-VL-8B-Thinking} and $52.7\%$ in \texttt{Qwen3-VL-32B}). Unlike the other four occlusion readers, both gain from writing: thinking adds $+5.7$ and $+8.2$ points on the cue set and $+4.0$ and $+9.6$ on the twins (short \texttt{CoT} $+3.2$ to $+7.1$), lowering height AME and raising occlusion AME ($+6.2$ to $+16.7$) in every cell, and thinking also gains $+4.3$ points on \texttt{CV-Bench} 3D Depth (\texttt{BLINK} changes are not significant). Both models still obey Stage 2 and the accounting of Analysis 4: their answers agree with the judge-extracted written relation ($98.6\%$ judge agreement) in $98.9\%$--$99.2\%$ of true and $94.2\%$--$99.3\%$ of false cue-set items, with $|y-x|\le1.0$ on the synthetic sets. They gain because their written relations are more accurate than \texttt{Direct} ($x=+3.8$ to $+9.5$, all CIs above $0$), making them the two exceptions to Stage-1 degradation on the cue-factorial set, occlusion twins, and \texttt{CV-Bench} 3D Depth among models that pass the occlusion rule. Frontier closed models follow the same written-relation accounting (\cref{tab:supp_e45}) yet diverge on both written accuracy and answer-step lock-in, as we examine next.

\section{Frontier Closed Models}
\label{sec:frontier}

\begin{table*}[t]
\centering
\caption{\textbf{Closed API models.} (a) \texttt{Direct}/\texttt{CoT} accuracy (\%) and pooled occlusion $d'$, with lowest-reasoning \texttt{Direct} in brackets. \texttt{Direct} $d'$ of \texttt{Claude Opus 5.5} and \texttt{DeepSeek-V4.1-Flash} uses the $793$ of $800$ parsed replies; accuracies use the $n$ shown, and failed or unparsed replies count as wrong. (b) Hard occlusion set, where ``low'' is the lowest setting each API accepts ($^a$minimal, where $90\%$ of calls still reason; $^b$low, $29\%$; \texttt{DeepSeek-V4.1-Flash} and \texttt{Qwen3.8-Max} emit no reasoning tokens). Here \texttt{Dir.} is the default call with hidden reasoning, so \texttt{CoT}$-$\texttt{Dir.} compares visible-plus-hidden reasoning with hidden reasoning; \texttt{HY4} documents no low-reasoning option (\cref{sec:supp_closed}). CIs use a $95\%$ twin-pair bootstrap.}
\label{tab:supp_closed}
\scriptsize
\setlength{\tabcolsep}{3pt}
\textbf{(a)}\ \fitto{0.95\textwidth}{\begin{tabular}{lccccc}
\toprule
Model & Occl. twins & Syn. Depth & BLINK Depth & BLINK Rel. & $d'$ Dir./CoT \\
\midrule
Claude Opus 5.5 & 99.1 [94.0]/100.0 ($n$=800) & 95.6/95.6 ($n$=250) & 95.2/94.4 ($n$=124) & 93.0/93.7 ($n$=143) & ceiling [ceiling]/ceiling \\
GPT-6-Astra & 100.0 [100.0]/100.0 ($n$=800) & 98.8/98.0 ($n$=250) & 93.4/94.3 ($n$=122) & 88.1/85.3 ($n$=143) & ceiling [ceiling]/ceiling \\
Gemini-4-Argon & 100.0/100.0 ($n$=800) & 98.4/98.0 ($n$=250) & 97.6/97.6 ($n$=124) & 95.8/95.8 ($n$=143) & ceiling/ceiling \\
Qwen3.8-Max & 100.0 [100.0]/100.0 ($n$=800) & 98.4 [98.0]/97.2 ($n$=250) & 90.3 [91.9]/94.4 ($n$=124) & 93.0/93.7 ($n$=143) & ceiling [ceiling]/ceiling \\
DeepSeek-V4.1-Flash & 90.1 [88.1]/90.9 ($n$=800) & 88.0 [86.0]/89.2 ($n$=250) & 86.3 [89.5]/86.3 ($n$=124) & 89.5/91.6 ($n$=143) & 2.66 [2.35]/2.66 \\
HY4 & 99.9/100.0 ($n$=800) & 92.8/92.4 ($n$=250) & 84.7/85.5 ($n$=124) & 88.1/87.4 ($n$=143) & ceiling/ceiling \\
\bottomrule
\multicolumn{6}{l}{\scriptsize Brackets: Direct at each API's lowest reasoning setting: reasoning\_effort=none for DeepSeek-V4.1-Flash and Qwen3.8-Max (0\% of calls reason),} \\
\multicolumn{6}{l}{\scriptsize minimal for Claude Opus 5.5 (22.8\% of occlusion-twin calls still produce reasoning tokens), low for GPT-6-Astra (0.3\% of twin calls).} \\
\end{tabular}}\\[3pt]
\setlength{\tabcolsep}{2pt}
\textbf{(b)}\ \fitto{0.95\textwidth}{\begin{tabular}{lccccc}
\toprule
\multicolumn{6}{l}{\textit{Hard occlusion set, closed models ($576$ prompts per condition); $d'$ by reasoning condition}} \\
\textbf{Model} & \textbf{Dir.\ acc.} & \textbf{low / Dir.\ / CoT} & \textbf{Dir.$-$low} & \textbf{CoT$-$low} & \textbf{CoT$-$Dir.} \\
\midrule
Claude Opus 5.5 & $85.2$ & $2.08^a$ / $2.08$ / $2.04$ & $+0.00$ {\scriptsize$[-0.09, +0.10]$} & $-0.04$ {\scriptsize$[-0.20, +0.12]$} & $-0.04$ {\scriptsize$[-0.19, +0.11]$} \\
GPT-6-Astra & $91.3$ & $2.77^b$ / $2.72$ / $2.98$ & $-0.05$ {\scriptsize$[-0.26, +0.15]$} & $+0.20$ {\scriptsize$[+0.02, +0.43]$} & $+0.26$ {\scriptsize$[+0.06, +0.46]$} \\
Qwen3.8-Max & $77.3$ & $1.76$ / $1.50$ / $1.60$ & $-0.26$ {\scriptsize$[-0.50, -0.02]$} & $-0.16$ {\scriptsize$[-0.40, +0.06]$} & $+0.10$ {\scriptsize$[-0.10, +0.28]$} \\
HY4 & $67.9$ & -- / $0.99$ / $1.08$ & -- & -- & $+0.09$ {\scriptsize$[-0.14, +0.32]$} \\
Gemini-4-Argon & $96.7$ & -- / $3.66$ / $3.70$ & -- & -- & $+0.04$ {\scriptsize$[-0.16, +0.36]$} \\
DeepSeek-V4.1-Flash & $60.1$ & $0.38$ {\scriptsize$[0.23, 0.52]$} / $0.56$ / $0.48$ & $+0.18$ {\scriptsize$[+0.02, +0.36]$} & $+0.11$ {\scriptsize$[-0.11, +0.32]$} & $-0.07$ {\scriptsize$[-0.26, +0.11]$} \\
\bottomrule
\end{tabular}}
\end{table*}

\paragraph{Accuracy and hard occlusion performance.}
We evaluate six frontier closed API models (\texttt{Claude Opus 5.5}, \texttt{GPT-6-Astra}, \texttt{Gemini-4-Argon}, \texttt{Qwen3.8-Max}, \texttt{HY4}, and \texttt{DeepSeek-V4.1-Flash}) at default \texttt{reasoning\_effort=high} and at each API's lowest accepted reasoning setting (\cref{tab:supp_closed}; API call details in \cref{sec:supp_closed}). Because closed \texttt{Direct} calls already emit hidden reasoning tokens on $51.7\%$--$100\%$ of queries (depending on model and set), comparing \texttt{CoT} against \texttt{Direct} contrasts visible-plus-hidden reasoning against hidden reasoning rather than against a reasoning-free readout. Lowest-reasoning \texttt{Direct} removes reasoning tokens entirely in \texttt{DeepSeek-V4.1-Flash} and \texttt{Qwen3.8-Max} and in all but $0.3\%$ of \texttt{GPT-6-Astra} twin calls, lowers the reasoning share in \texttt{Claude Opus 5.5} twin calls from $77.6\%$ to $22.8\%$, is rejected by \texttt{Gemini-4-Argon}, and is undocumented for \texttt{HY4}.

On the standard occlusion twins (\cref{tab:supp_closed}a), five of the six closed models sit at ceiling ($99.1\%$--$100.0\%$). To move them off ceiling, we evaluate them on the $576$-prompt hard occlusion set, which preserves the balanced twin design while adding thin overlap slivers, low contrast, distractor shapes, blur, noise, and smaller objects (\cref{tab:supp_closed}b; open models are near chance there, \texttt{Direct} $d'$ $-0.04$ to $0.62$). On the hard set, closed models drop off ceiling ($60.1\%$--$96.7\%$ \texttt{Direct} accuracy), yet visible \texttt{CoT} never significantly lowers $d'$ in any model (\texttt{GPT-6-Astra} rises by $+0.26$ $[+0.06, +0.46]$). Turning hidden reasoning on itself has mixed effects: it lowers \texttt{Qwen3.8-Max} from $1.76$ to $1.50$ ($-0.26$ $[-0.50, -0.02]$) and raises \texttt{DeepSeek-V4.1-Flash} from $0.38$ $[0.23, 0.52]$ to $0.56$ ($+0.18$ $[+0.02, +0.36]$). Auditing the closed models' visible \texttt{CoT} texts with two text-only judges shows $100\%$ text--answer agreement across all $13$ model--set cells with conclusions kept (\cref{tab:supp_e38}), at least $99\%$ ($100\%$ in $12$ of $13$ cells) after uniform conclusion removal (\cref{tab:supp_e45}), and no net Stage-1 deficit. On hard-set items where \texttt{Direct} is right, false written relations occur on $0.5\%$ (\texttt{Gemini-4-Argon}), $0.6\%$ (\texttt{GPT-6-Astra}), $4.1\%$ (\texttt{Claude Opus 5.5}), $7.9\%$ (\texttt{Qwen3.8-Max}), $14.9\%$ (\texttt{HY4}), and $19.1\%$ (\texttt{DeepSeek-V4.1-Flash}), and the resulting \texttt{D+C$-$} losses are offset by \texttt{D$-$C$+$} rescues.

\paragraph{Regenerated rationales and image-blind controls.}
To test whether closed models read the written relation from the image during generation, we regenerate \texttt{CoT} replies under the real image, a uniform gray image, and the identity-swapped twin ($150$ prompts per set on the hard set and cue set; the sampled hard items are $1.2$--$4.1$ points easier than the full set; \cref{tab:supp_e50}). Under the twin, both the written relation and the final answer track the image: the regenerated twin image effect is $+21.3$ to $+95.3$ points on the hard set and $+81.3$ to $+100.0$ on the cue set, matching \texttt{Direct} on the hard set (\texttt{CoT}$-$\texttt{Direct} $-3.3$ to $+6.0$, every CI covering or touching $0$), with $100.0\%$ agreement between the written relation and the answer whether the conclusion is kept or removed. Under a uniform gray image, five models state that the shapes are absent in at least $99.3\%$ of replies and answer ``no''; \texttt{Gemini-4-Argon} instead describes faint shapes and a depth order in $9.3\%$ (hard set) and $17.3\%$ (cue set) of replies and answers by it, naming the true front shape at chance ($19/40$). Because the rationale is regenerated under each image here, this test establishes that closed models write from the image; it does not yet test whether they can override a fixed written rationale.

\begin{table*}[t]
\centering
\caption{\textbf{Closed models with their own rationale held fixed.} Image effect $P(\text{answer}=\text{gold}_v\mid v)-P(\text{answer}=\text{gold}_v\mid v_\mathrm{cf})$ (points; pair-cluster bootstrap CI for the primary \emph{removed} variant). Regen.: rationale regenerated under each image (\cref{tab:supp_e50}). Fixed: own stored reply under $v$ with the answer line removed (\emph{kept}), the conclusion removed (\emph{removed}, uniform rule of \cref{sec:supp_engine}), or also mid-sentence conclusions removed (\emph{strict}, \cref{sec:supp_engine}). Direct: letter response at lowest reasoning (reasoning high in parentheses). Last column: removed-variant effect by the truth of the last written relation for $v$ (two-judge consensus; $n$). \texttt{Gemini-4-Argon} (evaluated on the primary variant, as kept and strict returned endpoint errors), \texttt{DeepSeek-V4.1-Flash}, and \texttt{HY4} were completed in a later run.}
\label{tab:supp_e61}
\scriptsize
\setlength{\tabcolsep}{3pt}
\fitto{\textwidth}{\begin{tabular}{llcccccc}
\toprule
Model & Set & Regen. & \multicolumn{3}{c}{Fixed own rationale} & Direct & True / false relation \\
\cmidrule(lr){4-6}
 & & & kept & removed & strict & & (removed) \\
\midrule
Claude Opus 5.5 & Hard occl. & $+72.7$ & $+75.3$ & $+74.0$ {\scriptsize[+62.5, +84.8]} & $+74.0$ & $+74.0$ ($+76.0$) & $+90.2$ (133) / $-52.9$ (17) \\
GPT-6-Astra & Hard occl. & $+90.0$ & $+90.0$ & $+90.0$ {\scriptsize[+84.4, +94.8]} & $+92.0$ & $+90.0$ ($+88.7$) & $+91.8$ (146) / $+25.0$ (4) \\
Gemini-4-Argon & Hard occl. & $+95.3$ & -- & $+92.0$ {\scriptsize[+86.8, +96.4]} & -- & $+93.3$ ($+93.3$) & $+92.0$ (150) / -- (0) \\
Qwen3.8-Max & Hard occl. & $+58.0$ & $+0.7$ & $+1.3$ {\scriptsize[$-$2.0, +4.7]} & $+18.7$ & $+62.0$ ($+52.0$) & $+1.7$ (120) / $+0.0$ (30) \\
DeepSeek-V4.1-Flash & Hard occl. & $+21.3$ & $+0.7$ & $+2.7$ {\scriptsize[$-$0.7, +6.3]} & $+1.6$ & $+14.0$ ($+21.3$) & $+5.3$ (94) / $-1.9$ (54) \\
HY4 & Hard occl. & $+44.0$ & $+2.0$ & $+4.1$ {\scriptsize[$-$1.4, +9.5]} & $+10.9$ & $+18.0$ ($+39.3$) & $+6.4$ (109) / $-2.6$ (38) \\
\midrule
Claude Opus 5.5 & Cue set & $+100.0$ & $+100.0$ & $+100.0$ {\scriptsize[+100.0, +100.0]} & $+100.0$ & $+100.0$ & $+100.0$ (150) / -- (0) \\
GPT-6-Astra & Cue set & $+100.0$ & $+100.0$ & $+100.0$ {\scriptsize[+100.0, +100.0]} & $+100.0$ & $+100.0$ & $+100.0$ (150) / -- (0) \\
Gemini-4-Argon & Cue set & $+100.0$ & -- & $+100.0$ {\scriptsize[+100.0, +100.0]} & -- & $+100.0$ & $+100.0$ (150) / -- (0) \\
Qwen3.8-Max & Cue set & $+99.3$ & $+0.7$ & $+4.7$ {\scriptsize[+1.8, +8.3]} & $+21.3$ & $+99.3$ & $+4.7$ (150) / -- (0) \\
DeepSeek-V4.1-Flash & Cue set & $+81.3$ & $+4.7$ & $+12.8$ {\scriptsize[+7.5, +18.5]} & $+24.3$ & $+76.0$ & $+14.3$ (133) / $+0.0$ (16) \\
HY4 & Cue set & $+100.0$ & $+4.7$ & $+10.7$ {\scriptsize[+5.7, +16.0]} & $+24.7$ & $+76.7$ & $+10.7$ (150) / -- (0) \\
\bottomrule
\end{tabular}}
\end{table*}

\paragraph{Fixed-rationale twin swaps and answer-step re-grounding.}
\cref{tab:supp_e61} applies the fixed-rationale protocol to all six closed models (pre-specified; same $150+150$ prompts). Each model's stored reply under $v$ from \cref{tab:supp_e50}, truncated under three rules (\emph{kept}, primary \emph{removed}, and \emph{strict}), is supplied as its prior assistant turn, and a second user turn asks for the option letter only under $v$ or under the swapped twin $v_\mathrm{cf}$ at the lowest reasoning setting (analogous to the two-turn \texttt{CoT+turn} and \texttt{CoT+reimage+restate} controls of \cref{tab:behavioral_main}, which recover at most $1.5$ points in open models). Three of the six closed models re-ground at the answer step, following the swapped image as strongly as when regenerating the rationale or answering directly. On the hard set (primary \emph{removed} variant), \texttt{Claude Opus 5.5} reaches $+74.0$ $[+62.5, +84.8]$ (vs.\ $+72.7$ regenerated, explicitly noting the correction on $8\%$ of hard-set and $33\%$ of cue-set twin replies), \texttt{GPT-6-Astra} reaches $+90.0$ $[+84.4, +94.8]$ (vs.\ $+90.0$), and \texttt{Gemini-4-Argon} reaches $+92.0$ $[+86.8, +96.4]$ (vs.\ $+95.3$; kept and strict variants returned endpoint errors), while all three reach $+100.0$ on the cue set. Splitting by the truth of the written relation under $v$ (\cref{tab:supp_e61}, last column) adds an important nuance: on the $17$ hard-set items where \texttt{Claude Opus 5.5}'s own written relation is false for $v$, its image effect is $-52.9$ (vs.\ $+90.2$ on $133$ true relations), so its re-grounding reflects noticing a changed image under $v_\mathrm{cf}$ rather than repairing an already-false rationale under $v$.

The other three closed models lock onto their own fixed text just like the open models, satisfying the pre-specified lock-in threshold (below one-third of the regenerated effect). On the hard set and cue set under the primary \emph{removed} variant, \texttt{Qwen3.8-Max} yields $+1.3$ $[-2.0, +4.7]$ and $+4.7$ $[+1.8, +8.3]$, \texttt{DeepSeek-V4.1-Flash} yields $+2.7$ $[-0.7, +6.3]$ and $+12.8$ $[+7.5, +18.5]$, and \texttt{HY4} yields $+4.1$ $[-1.4, +9.5]$ and $+10.7$ $[+5.7, +16.0]$. Removing mid-sentence conclusions as well (\emph{strict}) raises all three models' effects on the cue set (to $+21.3$, $+24.3$, and $+24.7$). On the hard set, strict truncation raises \texttt{Qwen3.8-Max} ($+18.7$) and \texttt{HY4} ($+10.9$) while leaving \texttt{DeepSeek-V4.1-Flash} at $+1.6$, and all three remain well below their regenerated and \texttt{Direct} baselines.

An important caveat accompanies this split: \emph{re-grounding is confounded with answer-turn reasoning}. The three models that lock in (\texttt{Qwen3.8-Max}, \texttt{DeepSeek-V4.1-Flash}, and \texttt{HY4}) are precisely the three queried with reasoning turned off (\texttt{HY4} accepted reasoning effort none yet still returned reasoning text on $80$ of $1{,}752$ calls; restricted to the $130$ hard-set and $136$ cue-set pairs without reasoning in either call, its effect is $+0.0$ $[-4.3, +4.4]$ and $+8.8$ $[+3.7, +14.3]$). Conversely, among the three re-grounding models, \texttt{Claude Opus 5.5} produces hidden reasoning tokens on every twin call, \texttt{Gemini-4-Argon} does not report reasoning-token counts, and \texttt{GPT-6-Astra} uses reasoning tokens on $145$ of $150$ hard-set twin calls (image effect $+93.1$ $[+88.4, +96.9]$ on those pairs) while giving identical answers under both images on the $5$ hard-set pairs without reasoning tokens in either call. On the cue set, however, $59$ \texttt{GPT-6-Astra} pairs ($53$ scenes) use no reasoning tokens in either call yet still achieve $+100.0$ $[+100.0, +100.0]$ (noting that zero-reasoning subsets are self-selected by the model). Text lock-in is therefore avoidable at the answer step, motivating answer-step re-grounding as a concrete target for open VLMs.

\section{Towards Answer-Step Re-grounding}
\label{sec:regrounding}

Because open VLMs retain the true relation in mid-layer representations (\cref{sec:error_onset}) and frontier models show that text lock-in is avoidable (\cref{sec:frontier}), we test whether label-free inference-time interventions can break lock-in and whether answer-step re-grounding can be learned directly.

\subsection{Label-free inference-time interventions}
\label{sec:label_free}

No label-free inference-time intervention we tested beats \texttt{Direct} on average on its evaluated sets (or out of sample where tested; \cref{tab:interventions_summary}). Full per-model tables and controls are in \cref{sec:supp_interventions}.

\begin{table*}[t]
\centering
\caption{\textbf{Summary of label-free inference-time interventions against \texttt{Direct}.} All entries are compiled verbatim from the full tables in \cref{sec:supp_interventions} (\cref{tab:supp_gate,tab:supp_state,tab:supp_e58a,tab:supp_e55}) and \cref{sec:supp_heldout}. Brackets: $95\%$ cluster-bootstrap CIs; --: not tested out of sample.}
\label{tab:interventions_summary}
\scriptsize
\setlength{\tabcolsep}{3pt}
\fitto{\textwidth}{\begin{tabular}{@{}p{0.20\textwidth}p{0.24\textwidth}p{0.22\textwidth}p{0.29\textwidth}@{}}
\toprule
\textbf{Intervention} & \textbf{In-sample / evaluated $\Delta$ vs.\ \texttt{Direct}} & \textbf{Out-of-sample $\Delta$ vs.\ \texttt{Direct}} & \textbf{Primary failure mode} \\
\midrule
Commit-sentence gate (\cref{tab:supp_gate}; 4 models, 3 synthetic sets) & $-6.4$ to $+0.6$ pts (CI $<0$ in $5/12$ cells; $+9.8$ to $+21.8$ over \texttt{CoT} in $6$ loss cells) & -- & Loses $59\%$--$96\%$ of \texttt{CoT} rescues; restores height shortcut ($-2.9$ to $-5.0$ vs.\ \texttt{CoT} where \texttt{CoT} gains); $2.5$--$5.0\times$ tokens, $8$--$25$ readouts \\
Structured JSON state (\cref{tab:supp_state}; 7 models, twins \& \texttt{Syn.\ Depth}) & $-13.2$ to $+2.4$ pts (\texttt{Syn.\ Depth}); $-1.14$ to $+0.35$ $d'$ (twins) & -- & ``front'' field right on $49.6\%$--$72.6\%$ of twins (upper front $19.9\%$--$45.9\%$); encodes height heuristic, not occlusion \\
Body-relation readout BR$_C$ (\cref{tab:supp_e58a}; 7-model mean) & $+4.0$ $[+1.5, +6.3]$ pts (cue set); other synthetic sets: $+3.9$ (twins), $-13.0$ (\texttt{Syn.\ Depth} $431$) & $-5.4$ $[-6.7, -4.1]$ pts (natural depth: \texttt{CV-Bench} 3D $+$ \texttt{BLINK}; $+0.1$ vs.\ \texttt{CoT}) & Removes Stage-3 conclusion mis-mapping; leaves Stage-1 false body relations unchanged \\
Perception-first prompts P1 / U1 (\cref{tab:supp_e55}; 7-model mean) & $-4.2$ $[-5.6, -2.8]$ (P1) / $-4.5$ $[-6.3, -2.6]$ (U1) on cue set & $-2.1$ $[-3.0, -1.2]$ (P1) / $-4.4$ $[-5.7, -3.2]$ (U1) on natural depth & P1 induces empty turns; U1 writes false overlap claims (\texttt{Molmo-7B-D} $-27.2$) and does not beat \texttt{Direct} on natural depth \\
\texttt{Direct}-margin gating $\tau$ (\cref{sec:supp_heldout,sec:supp_interventions}; 7 models) & $+1.1$ to $+2.5$ pts (main pool, $4/7$ models) & $-0.8$ to $+2.1$ pts (held-out $767$; every CI includes $0$) & Cross-validated margin threshold does not transfer significantly to held-out tasks \\
\bottomrule
\end{tabular}}
\end{table*}

\paragraph{Commit-sentence gate.}
Replacing the oracle claim edit of \cref{sec:answer_step} with a label-free criterion, the commit-sentence gate (\cref{tab:supp_gate}) tracks the turn-closed forced readout $P_j(d)$ of the model's own \texttt{Direct} answer $d$ after each sentence $s_j$ (no gold labels enter). Whenever $s_j$ drops $P(d)$ from $\ge 0.5$ to $< 0.5$, the gate resamples $s_j$ up to $K=4$ times at $T=0.7$, keeping the first sample that preserves $P(d)\ge 0.5$ or dropping $s_j$ and re-gating the continuation if all $K$ flip (at most $8$ gate events per item, reached by at most $7.7\%$ of items). Across four models and three synthetic sets, the gate recovers $+9.8$ to $+21.8$ points over \texttt{CoT} in the six cells where \texttt{CoT} loses significantly, outperforming a rule-parsed text gate by $4.6$--$10.9$ points. However, it never significantly beats \texttt{Direct} ($-6.4$ to $+0.6$ points, CI below zero in $5/12$ cells), discards $59\%$--$96\%$ of \texttt{CoT}'s rescues, costs $2.5$--$5.0\times$ \texttt{CoT}'s tokens plus $8$--$25$ forced readouts per item, and where \texttt{CoT} gains on the cue set and twins it drops $2.9$--$5.0$ points below \texttt{CoT} by restoring \texttt{Direct}'s lower-is-closer prior rather than occlusion. Count-matched controls confirm that the gate acts as \texttt{Direct} with a rationale attached (\cref{sec:supp_gate}, which also reports post-hoc answer-first explanations).

\paragraph{Structured spatial state.}
To test whether replacing free prose with a structured schema avoids transcription errors, we prompt the assistant to complete a JSON object naming both objects, their bounding boxes, which object is occluded, and which is in front ($\le 300$ tokens) before the same-turn readout (\cref{tab:supp_state}). The schema parses on $97\%$--$100\%$ of items and recovers $+13.2$ to $+19.6$ points of the free-\texttt{CoT} loss on \texttt{Synthetic Depth} in five models, but stays below \texttt{Direct} in \texttt{Qwen2.5-VL-3B} ($-13.2$) and \texttt{InternVL3-8B} ($-7.2$). On the occlusion twins, it removes the \texttt{CoT} loss in \texttt{Molmo-7B-D} ($+1.24$ $d'$ over \texttt{CoT}) but lowers $d'$ below free \texttt{CoT} in \texttt{Qwen3-VL-8B} ($-0.62$) and \texttt{Phi-3.5-Vision} ($-0.38$). The ``front'' field is correct on $49.6\%$--$72.6\%$ of twin images and names the upper object as front in $19.9\%$--$45.9\%$ (against a $50/50$ design), showing that the structured state largely encodes the lower-is-closer heuristic rather than occlusion perception.

\paragraph{Body-relation readout and additional label-free controls.}
Because Stage-3 conclusion mis-mapping costs accuracy on the cue set, we tested a pre-specified label-free readout that answers deterministically from the last body relation (two-judge consensus on conclusion-removed text, falling back to \texttt{CoT} in BR$_C$ or \texttt{Direct} in BR$_D$ when no relation is stated; \cref{tab:supp_e58a}). In sample on the cue set, BR$_C$ beats \texttt{Direct} by $+4.0$ $[+1.5, +6.3]$ points on the seven-model mean, driven by the height-reliant \texttt{InternVL} models ($+10.0$, $+8.1$) and \texttt{Molmo-7B-D} ($+2.2$). Across the remaining synthetic sets, it gives $+3.9$ $[+0.4, +7.3]$ on the twins and $-13.0$ $[-15.0, -11.0]$ on \texttt{Synthetic Depth} $431$. On the pre-specified out-of-sample primary (pooled natural depth: \texttt{CV-Bench} 3D Depth plus strictly held-out \texttt{BLINK} Relative Depth), however, BR$_C-$\texttt{Direct} falls to $-5.4$ $[-6.7, -4.1]$, changing accuracy by $+0.1$ $[-0.2, +0.4]$ over \texttt{CoT} because natural depth rationales carry little Stage-3 loss. Additional label-free controls in \cref{sec:supp_interventions} likewise do not surpass \texttt{Direct} on average: pre-specified perception-first prompts in the assistant turn (P1) or user turn (U1; \cref{tab:supp_e55}), order-swap probability averaging (gains held-out \texttt{Direct} in $3/7$ models), and frozen \texttt{Direct}-margin gating (\cref{tab:interventions_summary}).

\subsection{Learning answer-step re-grounding (pilot)}
\label{sec:supp_regrounding_pilot}

\begin{table*}[t]
\centering
\caption{\textbf{Pre-specified LoRA pilot of the re-grounding recipe} (\texttt{Qwen3-VL-8B}). Image effects in points (COCO: probability scale; \texttt{CV-Bench}: answer scale, as pre-specified); accuracies in \%; cluster-bootstrap $95\%$ CIs ($2{,}000$ draws shared across models, so differences are paired). Base: the untuned model re-run with the same code (its fixed readouts reproduce the stored ones on $100\%$ of items; its regenerated main-pool \texttt{CoT}$-$\texttt{Direct} is $-3.0$ vs.\ $-3.5$ in \cref{tab:behavioral_main}).}
\label{tab:supp_e63_pilot}
\scriptsize
\setlength{\tabcolsep}{2pt}
\fitto{\textwidth}{\begin{tabular}{llrrrrr}
\toprule
Endpoint & Set & Base & MATCH & CF & CF$-$base & CF$-$MATCH \\
\midrule
H1: image effect, base $R_\mathrm{cut}$ fixed & COCO twins & $2.9$ & $51.7$ & $58.9$ & $+55.9$ {\tiny[51.5, 59.9]} & $+7.2$ {\tiny[5.0, 9.5]} \\
 & mirrored CV-Bench & $1.3$ & $90.7$ & $93.4$ & $+92.1$ {\tiny[86.8, 96.7]} & $+2.6$ {\tiny[0.0, 6.0]} \\
 & rendered twins / cue set & $\le 10$ & ${\approx}100$ & ${\approx}100$ & & \\
H2: \texttt{Direct} accuracy & natural (pooled, $4{,}015$) & $81.3$ & $84.4$ & $84.2$ & $+2.9$ {\tiny[2.0, 3.9]} & $-0.2$ {\tiny[$-$0.9, 0.4]} \\
 & held-out pool ($767$) & $70.7$ & $70.0$ & $68.8$ & $-1.8$ {\tiny[$-$4.4, 0.5]} & $-1.2$ {\tiny[$-$3.0, 0.4]} \\
H3: own \texttt{CoT}$-$\texttt{Direct} & main pool & $-3.0$ & $-0.1$ & $+0.1$ & $+3.2$ {\tiny[2.0, 4.4]} & $+0.2$ {\tiny[$-$0.3, 0.7]} \\
 & cue set & $-1.5$ & $-0.1$ & $0.0$ & $+1.5$ {\tiny[$-$1.8, 4.6]} & $+0.1$ {\tiny[0.0, 0.2]} \\
H4: gray image, answer = base & four sets pooled ($2{,}402$) & $97.6$ & $68.7$ & $47.3$ & $-50.3$ {\tiny[$-$52.2, $-$48.5]} & $-21.4$ {\tiny[$-$23.1, $-$19.8]} \\
\quad $R_\mathrm{cut}$'s written relation & COCO twins ($326$) & $96.3$ & $83.7$ & $65.6$ & $-30.7$ {\tiny[$-$36.3, $-$24.6]} & $-18.1$ {\tiny[$-$22.9, $-$12.8]} \\
 & mirrored CV-Bench ($147$) & $100.0$ & $95.9$ & $57.1$ & $-42.9$ {\tiny[$-$50.7, $-$35.1]} & $-38.8$ {\tiny[$-$46.6, $-$31.3]} \\
 & cue set / twins (rendered) & $97.7$ / $97.4$ & $62.4$ / $67.3$ & $43.0$ / $44.2$ & & \\
Base-correct \texttt{D+C$+$} kept (base $R_\mathrm{cut}$) & held-out pool ($469$) & $98.9$ & $96.2$ & $92.8$ & $-6.2$ {\tiny[$-$8.7, $-$3.9]} & \\
\bottomrule
\end{tabular}}
\end{table*}

Since inference-time interventions do not supply answer-step image checking, we test whether re-grounding is learnable in a pre-specified single-model pilot (H1--H4) on \texttt{Qwen3-VL-8B} (\cref{tab:supp_e63_pilot}). We LoRA-fine-tune the model (rank $16$, one epoch, loss on the answer letter only) on its own conclusion-removed rationales for newly rendered scenes ($1{,}512$ occlusion-twin pairs and $1{,}008$ left/right mirror pairs, with no overlap with any evaluation image). Both training arms use $5{,}040$ examples with labels always determined by the shown image: the counterfactual arm (CF) pairs one rationale per scene pair ($2{,}520$ rationales) with both images so that half of training examples contradict the text, whereas the matched control (MATCH) pairs each of the $5{,}040$ rationales with its own image (which still contradicts the image label on $13.3\%$ of examples where the base model's rationale was wrong). Because the \texttt{CV-Bench} left/right template and mirroring appear in rendered training, mirrored \texttt{CV-Bench} tests transfer to the natural-image domain rather than a new question template.

Fine-tuning on rendered scenes alone restores answer-step image use on natural photographs (\cref{tab:supp_e63_pilot}). With the base model's $R_\mathrm{cut}$ held fixed (H1), the twin image effect rises from $2.9$ to $51.7$ (MATCH) and $58.9$ (CF) on COCO occlusion twins and from $1.3$ to $90.7$ and $93.4$ on mirrored \texttt{CV-Bench}. On the pooled $4{,}015$-item natural set (H2), \texttt{Direct} accuracy rises from $81.3\%$ to $84.4\%$ (MATCH, $+3.1$) and $84.2\%$ (CF, $+2.9$ $[+2.0, +3.9]$). On the held-out pool, \texttt{Direct} accuracy reaches $70.0\%$ (MATCH) and $68.8\%$ (CF, $-1.8$ $[-4.4, +0.5]$). On the main pool (H3, where the pilot's base re-run has \texttt{CoT}$-$\texttt{Direct} $=-3.0$ vs.\ $-3.5$ in \cref{tab:behavioral_main}), the \texttt{CoT} loss disappears ($-0.1$ MATCH, $+0.1$ CF). At the same time, retention of base-correct \texttt{D+C$+$} held-out answers ($n=469$) moves from $98.9\%$ to $96.2\%$ (MATCH) and $92.8\%$ (CF, $-6.2$ $[-8.7, -3.9]$). Counterfactual pairing adds little re-grounding over MATCH ($+7.2$ $[+5.0, +9.5]$ on COCO, $+2.6$ $[+0.0, +6.0]$ on \texttt{CV-Bench}; H1 partial) while altering how the text is used when the image is uninformative. Under a uniform gray image (H4), MATCH still follows the base rationale's written relation on $83.7\%$ of COCO twins and $95.9\%$ of mirrored \texttt{CV-Bench} items (base: $96.3\%$ and $100.0\%$), whereas CF drops to $65.6\%$ and $57.1\%$; on the rendered training domain both arms fall to $43.0\%$--$67.3\%$, so the pooled H4 endpoint is not met ($68.7\%$ MATCH, $47.3\%$ CF). On natural photographs, MATCH therefore consults the image when it is informative and follows the rationale when it is not, whereas counterfactual pairing mainly teaches the model to discount the text. This single-model pilot suggests that answer-step re-grounding is learnable from fixed rationales with image-determined labels on rendered scenes without requiring contradicting twin pairs.

\section{Discussion}
\label{sec:discussion}

\paragraph{Why benchmarks miss the failure.}
Shortcut relief and lost valid cues offset each other on average, so standard aggregate scores on natural benchmarks often show a small or null net \texttt{CoT}$-$\texttt{Direct} gap despite heavy item-level answer churn (\cref{tab:supp_heldout,tab:supp_churn}b). Controlled cue conflicts and fixed-rationale counterfactual image swaps expose that the answer has moved from the image to the written text, showing that rationales of open VLMs are not faithful records of visual perception.

\paragraph{Implications for open VLMs.}
Open models already hold the required perceptual signal: the true depth order remains decodable at the written relation token (\cref{sec:error_onset,tab:supp_probe_align}), and rationale-body checkers use the image to reject an appended wrong conclusion (\cref{sec:answer_step,fig:swaps,tab:supp_e59}). What open VLMs lack is an answer step that re-consults the image when it conflicts with their own generated text. Prompting and inference-time heuristics do not supply this behavior: across re-shown images or questions, perception-first instructions, structured JSON state, body-relation readouts, and commit-sentence gating, none beats \texttt{Direct} on average on its evaluated sets (or out of sample where tested; \cref{sec:regrounding,tab:interventions_summary}). By contrast, an unconditional oracle edit replacing the first claim with the true relation beats \texttt{Direct} by $14.3$--$34.9$ points in five of six core models tested (\texttt{Molmo-7B-D} $-12.3$; \cref{sec:answer_step,tab:supp_claim_uncond}). Three of six frontier closed models show that answer-step re-grounding is achievable, though in those APIs it is confounded with answer-turn reasoning (\cref{sec:frontier}), and our pre-specified single-model LoRA pilot suggests that fine-tuning on fixed rationales with image-determined labels (MATCH) restores answer-step re-grounding on natural photographs from rendered scenes alone (\cref{sec:supp_regrounding_pilot}). Combining answer-step re-grounding with reinforcement learning rewards for visually consistent intermediate claims~\citep{yu2025perceptionr1,chen2026rememberr1} is a natural next step.

\section{Conclusion}
\label{sec:conclusion}
We showed that short chain-of-thought turns spatial VQA from a perception problem into a transcription problem. Writing re-weights visual cues and produces false spatial relations, the forced readout then locks onto the written text instead of the image across both rendered scenes and natural photographs, and concluding sentences can mis-map even true relations on converse questions. Consequently, written-relation accuracy minus \texttt{Direct} accuracy accounts for when CoT helps or hurts. Model scale alone does not remove this bottleneck, and label-free inference-time interventions do not beat \texttt{Direct} on average out of sample. Answer-step re-grounding, observed in three of six frontier closed models and demonstrated in our single-model LoRA pilot on natural photographs, offers a direct path to breaking text lock-in.

{
    \small
    \bibliographystyle{ieeenat_fullname}
    \bibliography{references}
}

\clearpage
\appendix
\renewcommand{\theHtable}{S\arabic{table}}
\renewcommand{\theHfigure}{S\arabic{figure}}
\renewcommand{\dbltopfraction}{0.95}
\renewcommand{\topfraction}{0.95}
\renewcommand{\bottomfraction}{0.5}
\renewcommand{\textfraction}{0.05}
\renewcommand{\dblfloatpagefraction}{0.8}
\renewcommand{\floatpagefraction}{0.8}
\setcounter{dbltopnumber}{4}
\setcounter{topnumber}{4}
\setcounter{totalnumber}{6}
\setlength{\textfloatsep}{8pt plus 2pt minus 2pt}
\setlength{\dbltextfloatsep}{8pt plus 2pt minus 2pt}
\setlength{\floatsep}{6pt plus 2pt minus 2pt}
\setlength{\dblfloatsep}{6pt plus 2pt minus 2pt}
\setlength{\intextsep}{6pt plus 2pt minus 2pt}
\captionsetup{font=footnotesize,skip=3pt}
\makeatletter\setlength{\@fptop}{0pt}\setlength{\@dblfptop}{0pt}\makeatother
\newpage
\twocolumn[
    \centering
    \Large
    \textbf{\thetitle}\\
    \vspace{0.5em}Appendix \\
    \vspace{1.0em}
]
\pdfbookmark[0]{Appendix}{appendix}
\renewcommand{\thefigure}{S\arabic{figure}}
\renewcommand{\thetable}{S\arabic{table}}
\setcounter{figure}{0}
\setcounter{table}{0}

\noindent\textbf{Reading guide.} \cref{sec:supp_impl} details the models, prompts, engines, and statistical tests, along with truncation-rule definitions (\cref{sec:supp_engine}), the occlusion rule and heuristic baselines (\cref{sec:supp_competence,sec:supp_cues}), sentence judges (\cref{sec:supp_judge}), frozen held-out protocol details (\cref{sec:supp_heldout}), and closed-model API settings supporting \cref{sec:frontier} (\cref{sec:supp_closed}). \cref{sec:supp_a1,sec:supp_a2,sec:supp_a3,sec:supp_a4} supplement Analyses 1--4 of the main paper. Specifically, we cover format controls and answer churn (\cref{sec:supp_a1}), commit-sentence checks, the full \texttt{Qwen3-VL-32B} pipeline, uniform-attenuation tests, probes, and steering (\cref{sec:supp_a2}). We then report phrasing-form claim edits, attention-edge knockouts, body-vs.-image checking, and regenerated closed-model controls (\cref{sec:supp_a3}), followed by full two-channel, larger/reasoning-tuned cohort, closed-model, and multi-image benchmark tables (\cref{sec:supp_a4}). \cref{sec:supp_interventions} supplements \cref{sec:regrounding} with full per-model tables for the label-free interventions, gate controls, answer-first explanations, perception-first prompts, order swaps, and margin gating, and \cref{sec:supp_repro} provides reproducibility notes, verbatim prompts, and model checkpoints. \cref{tab:supp_evidence} maps each claim of the main paper to its supporting evidence, and \cref{tab:supp_tab2_ci} reports the $95\%$ CIs for every cell of main \cref{tab:profile}. Additional per-model breakdowns and per-item outputs are available in the released repository (\href{https://github.com/MING-ZCH/seeing-right-writing-wrong}{github.com/MING-ZCH/seeing-right-writing-wrong}).

\smallskip
\noindent\textbf{Appendix summary tables.} \cref{tab:supp_evidence,tab:supp_tasks,tab:supp_e59,tab:supp_e58b,tab:supp_tab2_ci} collect the evidence map, main-pool task breakdown, conclusion-removed attention knockouts, \texttt{Synthetic Depth} leverage checks, and per-cell CIs for \cref{tab:profile}; additional supporting tables (\cref{tab:supp_multiplicity} onward) follow in their respective sections.
\smallskip

\begin{table*}[t]
\centering
\caption{\textbf{Evidence map.} Each claim of the main paper with the experiments, figures, and tables that support it and its main caveat.}
\label{tab:supp_evidence}
\fontsize{6}{7}\selectfont
\setlength{\tabcolsep}{2.5pt}
\renewcommand{\arraystretch}{0.84}
\fitto{\textwidth}{\begin{tabular}{p{0.26\textwidth}p{0.22\textwidth}p{0.12\textwidth}p{0.13\textwidth}p{0.21\textwidth}}
\toprule
Claim & Evidence & Main & Appendix & Main caveat \\
\midrule
\textbf{Not format, not lost input.} Turn-closed filler of length $|R|$ stays within $-2.3$ to $+0.7$ points of \texttt{Direct} in all eight open models; re-showing the image or question adds at most $0.8$ (both together at most $1.5$) & Same-turn vs.\ turn-closed filler, long filler, re-image, restate, both & \cref{sec:behavioral}, \cref{tab:behavioral_main} & \cref{sec:supp_a1}, \cref{tab:supp_prior,tab:supp_tasks} & Same-turn filler shifts the option-letter prior (up to $-9.3$); \texttt{Molmo-7B-D} re-image only on the cue set (HuggingFace readout) \\
\textbf{What \texttt{Direct} relies on.} Two of seven models read occlusion on the cue set; four follow lower-is-closer on $85\%$--$96\%$ of items; in \texttt{Synthetic Depth} the size rule is right on $98.6\%$ & Occlusion rule on the cue-factorial set; rule baselines; \texttt{Direct} agreement with each rule & \cref{sec:behavioral}, \cref{tab:profile} & \cref{sec:supp_competence}, \cref{tab:supp_heuristics,tab:supp_glm} & Rule explicit but not blind; rendered shapes; both occlusion readers near chance on the hard set \\
\textbf{Stage 2: the readout locks onto the written relation.} Readout gives in at one sentence, relational in $69.3\%$--$88.5\%$ of committed trajectories (all seven); answer agrees with the last written relation in $77.2\%$--$99.6\%$ (true) and $75.5\%$--$94.8\%$ (false); a true first claim beats \texttt{Direct} by $14.3$--$34.9$ points in five of six; a twin image under the model's own conclusion-removed rationale moves the answer by $2.0$--$8.9$ points (full rationale: $\le0.4$ in five of seven), and likewise on natural photographs (mirrored \texttt{CV-Bench}: $69.6\to3.4$; COCO occlusion twins: $20.5\to3.0$); copiers and checkers both follow text; on $R_\mathrm{cut}$ removing the image from all text flips $2.1\%$--$9.0\%$ of answers ($26.6\%$--$53.9\%$ under \texttt{Direct}), blocking the last relation $10\%$--$24\%$ & Commit trajectories with two nulls (uniform and position-matched); attention-edge knockouts; all-item relation-by-answer tables; unconditional claim edits; own-rationale twin image swap on rendered and natural images; conclusion swaps on original, gray, mirrored images & \cref{sec:error_onset,sec:answer_step}, \cref{fig:error_onset,fig:swaps}, \cref{tab:supp_e60,tab:supp_e62,tab:supp_claim_uncond} & \cref{sec:supp_engine,sec:supp_a2,sec:supp_a3}, \cref{tab:supp_x2_trunc,tab:supp_e57,tab:supp_recheck,tab:supp_e59} & $j^\ast$ is selected by the readout collapse; \texttt{Molmo-7B-D}'s true-claim edit lowers accuracy ($-12.3$); the two \texttt{InternVL} models move with the twin image ($+6.8$, $+8.5$); claim edit not run for \texttt{InternVL3.5-8B}; three of six closed models (\texttt{Claude Opus 5.5}, \texttt{GPT-6-Astra}, \texttt{Gemini-4-Argon}) re-read the image under a fixed rationale (\cref{tab:supp_e61}) \\
\textbf{Stages 1 and 3: the written relation is often wrong.} Of $3{,}014$ \texttt{CoT} losses, $80\%$ end in a false written statement ($57\%$ false relation, $24\%$ true relation with a mis-mapped conclusion, mostly converse questions), under $1\%$ depart from the text at the readout; commit sentences contradict the image in $184/209 = 88.0\%$ vs.\ $19/203 = 9.4\%$ (\texttt{Synthetic Depth} $431$, seven models) and $252/384 = 65.6\%$ vs.\ $6.6\%$ (cue set, two models); a probe at the relation word decodes the true relation ($0.88$--$0.93$, three models); writing attenuates all cues and discounts height beyond a uniform-attenuation null; a question-blind description also costs accuracy & Loss decomposition; attenuation null; two-judge audits with matched comparators; human check; within-claim probe with gray, text-only, and twin-swap controls; description arms & \cref{sec:error_onset,sec:answer_step}, \cref{fig:teaser}b, \cref{tab:supp_written_ame,tab:supp_e54a,tab:cohorts_summary} & \cref{sec:supp_judge,sec:supp_a2}, \cref{tab:supp_syn431,tab:supp_judge,tab:supp_e57,tab:supp_probe_align,tab:supp_e54b,tab:supp_e48} & Body relations of \texttt{Molmo-7B-D} and \texttt{Qwen3-VL-8B} are as accurate as \texttt{Direct} on the cue set once the conclusion is removed; two reasoning-tuned, height-leaning models that pass the occlusion rule write relations more accurate than \texttt{Direct} (\cref{tab:supp_e48}); LLM judges; a $30$-sentence human check agrees with their consensus ($\kappa = 0.91$); in \texttt{Synthetic Depth} \texttt{Direct} may rest on size; per-model cue-set rates differ ($91.1\%$, $37.4\%$); decodability is not use (steering gives no gain) \\
\textbf{Net effect follows the written relation.} $\texttt{CoT}-\texttt{Direct}=x+A+$ unparsed; $r=0.93$ over $21$ model--dataset points; $A$ down to $-15.6$ with conclusions removed, near $0$ on natural and closed points; shortcut models gain on the cue set ($+0.6$ to $+4.8$), \texttt{Synthetic Depth} losses up to $25.8$ points; on natural depth questions $y=x$ predicts \texttt{CoT}$-$\texttt{Direct} within $1.3$ points ($13$ pairs); on \texttt{VSI-Bench} and \texttt{MindCube} net $-0.6$ and $+0.1$ (null) but $y$ tracks $x$ ($r=0.84$); held-out net $-3.1$ to $+1.3$; on COCO occlusion twins the sign of $x$ matches the sign of \texttt{CoT}$-$\texttt{Direct} in $8/8$ models & Two-channel fit with shared-\texttt{Direct} permutation null; cue-set AME; held-out pool by task and depth split; scale ladder; closed APIs & \cref{sec:consequences,sec:frontier}, \cref{fig:two_channel}, \cref{tab:profile,tab:supp_e45,tab:supp_e35,tab:supp_heldout,tab:cohorts_summary,tab:supp_closed,tab:supp_e62} & \cref{sec:supp_heldout,sec:supp_closed}, \cref{tab:supp_two_channel,tab:supp_r29_ladder,tab:supp_e31,tab:supp_e43,tab:supp_e48,tab:supp_e51,tab:supp_cue_ame,tab:supp_e58b} & Largely an accounting consequence of Stage 2 ($y=x$ predicts natural points as well); both axes subtract \texttt{Direct} (null $r\approx0.72$); outliers \texttt{Phi-3.5-Vision} (low parse coverage) and \texttt{Molmo-7B-D}; occlusion AME rises in three height-reliant models; held-out CIs about $\pm4$ points \\
\textbf{Label-free fixes do not beat \texttt{Direct}.} & Symmetrization, margin gate, commit-sentence gate with count-matched controls, answer-first explanations, structured state, probe steering (oracle sign, an upper bound), pre-specified perception-first prompts (assistant and user turn; prediction files written and timestamped before the run) & \cref{sec:regrounding}, \cref{tab:interventions_summary} & \cref{sec:supp_interventions}, \cref{tab:supp_gate,tab:supp_state,tab:supp_e58a,tab:supp_steering_nonoracle,tab:supp_e55} & Gate criterion is the model's own \texttt{Direct} answer \\
\textbf{Lock-in is avoidable; answer-step re-grounding is a training target.} Three of six frontier closed models (\texttt{Claude Opus 5.5}, \texttt{GPT-6-Astra}, \texttt{Gemini-4-Argon}) follow the swapped twin image under their own fixed $R_\mathrm{cut}$ ($+74.0$ to $+92.0$ points on the hard set, $+100.0$ on the cue set); the three queried with reasoning off (\texttt{Qwen3.8-Max}, \texttt{DeepSeek-V4.1-Flash}, \texttt{HY4}) lock onto text ($+1.3$ to $+4.1$ hard, $+4.7$ to $+12.8$ cue), motivating re-grounding as a training target; a pre-specified LoRA pilot restores answer-step image use on natural photographs; fixed rationales with image-determined labels (MATCH) show re-grounding there, whereas counterfactual pairing mainly teaches discounting the text & Closed-model regenerated vs.\ fixed-rationale twin and gray-image swaps across three truncation rules; counterfactual-twin re-grounding pilot & \cref{sec:intro,sec:frontier,sec:regrounding,sec:discussion}, \cref{fig:teaser}c, \cref{tab:supp_e61,tab:supp_closed,tab:supp_e63_pilot} & \cref{sec:supp_closed,sec:supp_a3}, \cref{tab:supp_e50,tab:supp_e38} & Closed-model fixed-$R_\mathrm{cut}$ readout uses a new user turn; \texttt{Claude Opus 5.5} uses answer-turn hidden reasoning tokens (\texttt{Gemini-4-Argon} does not report them); \texttt{GPT-6-Astra} reasons on $145$ of $150$ hard-set twin calls, and its $5$ zero-reasoning hard pairs show no image effect \\
\bottomrule
\end{tabular}}
\end{table*}

\begin{table*}[t]
\centering
\caption{\textbf{Task-level accuracy (\%) on the $2{,}850$-item binary spatial pool.} \texttt{Direct} / \texttt{CoT} (\texttt{Filler($|R|$)+turn}; \texttt{Filler($|T|$)+turn} for \texttt{Qwen3-VL-8B-Thinking}) and paired differences (points) with $95\%$ image-cluster bootstrap CIs. First row per model (All): all $2{,}850$ items, the CIs behind the $^\ast$ marks of \cref{tab:behavioral_main}; \texttt{Fil}: same-turn \texttt{Filler($|R|$)}.}
\label{tab:supp_tasks}
\scriptsize
\setlength{\tabcolsep}{4pt}
\fitto{\textwidth}{\begin{tabular}{@{}llccccc@{}}
\toprule
Model & Task & \texttt{Direct} / \texttt{CoT} (\texttt{Fil+t}) & $\Delta$(\texttt{CoT}$-$\texttt{Direct}) & $\Delta$(\texttt{CoT}$-$\texttt{Fil+t}) & $\Delta$(\texttt{Fil+t}$-$\texttt{Direct}) & $\Delta$(\texttt{Fil+t}$-$\texttt{Fil}) \\
\midrule
Qwen2.5-VL-3B & \textbf{All ($2{,}850$)} & $80.3$ / $71.5$ ($79.8$) & $-8.8$\,{\scriptsize$[-10.6, -7.1]$} & $-8.3$\,{\scriptsize$[-10.0, -6.5]$} & $-0.5$\,{\scriptsize$[-1.3, +0.3]$} & $+2.0$\,{\scriptsize$[+1.0, +3.0]$} \\
 & CVB 2D Relation ($650$) & $73.7$ / $75.2$ ($74.2$) & $+1.5$\,{\scriptsize$[-1.7, +4.9]$} & $+1.1$\,{\scriptsize$[-2.3, +4.6]$} & & \\
 & CVB 3D Depth ($600$) & $78.8$ / $70.8$ ($79.0$) & $-8.0$\,{\scriptsize$[-11.8, -4.5]$} & $-8.2$\,{\scriptsize$[-12.0, -4.5]$} & & \\
 & CVB 3D Distance ($600$) & $68.7$ / $59.3$ ($67.5$) & $-9.3$\,{\scriptsize$[-14.5, -4.7]$} & $-8.2$\,{\scriptsize$[-13.5, -3.3]$} & & \\
 & Syn.\ RelPos ($500$) & $100.0$ / $96.2$ ($100.0$) & $-3.8$\,{\scriptsize$[-5.6, -2.2]$} & $-3.8$\,{\scriptsize$[-5.6, -2.2]$} & & \\
 & Syn.\ Depth ($500$) & $84.8$ / $57.4$ ($82.6$) & $-27.4$\,{\scriptsize$[-32.0, -22.8]$} & $-25.2$\,{\scriptsize$[-30.0, -20.4]$} & & \\
\midrule
Qwen2.5-VL-7B & \textbf{All ($2{,}850$)} & $85.9$ / $82.1$ ($86.6$) & $-3.9$\,{\scriptsize$[-5.4, -2.4]$} & $-4.6$\,{\scriptsize$[-6.1, -3.1]$} & $+0.7$\,{\scriptsize$[+0.1, +1.3]$} & $+10.0$\,{\scriptsize$[+8.6, +11.5]$} \\
\bottomrule
\end{tabular}}
\end{table*}

\begin{table*}[t]
\centering
\caption{\textbf{Attention knockouts on the conclusion-removed rationale ($R_\mathrm{cut}$), all seven models.} Flips and $|\Delta P(\text{gold})|$ (points) when the image is blocked from every post-image text row (\emph{image (all text)}), for \texttt{Direct} and $R_\mathrm{cut}$, and when the forced \texttt{``\textbackslash nAnswer: (''} rows are blocked from attending to the last remaining relation sentence through the end of $R_\mathrm{cut}$ (\emph{last relation}); population-weighted, twin-pair bootstrap CIs (flips). Full R: rationale with its conclusion. Lower block: 2D left/right items with appended gold/wrong conclusions (exploratory).}
\label{tab:supp_e59}
\scriptsize
\setlength{\tabcolsep}{2pt}
\fitto{\textwidth}{\begin{tabular}{llcccccccccc}
\toprule
 & & \multicolumn{2}{c}{Direct, image (all text)} & \multicolumn{2}{c}{R$_\mathrm{cut}$, image (all text)} & & Full R, image (all text) & \multicolumn{2}{c}{R$_\mathrm{cut}$, last relation} & \multicolumn{2}{c}{R$_\mathrm{cut}$ + concl.: $S$} \\
\cmidrule(lr){3-4}\cmidrule(lr){5-6}\cmidrule(lr){8-8}\cmidrule(lr){9-10}\cmidrule(lr){11-12}
Items & Model & flips & $|\Delta P|$ & flips & $|\Delta P|$ & ratio & flips & flips & $|\Delta P|$ & none & image (all text) \\
\midrule
Occlusion & Qwen2.5-VL-3B & 53.9\,{\tiny[50.2,57.7]} & 38.2 & 9.0\,{\tiny[6.7,11.6]} & 10.7 & 0.28 & 0.0\,{\tiny[0.0,0.0]} & 17.2\,{\tiny[14.7,19.8]} & 19.4 & -- & -- \\
 & Qwen2.5-VL-7B & 49.6\,{\tiny[45.9,53.1]} & 41.0 & 2.1\,{\tiny[1.0,3.4]} & 3.2 & 0.08 & 0.0\,{\tiny[0.0,0.0]} & 10.3\,{\tiny[8.6,12.4]} & 12.4 & -- & -- \\
 & InternVL3-8B & 49.6\,{\tiny[46.4,52.8]} & 30.2 & 2.2\,{\tiny[1.2,3.3]} & 2.9 & 0.10 & 1.3\,{\tiny[0.5,2.2]} & 23.6\,{\tiny[20.7,26.7]} & 26.6 & -- & -- \\
 & Phi-3.5-V & 26.6\,{\tiny[23.5,30.1]} & 18.0 & 7.7\,{\tiny[5.9,9.6]} & 7.9 & 0.44 & 1.2\,{\tiny[0.5,2.2]} & 22.5\,{\tiny[18.9,26.4]} & 26.6 & -- & -- \\
 & Molmo-7B-D & 43.5\,{\tiny[40.0,47.2]} & 25.4 & 7.4\,{\tiny[5.5,9.4]} & 7.4 & 0.29 & 0.2\,{\tiny[0.0,0.7]} & 17.8\,{\tiny[15.0,20.7]} & 20.0 & -- & -- \\
 & Qwen3-VL-8B & 38.3\,{\tiny[34.8,41.9]} & 38.6 & 4.6\,{\tiny[2.9,6.4]} & 5.1 & 0.13 & 0.0\,{\tiny[0.0,0.0]} & 19.1\,{\tiny[16.2,22.2]} & 19.9 & -- & -- \\
\bottomrule
\end{tabular}}
\end{table*}

\begin{table*}[t]
\centering
\caption{\textbf{Leverage of \texttt{Synthetic Depth}.} (a) Main-pool \texttt{CoT}$-$\texttt{Direct} with and without \texttt{Synthetic Depth}; (b) two-channel fits with and without its seven points; (c) agreement with the conclusion-removed body relation ($a_T$ / $a_F$).}
\label{tab:supp_e58b}
\scriptsize
\setlength{\tabcolsep}{2pt}
\fitto{\textwidth}{\input{tables/supp_e58b}}
\end{table*}

\noindent\fbox{\parbox{\dimexpr\columnwidth-2\fboxsep-2\fboxrule\relax}{\scriptsize\textbf{Glossary.}\\
\textbf{D$\pm$C$\pm$}: \texttt{Direct} correct ($+$) or wrong ($-$) and \texttt{CoT} correct or wrong on the same item (\eg, D+C$-$).\\
\textbf{Occlusion-competent} (main paper: passes the occlusion-reading rule): \texttt{Direct} cue-set accuracy CI low $>70\%$ and occlusion weight CI low $>0$ (\cref{sec:supp_competence}).\\
\textbf{$d'_\mathrm{pair}$}: pair-matched sensitivity from both queries of an image or twin pair, $\ln[(n_\mathrm{BC}+0.5)/(n_\mathrm{BW}+0.5)]/1.702$.\\
\textbf{$j^\ast$}: commit sentence, the first sentence after which forced-readout $P(\text{gold})$ falls below $0.5$ and stays there.\\
\textbf{$R_{\text{cut}}$, $S$, $\Delta_{\text{vis}}$}: rationale without its concluding answer sentence (its last spatial relation is the body relation); conclusion-swap selectivity $F(\text{gold})-F(\text{wrong})$, where $F(c)$ is the rate of answering the option stated by an appended conclusion $c$; for twin image swaps the (twin) image effect $P(\text{answer}=\text{gold}_v\mid v)-P(\text{answer}=\text{gold}_v\mid v_{\text{cf}})$.\\
\textbf{Pre-specified}: hypotheses and decision rules written to a timestamped file before the run (internal, not a public registry).\\
\textbf{F+t}: \texttt{Filler($|R|$)+turn}, length-matched neutral filler with the assistant turn closed before the answer prompt.\\
\textbf{$s{-}1$, $s$, $r{-}1$, $r$, $a$}: probe positions writing the subject, at the end of the subject noun, writing the relation word, after reading it, and the answer token (\texttt{answer} in \cref{tab:supp_probe_align}).}}
\vspace{4pt}

\begin{table*}[t]
\centering
\caption{\textbf{95\% CIs for main \cref{tab:profile}.} Each cell repeats the main-table value with its 95\% CI in brackets; $n$ is in the group header or row label or, where it varies by model, in the small row below ($k/n$ for the two-judge \texttt{Syn.\ Depth} $431$ audit). Cue-factorial set ($1{,}600$ queries, $400$ base scenes): \texttt{Direct} accuracy uses the bootstrap of the occlusion-reading rule (twin pair $\times$ phrasing clusters, $2{,}000$ resamples); lower-is-closer and the written-relation agreement rates resample base scenes ($2{,}000$); the height AME and the cue-set accuracy change resample twin pairs with \texttt{Direct} and \texttt{CoT} jointly ($1{,}000$ and $300$ draws). Commit at a relational sentence: relational share of the commit sentences of committed \texttt{D+C$-$} rationales (2,850-item pool), image-cluster bootstrap ($2{,}000$). Answer = written relation: cue queries with a parsed last relation that is true / false. True first-claim edit: \texttt{Synthetic Depth} and twin items with a parsable first claim, image clusters. Twin image effect ($800$ twin queries, $200$ twin pairs): \texttt{Direct} $=2\times$ accuracy $-100$, own $R_\mathrm{cut}$ = rationale held fixed while the image is swapped; twin-pair bootstrap ($2{,}000$). \texttt{Syn.\ Depth} $431$ commit / matched: share of judgeable commit sentences (\texttt{D+C$-$}) and of matched relational \texttt{D+C$+$} sentences that contradict the image, Wilson intervals. \texttt{Syn.\ Depth} $431$ accuracy: image bootstrap ($1{,}000$); held-out $767$ and depth $324$: image-cluster bootstrap ($2{,}000$). $^\ast$: CI excludes zero; $^\ddagger$: empty \texttt{BLINK} rationales (format effect); --: not run. Percentile intervals throughout; not corrected for multiple comparisons.}
\label{tab:supp_tab2_ci}
\scriptsize
\setlength{\tabcolsep}{1.25pt}
\renewcommand{\arraystretch}{0.92}
\fitto{\textwidth}{\begin{tabular}{@{}lccccccc@{}}
\toprule
\textbf{Model} & \textbf{\shortstack{Qwen2.5-VL\\3B}} & \textbf{\shortstack{Qwen2.5-VL\\7B}} & \textbf{\shortstack{InternVL3\\8B}} & \textbf{\shortstack{Phi-3.5\\Vision}} & \textbf{\shortstack{Molmo\\7B-D}} & \textbf{\shortstack{Qwen3-VL\\8B}} & \textbf{\shortstack{InternVL3.5\\8B}} \\
\midrule
\rowcolor{cambriangray}\multicolumn{8}{@{}l@{}}{\textit{\textbf{What \texttt{Direct} relies on (cue-factorial set, $1{,}600$ queries)}}} \\
Accuracy (\%) & $52.4${\tiny[49.2,55.8]} & $53.4${\tiny[50.2,56.6]} & $63.7${\tiny[60.8,66.5]} & $59.5${\tiny[57.5,61.6]} & $88.2${\tiny[86.6,89.7]} & $78.9${\tiny[77.0,80.8]} & $55.6${\tiny[52.5,58.8]} \\
Follows lower-is-closer (\%) & $96.2${\tiny[95.2,97.1]} & $95.6${\tiny[94.4,96.7]} & $84.9${\tiny[82.9,86.9]} & $62.6${\tiny[60.6,64.8]} & $57.2${\tiny[53.4,61.1]} & $62.2${\tiny[59.4,65.4]} & $93.8${\tiny[92.5,95.1]} \\
\midrule
\rowcolor{cambriangray}\multicolumn{8}{@{}l@{}}{\textit{\textbf{Stage 1: written relations are often false (commit sentences contradicting the image)}}} \\
\texttt{Syn.\ Depth} 431: commit (\%) & $100${\tiny[88.6,100.0]} & $52${\tiny[34.4,68.6]} & $97${\tiny[83.3,99.4]} & $100${\tiny[89.6,100.0]} & $77${\tiny[57.9,89.0]} & $96${\tiny[81.7,99.3]} & $91${\tiny[77.0,97.0]} \\
{\tiny \quad $k/n$ judgeable} & {\tiny 30/30} & {\tiny 15/29} & {\tiny 29/30} & {\tiny 33/33} & {\tiny 20/26} & {\tiny 26/27} & {\tiny 31/34} \\[-1pt]
\quad matched (\%) & $7${\tiny[2.0,22.6]} & $7${\tiny[1.8,21.3]} & $16${\tiny[6.9,31.8]} & $19${\tiny[8.9,35.3]} & $10${\tiny[3.5,25.6]} & $4${\tiny[0.8,21.0]} & $0${\tiny[0.0,12.1]} \\
\bottomrule
\end{tabular}}
\end{table*}

\section{Implementation, Stimuli, and Judges}
\label{sec:supp_impl}

\paragraph{Models, engines, and image budgets.}
All open models use \texttt{bfloat16} with greedy decoding (temperature $0$), prefix caching, and seed $0$. \texttt{Qwen2.5-VL-3B/7B}~\citep{bai2025qwen25vl}, \texttt{InternVL3-8B}~\citep{zhu2025internvl3}, and \texttt{Phi-3.5-Vision}~\citep{abdin2024phi3} run on the \texttt{vLLM} 0.8.4 V0 engine~\citep{kwon2023vllm} ($12{,}288$-token context, up to two images per prompt), whereas \texttt{Molmo-7B-D}~\citep{deitke2024molmo} uses the V1 engine of the same version ($4{,}096$-token context, one image slot). \texttt{Qwen3-VL-8B} (Instruct and Thinking) and \texttt{InternVL3.5-8B} run on \texttt{vLLM} 0.11 (V1 only). Per-image visual budgets follow each model's standard configuration: at most $1280 \times 28 \times 28$ pixels for the \texttt{Qwen2.5-VL} models (\texttt{vLLM} 0.11 ignores \texttt{max\_pixels} for \texttt{Qwen3-VL}, which therefore ran at its processor default; the rendered sets and COCO twins are $\le 1$\,MP and unaffected), up to $6$ dynamic tiles for the InternVL models (\texttt{max\_dynamic\_patch=6}), and $4$ crops for \texttt{Phi-3.5-Vision} (\texttt{num\_crops=4}). Each model's rationale $R$ is generated once per item as a greedy continuation of \texttt{``Let's think step by step.''} (at most $512$ new tokens) and kept fixed across all post-hoc conditions. Mean $|R|$ is $124$, $75$, $131$, $51$, and $99$ tokens for \texttt{Qwen2.5-VL-3B}, \texttt{Qwen2.5-VL-7B}, \texttt{InternVL3-8B}, \texttt{Phi-3.5-Vision}, and \texttt{Molmo-7B-D}, $217.7$ (median $192$; $2.8\%$ reach $512$) for \texttt{Qwen3-VL-8B}, and $138.0$ (median $142$; none truncated) for \texttt{InternVL3.5-8B}; among the first five models, truncation at $512$ tokens occurs on $<0.1\%$ of \texttt{Qwen2.5-VL-3B} items and never in the other four.

\paragraph{Prompt templates and neutral filler.}
Every condition shares the first user turn \texttt{[<image>, Q]} with lettered options (\texttt{(A) yes} / \texttt{(B) no} on synthetic items) and ends with the open assistant prefix \texttt{``Answer: (''}:
\begin{itemize}
    \item \texttt{Direct}: user \texttt{[<image>, Q]}; assistant \texttt{"Answer: ("}.
    \item \texttt{Direct + restate}: user \texttt{[<image>, "Q\textbackslash n\textbackslash nQuestion again: Q\textbackslash nNow give your final answer."]}; assistant \texttt{"Answer: ("}.
    \item \texttt{Same-turn text} ($C \in \{R, \text{Filler}\}$): user \texttt{[<image>, Q]}; assistant \texttt{"C\textbackslash nAnswer: ("}.
    \item \texttt{Extra-turn text}: user \texttt{[<image>, Q]}; assistant \texttt{"C"} (turn closed); second user turn (\texttt{"Now give your final answer."} for \texttt{turn}; \texttt{"Question again: Q\textbackslash nNow give your final answer."} for \texttt{restate}; \texttt{[<image>, "Here is the image again.\textbackslash nNow give your final answer."]} for \texttt{reimage}; or both for \texttt{reimage+restate}); assistant \texttt{"Answer: ("}.
\end{itemize}
Neutral filler cycles twelve non-spatial sentences joined by spaces and truncated to exact token count $n \in \{|R|, 512, 1024\}$ under each model's tokenizer: (1) \emph{The history of paper making spans many centuries and several cultures.} (2) \emph{Early writers recorded ideas on bark, cloth, clay and animal skin.} (3) \emph{Later, workers learned to soak plant fibers and press them into thin sheets.} (4) \emph{The method spread slowly through trade and the movement of skilled craftsmen.} (5) \emph{Mills powered by water made production faster and cheaper.} (6) \emph{Printed books became common, and literacy rates rose over time.} (7) \emph{Today most paper is produced from wood pulp in large industrial plants.} (8) \emph{Recycling programs recover a large share of used paper every year.} (9) \emph{Some artists still prefer handmade sheets for their texture and strength.} (10) \emph{Libraries preserve old documents by controlling humidity and temperature.} (11) \emph{Scholars study watermarks to estimate the age of historical manuscripts.} (12) \emph{The chemistry of acid-free paper helps documents last for generations.}

\paragraph{Free-text parsing and statistics.}
The option stated inside $R$ is parsed with three ordered patterns (\texttt{``answer is (X)''}, \texttt{``(X)''}, and a bare letter before \texttt{.}/\texttt{)}/\texttt{:}), taking the last match. For spatial relations inside $R$, rule parser v3 (\texttt{claim\_parse v3}) splits $R$ after the \texttt{CoT} prefix into sentences, matches sentences that name both shapes with an active (\eg, \emph{in front of}, \emph{on top of}, \emph{covers}, \emph{overlaps over}, \emph{closer}) or passive (\emph{behind}, \emph{covered/hidden/occluded by}, \emph{underneath}, \emph{farther}) depth predicate while skipping symmetric verbs (\emph{overlaps}, \emph{touches}, \emph{intersects}) and question/hypothetical clauses, and flips the claimed front shape under negation (\eg, \emph{is not in front of}). Accuracies and paired differences use $2{,}000$-resample image-cluster bootstrap $95\%$ CIs and two-sided image-cluster sign-flip $p$-values; occlusion $d'$ uses a $2{,}000$-resample twin-pair bootstrap. Per-model CIs are not corrected for multiple comparisons. Under Benjamini--Hochberg FDR control ($q < 0.05$) jointly over the $60$ per-model tests behind the main claims, including the \texttt{Synthetic Depth} $\Delta d'_\mathrm{pair}$, $52$ of the $53$ effects whose nominal CI excludes zero survive, and the one that does not is the \texttt{Qwen2.5-VL-7B} twin $d'$ rise ($q = 0.052$; with Wald $p$, the \texttt{Qwen2.5-VL-7B} probit cue-occlusion shift also fails, $q = 0.053$; \cref{tab:supp_multiplicity}). Throughout, an effect is nominal only if the CI shown in the tables excludes zero at the shown precision.

\begin{table*}[t]
\centering
\caption{\textbf{Multiplicity control over the per-model tests} (Benjamini--Hochberg over all $60$ tests jointly; the family was fixed before the \texttt{Qwen3-VL-8B-Thinking} cue-set run, whose results are in \cref{tab:supp_e31}, hence its dashes; including the \texttt{Synthetic Depth} $\Delta d'_\mathrm{pair}$; $52$ of $53$ nominal effects survive). With Wald $p$ instead of resampling $p$, the \texttt{Qwen2.5-VL-7B} probit cue-occlusion shift exceeds the threshold ($q = 0.053$ vs.\ $0.020$), so the probit cue-occlusion shift survives in six of seven models under Wald-based FDR (probability-scale AME is the primary cue-effect scale; \cref{sec:supp_cues,tab:supp_cue_ame}). With the $431$-image and twin $\Delta d'_\mathrm{pair}$ estimators in place of the two \texttt{Synthetic Depth} and twin columns, all $50$ nominal effects survive, with the \texttt{Phi-3.5-Vision} $431$-image loss not counted because its shown CI touches zero.}
\label{tab:supp_multiplicity}
\scriptsize
\setlength{\tabcolsep}{3pt}
\fitto{\textwidth}{\begin{tabular}{lcccccccc}
\toprule
Model & CoT$-$Dir.\ acc. & CoT$-$F$+$t acc. & Syn.\ Depth pooled $\Delta d'$ & Syn.\ Depth $\Delta d'_\mathrm{pair}$ & Twin $\Delta d'$ & CV-B 3D $\Delta d'$ & Cue occl.\ $\Delta$ & Cue vert.\ $\Delta$ \\
\midrule
Qwen2.5-VL-3B & $\mathbf{-8.8}$ & $\mathbf{-8.3}$ & $\mathbf{-1.77}$ & $\mathbf{-1.50}$ & $+0.05$ & $\mathbf{-0.49}$ & $\mathbf{-0.51}$ & $\mathbf{-2.99}$ \\
Qwen2.5-VL-7B & $\mathbf{-3.9}$ & $\mathbf{-4.6}$ & $\mathbf{-0.58}$ & $\mathbf{-0.16}$ & $+0.21^{\circ}$ & $-0.25$ & $\mathbf{-0.34}$ & $\mathbf{-2.19}$ \\
InternVL3-8B & $\mathbf{-5.1}$ & $\mathbf{-4.1}$ & $\mathbf{-1.83}$ & $\mathbf{-0.94}$ & $\mathbf{+0.43}$ & $+0.05$ & $\mathbf{-0.56}$ & $\mathbf{-2.44}$ \\
Phi-3.5-V & $\mathbf{-9.2}$ & $\mathbf{-6.9}$ & $\mathbf{-0.40}$ & $\mathbf{-0.23}$ & $\mathbf{-0.23}$ & $\mathbf{-1.21}$ & $\mathbf{-0.31}$ & $\mathbf{-0.62}$ \\
Molmo-7B-D & $-1.7$ & $-1.6$ & $\mathbf{-1.19}$ & $\mathbf{-0.73}$ & $\mathbf{-0.94}$ & $\mathbf{-0.43}$ & $\mathbf{-1.18}$ & $\mathbf{-0.55}$ \\
Qwen3-VL-8B & $\mathbf{-3.5}$ & $\mathbf{-4.0}$ & $\mathbf{-1.18}$ & $\mathbf{-1.32}$ & $\mathbf{-0.46}$ & $\mathbf{-0.59}$ & $\mathbf{-1.01}$ & $\mathbf{-2.13}$ \\
Qwen3-VL-8B-Think. & $\mathbf{-1.4}$ & $\mathbf{-1.4}$ & -- & -- & $\mathbf{-0.65}$ & $+0.04$ & -- & -- \\
\end{tabular}}
\end{table*}

\subsection{Engine and readout notes}
\label{sec:supp_engine}

\paragraph{Exact readout.} For \texttt{Qwen2.5-VL}, \texttt{InternVL3-8B}, and \texttt{Phi-3.5-Vision}, a per-request logits processor on the V0 engine reads the raw logits row at \texttt{Answer: (} and stores the full-vocabulary log-softmax of the letter tokens (bare, space-prefixed, and lower-case variants; the main definition sums the variants). Option-letter mass is $0.97$--$1.00$, and exact and top-$20$ probabilities of the gold option agree to $10^{-4}$ on average.
\paragraph{Molmo-7B-D.} Its model processor wraps every prompt as ``User: \dots{} Assistant:'' by default, which appends a second ``Assistant:'' after \texttt{Answer: (}. We pass \texttt{message\_format="none"} so that the chat template alone produces the native ``User: \dots{} Assistant: Answer: ('' prompt. Under the V0 engine its option-letter mass is about $1\%$, so only V1 outputs are used.
\paragraph{V1-engine models.} \texttt{Molmo-7B-D}, \texttt{Qwen3-VL-8B} (Instruct and Thinking), and \texttt{InternVL3.5-8B} use the top-$20$ letter tokens, which never missed an option in our runs (median option mass $0.9987$ for \texttt{Molmo-7B-D}). \texttt{InternVL3.5-8B} runs with \texttt{TRITON\_ATTN} because the vision encoder's flash-attention kernel failed on our driver. The Thinking model generates its native thinking trace (up to $4{,}096$ tokens), which is closed with \texttt{</think>} before \texttt{Answer: (} (mean $307$ tokens on the $2{,}850$-item main pool, \cref{tab:behavioral_main}; $193$ tokens on the cue-factorial set, \cref{tab:supp_e31}).
\paragraph{Re-runs.} The \texttt{D$\pm$C$\pm$} groups of \cref{fig:error_onset} and of the commit, rescue, and steering analyses use the original runs' top-$20$ letter scores; in the four V0-engine models these cells differ from those of the exact \cref{tab:behavioral_main} readout on $83$, $27$, $22$, and $40$ near-tie items, and on none in the other models (\cref{tab:supp_churn}b). \cref{tab:behavioral_main} re-reads the original greedy rationales, and the within-claim probe teacher-forces those stored rationales in HuggingFace \texttt{Transformers} (\cref{sec:supp_a2}), whereas the probe-steering, structured-state, gate, and answer-first experiments regenerate rationales in later \texttt{vLLM} runs (whose greedy outputs differ from the stored ones on $35.8\%$ in \texttt{Qwen3-VL-8B} to $79.8\%$ in \texttt{Molmo-7B-D} of \texttt{Synthetic Depth} items due to \texttt{bfloat16} batch-reduction non-determinism across batch sizes and \texttt{vLLM} versions; each intervention table uses its own matched same-batch baseline). In those regenerated runs, the \texttt{CoT} answer is unchanged on $83.0\%$--$97.2\%$, and \texttt{CoT} accuracy moves by $-3.6$ (\texttt{Qwen2.5-VL-7B}), $-2.6$ (\texttt{InternVL3-8B}), and at most $1.2$ points in the other models; \texttt{Direct} answers are identical, except that the gate experiment's own \texttt{Direct} run changes one \texttt{Molmo-7B-D} answer per set (cues $1{,}410$ vs.\ $1{,}411$ of $1{,}600$ correct, twins $706$ vs.\ $705$ of $800$; \cref{tab:supp_gate}), and the \texttt{Qwen3-VL-8B} scale-ladder row uses a re-run of the twins.
\paragraph{Conclusion-removal rules.}
Four deterministic rules strip concluding text from a stored rationale $R$ before forced readout or relation extraction:
(1)~$R_\mathrm{cut}$ (\emph{paper default}; legacy \texttt{e5\_content\_edit.strip\_answer}): cuts $R$ at the sentence boundary (\texttt{". "}, \texttt{"\textbackslash n"}, or \texttt{": "}) preceding the last match of \texttt{answer|\textbackslash([A-D]\textbackslash)|\textbackslash b[A-D]\textbackslash)|option [A-D]\textbackslash b}, repeated up to four times while the tail matches and at least $30\%$ of $R$ remains (never cutting into \texttt{"Let's think step by step."}); used in \cref{tab:profile,sec:error_onset,sec:answer_step,tab:supp_e57,tab:supp_e59,tab:supp_e60,tab:supp_e62,tab:supp_recheck}.
(2)~\emph{kept} (\texttt{e38\_closed\_judge.strip\_answer}): removes from the last \texttt{"Answer:"} (or \texttt{answer\textbackslash s*:} (ASCII or full-width colon)) onward plus trailing sentences that match \texttt{\textbackslash b(answer|option|choice)\textbackslash b} or bare \texttt{(A)}/\texttt{(B)}/\texttt{yes}/\texttt{no} without a depth word, keeping the concluding sentence; used in \cref{tab:supp_e61,tab:supp_e38}.
(3)~\emph{uniform rule} (\texttt{e35\_natural\_two\_channel.truncate}): splits $R$ into sentences (\texttt{e2f\_claim\_edit.split\_sents}), drops everything after the last answer-stating sentence (\texttt{ANS}) and every earlier sentence matching \texttt{ANS}, drops any \texttt{Conclusion}/\texttt{Final Answer}/\texttt{Summary} header (\texttt{HEADER}, \texttt{HEADER\_INLINE}) and all following sentences, and strips trailing concluding sentences starting with \texttt{therefore}/\texttt{thus}/\texttt{so}/\texttt{hence}/\texttt{in conclusion}/\texttt{based on}/\texttt{overall} (\texttt{CONC}) or trailing sentences that merely restate an option (\texttt{restates}); used in \cref{tab:supp_e54a,tab:supp_e57,tab:supp_e61,tab:supp_e50,tab:supp_e45,tab:supp_e35,tab:supp_e54b,tab:supp_written_ame,tab:supp_e51,tab:supp_e58a}.
(4)~\emph{strict} (\texttt{e45\_uniform\_two\_channel.strict\_body}; on COCO twins, \cref{tab:supp_e62}, the strict variant is instead $R_\mathrm{cut}$ minus trailing concluding sentences, \texttt{e62\_run.strict\_cut}): applies \texttt{truncate} and then iteratively drops trailing sentences containing a mid-sentence conclusion clause (\texttt{,/;/-/--- so|therefore|thus|hence|which means|meaning|indicating that|confirming} or \texttt{this|that|which means|indicates|confirms|shows}, \texttt{MIDCONC}), re-applying \texttt{truncate} after each removal; used in \cref{tab:supp_e61}.
Full per-model truncation sensitivity across the $800$ occlusion-twin queries is reported in \cref{tab:supp_x2_trunc} (summarized in \cref{sec:answer_step}); on the cue set, the uniform rule yields $16.6$ $[15.0, 18.4]$ ($29.3$ $[22.6, 36.5]$ on \texttt{D+C$-$}) for \texttt{Qwen3-VL-32B} (vs.\ $0.2$ under $R_\mathrm{cut}$) and $4.1$ $[3.0, 5.4]$ for \texttt{Qwen3-VL-8B}, and on COCO twins the strict rule raises the seven-model mean from $+3.0$ to $+5.5$ (\texttt{Qwen3-VL-8B} $+2.9\to+5.0$, \texttt{Qwen3-VL-32B} $+0.4\to+3.1$).

\begin{table*}[t]
\centering
\caption{\textbf{Truncation sensitivity of the fixed-rationale twin swap} ($800$ occlusion-twin queries; image effect in points with twin-pair bootstrap CIs; GPU re-read). Rules as in \cref{sec:supp_engine}; changed: share of rationales the uniform / strict rule alters relative to $R_\mathrm{cut}$.}
\label{tab:supp_x2_trunc}
\scriptsize
\setlength{\tabcolsep}{4pt}
\fitto{\textwidth}{\begin{tabular}{lrrrrr}
\toprule
Model & \texttt{Direct} & $R_\mathrm{cut}$ & uniform & strict & changed (\%) \\
\midrule
Qwen2.5-VL-3B & $3.0$ {\tiny[$-$10.5, 15.8]} & $2.0$ {\tiny[$-$1.4, 5.4]} & $3.2$ {\tiny[$-$2.9, 9.4]} & $3.1$ {\tiny[$-$3.2, 9.4]} & 61 / 69 \\
Qwen2.5-VL-7B & $8.0$ {\tiny[$-$5.0, 20.2]} & $3.9$ {\tiny[1.1, 6.5]} & $6.8$ {\tiny[3.2, 10.1]} & $8.1$ {\tiny[3.7, 12.4]} & 38 / 54 \\
InternVL3-8B & $26.8$ {\tiny[16.0, 36.8]} & $5.0$ {\tiny[3.5, 6.8]} & $9.2$ {\tiny[6.0, 12.8]} & $10.5$ {\tiny[6.9, 14.4]} & 36 / 41 \\
Phi-3.5-V & $26.8$ {\tiny[20.5, 32.8]} & $4.4$ {\tiny[2.1, 6.8]} & $3.5$ {\tiny[1.1, 5.9]} & $8.1$ {\tiny[4.2, 11.9]} & 22 / 43 \\
Molmo-7B-D & $76.5$ {\tiny[72.2, 81.0]} & $8.9$ {\tiny[6.6, 11.1]} & $15.5$ {\tiny[12.9, 18.2]} & $15.9$ {\tiny[13.1, 19.0]} & 35 / 64 \\
Qwen3-VL-8B & $62.3$ {\tiny[56.8, 67.8]} & $8.4$ {\tiny[5.9, 11.1]} & $4.1$ {\tiny[2.4, 6.0]} & $9.4$ {\tiny[6.6, 12.1]} & 80 / 84 \\
InternVL3.5-8B & $13.3$ {\tiny[1.2, 25.0]} & $2.0$ {\tiny[$-$0.5, 4.6]} & $3.4$ {\tiny[0.0, 7.0]} & $0.9$ {\tiny[$-$3.5, 5.4]} & 24 / 34 \\
Qwen3-VL-32B & $92.8$ {\tiny[90.2, 95.0]} & $0.1$ {\tiny[0.0, 0.4]} & $14.1$ {\tiny[12.0, 16.5]} & $24.2$ {\tiny[21.4, 27.1]} & 75 / 86 \\
\bottomrule
\end{tabular}}
\end{table*}

\subsection{Occlusion competence, heuristic baselines, and Synthetic Depth}
\label{sec:supp_competence}

\paragraph{Competence rule.} A model is occlusion-competent if (i) its \texttt{Direct} accuracy on the size- and height-balanced cue-factorial set has a $95\%$ CI lower bound above $70\%$ and (ii) its \texttt{Direct} probit occlusion weight (\cref{tab:supp_glm}a) has a CI lower bound above $0$; CIs resample twin pairs, and \texttt{Qwen3-VL-8B-Thinking}, whose cue-set run was added later, is evaluated in \cref{tab:supp_e31}. The rule was fixed in the analysis code before the per-model table was produced; the cue-set accuracies had been reported in earlier analyses, so the rule is explicit but not blind. Any accuracy threshold on the CI lower bound above $60.8\%$ (\texttt{InternVL3-8B}) and below $77.0\%$ (\texttt{Qwen3-VL-8B}) gives the same split of the seven instruct models, and criterion (ii) holds for all seven. \texttt{MiMo-VL-7B-RL} and \texttt{GLM-4.1V-9B-Thinking} ($71.5\%$, $71.3\%$; \cref{tab:supp_e48}) pass only for thresholds below $71.3\%$; above it, all four remaining occlusion readers write relations less accurate than \texttt{Direct} on the cue set when the conclusion is kept (Stage 1; $x=-7.7$ to $-2.2$, \cref{tab:supp_two_channel}, \cref{sec:supp_a4}). \texttt{Molmo-7B-D} ($88.2\%$ $[86.6, 89.7]$; weight $2.72$ $[2.56, 2.93]$) and \texttt{Qwen3-VL-8B} ($78.9\%$ $[77.0, 80.8]$; $2.79$ $[2.53, 3.11]$) pass. \texttt{InternVL3-8B} fails (i) only ($63.7\%$ $[60.8, 66.5]$; weight $1.81$ $[1.61, 2.14]$), and the other four fail (i) at $52.4\%$--$59.5\%$ (\cref{tab:profile}). Competence does not extend to the hard occlusion set, where both competent models are near chance under \texttt{Direct} ($55.2\%$, $54.9\%$).

\paragraph{Heuristic baselines.} \cref{tab:supp_heuristics} scores rules that answer without reading occlusion. On \texttt{Synthetic Depth}, ``larger visible area is in front'' is right on $98.6\%$ $[97.1, 99.3]$ of the $500$ images (wrong on $7$; $98.4\%$ on the $431$ overlapping ones), ``lower is closer'' on $52.4\%$ $[48.0, 56.7]$, and a five-fold logistic on the visible-area ratio and height on $98.2\%$ (median visible-area ratio $4.86$). Size and occlusion are therefore nearly collinear there, and how much of \texttt{Direct}'s accuracy comes from size is not identifiable: each model's two-order \texttt{Direct} accuracy equals its agreement with the size rule within $0.8$ points. The other sets separate the cues. On the twins the drawn-size rule is right on $0\%$ and the visible-size rule on $61.0\%$; on the cue set drawn size gives $50\%$ by design and visible size $64.6\%$, because occlusion hides part of the back shape. No rule reaches the competent models' cue-set accuracy. The shortcut-reliant models follow ``lower is closer'' on $84.1\%$--$98.0\%$ of twin queries and $84.9\%$--$96.2\%$ of cue queries (\texttt{Qwen2.5-VL-3B/7B}, \texttt{InternVL3-8B}, \texttt{InternVL3.5-8B}), and \texttt{CoT} lowers the cue-set share to $58.6\%$--$81.6\%$.

\begin{table*}[t]
\centering
\caption{\textbf{Heuristic baselines and occlusion competence.} Top: accuracy (\%) of rules that ignore occlusion on each set (\texttt{Synthetic Depth}: all $500$ / $431$ visibly overlapping images; twins and cue set: images under the ``in front of'' reading). Bottom: per model, \texttt{Direct} two-order accuracy on \texttt{Synthetic Depth} and its agreement with the visible-size rule; twin and cue-set \texttt{Direct} accuracy and the share of answers naming the lower shape as closer (\texttt{CoT} in parentheses); competence under the rule of \cref{sec:supp_competence}.}
\label{tab:supp_heuristics}
\scriptsize
\setlength{\tabcolsep}{3pt}
\fitto{\textwidth}{\begin{tabular}{lcccc}
\toprule
Rule (answer without reading occlusion) & \texttt{Syn.\ Depth} 500 / 431 & Twins (400 img.) & Cue set (1{,}600) & Hard set (288) \\
\midrule
Larger drawn area is in front & -- & $0.0$ & $50.0$ & $0.0$ \\
Larger visible area is in front & $98.6$ / $98.4$ & $61.0$ & $64.6$ & $14.2$ \\
Lower in the image is closer & $52.4$ / $50.8$ & $50.0$ & $50.0$ & $50.0$ \\
Logistic on visible ratio + height (5-fold) & $98.2$ / $98.1$ & $63.0$ & $64.4$ & $86.5$ \\
\midrule
\multicolumn{5}{l}{\textit{Per model, \texttt{Direct} (two object orders on \texttt{Syn.\ Depth}; \%)}} \\
Model & \texttt{Syn.\ Depth} acc.\ / agrees with size rule & Twins acc.\ / follows lower & Cue acc.\ / follows lower (\texttt{CoT}) & Competent \\
\midrule
Qwen2.5-VL-3B & $82.2$ / $82.0$ & $51.5$ / $98.0$ & $52.4$ / $96.2$ ($69.4$) & no \\
Qwen2.5-VL-7B & $78.3$ / $78.1$ & $54.0$ / $95.5$ & $53.4$ / $95.6$ ($70.4$) & no \\
InternVL3-8B & $90.3$ / $90.1$ & $63.4$ / $84.1$ & $63.7$ / $84.9$ ($58.6$) & no \\
Phi-3.5-V & $65.0$ / $65.0$ & $63.4$ / $64.9$ & $59.5$ / $62.6$ ($55.2$) & no \\
Molmo-7B-D & $88.5$ / $88.1$ & $88.1$ / $58.4$ & $88.2$ / $57.2$ ($54.1$) & yes \\
Qwen3-VL-8B & $77.9$ / $77.1$ & $81.1$ / $61.4$ & $78.9$ / $62.2$ ($41.4$) & yes \\
InternVL3.5-8B & $55.8$ / $55.6$ & $56.6$ / $92.6$ & $55.6$ / $93.8$ ($81.6$) & no \\
\bottomrule
\end{tabular}}
\end{table*}

\paragraph{\texttt{Synthetic Depth}: overlapping images and pair-matched sensitivity.} Of the $500$ \texttt{Synthetic Depth} images, $431$ show visible overlap, $40$ only touch or are separated (gold follows size), and $29$ are nearly fully hidden. The main text analyzes the $431$; \cref{tab:supp_syn431}a gives both. On the $431$, $d'_\mathrm{pair}$ falls with CI below zero in four models ($-0.78$ to $-1.55$) against six on all $500$: \texttt{Qwen2.5-VL-7B} ($-0.14$ $[-0.30, +0.03]$) and \texttt{Phi-3.5-Vision} ($-0.21$ $[-0.40, +0.00]$) lose significance, and \texttt{InternVL3.5-8B} still rises ($+0.19$). Pooled $d'$ falls in six models on both sets ($-0.38$ to $-1.89$ on the $431$), and the commit audit is unchanged ($184/209 = 88.0\%$ vs.\ $19/203 = 9.4\%$).
Each image yields one of four answer pairs under the two object orders: both orders correct (BC), ``yes'' to both (YY, affirming both queried relations), ``no'' to both (NN), or both wrong (BW, a consistent reversed percept). We assume $P(\text{yes}) = \sigma(b_i \pm \Delta/2)$ for image $i$ under the two orders, with a free per-image bias $b_i$ and a sensitivity $\Delta$ that is constant over images, and conditionally independent answers. Then $n_\mathrm{BC}/n_\mathrm{BW}$ estimates $e^{\Delta}$ whatever the $b_i$, and $d'_\mathrm{pair} = \ln[(n_\mathrm{BC}+0.5)/(n_\mathrm{BW}+0.5)]/1.702$ is a pair-matched (conditional-logit) sensitivity on the probit scale that, under this common-sensitivity model, removes any per-image bias, including affirmation. Under \texttt{CoT}, $d'_\mathrm{pair}$ falls in six models, with CIs below zero ($-0.16$ to $-1.50$), and rises in \texttt{InternVL3.5-8B} ($+0.20$; \cref{tab:supp_syn_pairmh}b). The common-sensitivity model is testable: with a common $\Delta$, the share of images answered differently under the two orders, BC$+$BW, cannot exceed $\sigma(\Delta/2)^2+\sigma(-\Delta/2)^2$. Under \texttt{Direct} the observed share exceeds this bound in four models (\texttt{Qwen2.5-VL-7B}, \texttt{InternVL3-8B}, \texttt{Phi-3.5-Vision}, \texttt{Molmo-7B-D}), so sensitivity varies across images, and $d'_\mathrm{pair}$ is then a bias-weighted mean of image-level sensitivities, which a bias shift could lower by moving easy images out of the discordant set. We therefore fix the stratum weights across arms (size-standardised $d'_\mathrm{pair}$): over visible-area-ratio terciles it falls in five of seven models and over quintiles in six, and \texttt{InternVL3.5-8B} rises in both ($+0.20$). By tercile, $d'_\mathrm{pair}$ falls significantly in $1$, $5$, and $4$ of seven models (low, mid, high ratio), so the loss concentrates on size-asymmetric pairs. A logistic mixed model with an image random sensitivity slope, which does not assume a common sensitivity, is weakly identified with two answers per image and arm; its mean $d'$ falls with CI below zero in three of seven models (\texttt{Qwen2.5-VL-3B}, \texttt{Qwen2.5-VL-7B}, \texttt{InternVL3-8B}) and is indeterminate in the other four.

\begin{table*}[t]
\centering
\caption{\textbf{\texttt{Synthetic Depth} sensitivity checks} (both object orders; image bootstrap $95\%$ CIs). \textbf{(a)} All $500$ images vs.\ the $431$ visibly overlapping images (\texttt{CoT}$-$\texttt{Direct}); last columns: two-judge consensus, commit sentences contradicting the image / judgeable, and relational \texttt{D+C$+$} comparators contradicting / judgeable. \textbf{(b)} Common-sensitivity bound check, pooled and size-standardised $\Delta d'_\mathrm{pair}$, and $d'_\mathrm{pair}$ by visible-area-ratio tercile (from its own resampling, so pooled CIs can differ from (a) in the second decimal).}
\label{tab:supp_syn431}
\label{tab:supp_syn_pairmh}
\scriptsize
\setlength{\tabcolsep}{1.4pt}
\textbf{(a)}\ \fitto{0.96\textwidth}{\begin{tabular}{lcccccc}
\toprule
 & \multicolumn{2}{c}{$\Delta d'_\mathrm{pair}$ [95\% CI]} & \multicolumn{2}{c}{Pooled $\Delta d'$ [95\% CI]} & \multicolumn{2}{c}{Commit contradicts / relational \texttt{D+C$+$}} \\
\cmidrule(lr){2-3}\cmidrule(lr){4-5}\cmidrule(lr){6-7}
Model & all 500 & overlapping 431 & all 500 & overlapping 431 & all 500 & overlapping 431 \\
\midrule
Qwen2.5-VL-3B & $-1.50$ {\tiny[$-$2.01,$-$1.16]} & $-1.54$ {\tiny[$-$2.05,$-$1.14]} & $-1.77$ {\tiny[$-$2.12,$-$1.48]} & $-1.87$ {\tiny[$-$2.27,$-$1.53]} & $37/37$ / $2/30$ & $30/30$ / $2/28$ \\
Qwen2.5-VL-7B & $-0.16$ {\tiny[$-$0.30,$-$0.01]} & $-0.14$ {\tiny[$-$0.30,+0.03]} & $-0.58$ {\tiny[$-$0.78,$-$0.39]} & $-0.60$ {\tiny[$-$0.83,$-$0.38]} & $20/35$ / $3/32$ & $15/29$ / $2/30$ \\
InternVL3-8B & $-0.94$ {\tiny[$-$1.25,$-$0.67]} & $-0.94$ {\tiny[$-$1.25,$-$0.68]} & $-1.83$ {\tiny[$-$2.14,$-$1.54]} & $-1.89$ {\tiny[$-$2.24,$-$1.59]} & $33/34$ / $5/33$ & $29/30$ / $5/32$ \\
Phi-3.5-V & $-0.23$ {\tiny[$-$0.42,$-$0.05]} & $-0.21$ {\tiny[$-$0.40,+0.00]} & $-0.40$ {\tiny[$-$0.59,$-$0.20]} & $-0.38$ {\tiny[$-$0.59,$-$0.16]} & $36/36$ / $6/36$ & $33/33$ / $6/32$ \\
Molmo-7B-D & $-0.73$ {\tiny[$-$1.15,$-$0.41]} & $-0.78$ {\tiny[$-$1.34,$-$0.36]} & $-1.19$ {\tiny[$-$1.45,$-$0.95]} & $-1.26$ {\tiny[$-$1.57,$-$0.98]} & $24/31$ / $5/34$ & $20/26$ / $3/30$ \\
Qwen3-VL-8B & $-1.32$ {\tiny[$-$1.80,$-$1.01]} & $-1.55$ {\tiny[$-$2.14,$-$1.18]} & $-1.18$ {\tiny[$-$1.43,$-$0.97]} & $-1.71$ {\tiny[$-$2.12,$-$1.40]} & $26/27$ / $1/28$ & $26/27$ / $1/23$ \\
InternVL3.5-8B & $+0.20$ {\tiny[+0.04,+0.36]} & $+0.19$ {\tiny[+0.02,+0.38]} & $+0.24$ {\tiny[+0.05,+0.42]} & $+0.21$ {\tiny[+0.00,+0.43]} & $32/35$ / $0/33$ & $31/34$ / $0/28$ \\
\midrule
Pooled &  &  &  &  & $208/235$ (88.5\%) / $22/226$ (9.7\%) & $184/209$ (88.0\%) / $19/203$ (9.4\%) \\
\bottomrule
\end{tabular}}\\[4pt]
\setlength{\tabcolsep}{3pt}
\textbf{(b)}\ \fitto{0.96\textwidth}{\begin{tabular}{@{}c@{}}
\begin{tabular}{lccccc}
\toprule
 & \multicolumn{2}{c}{BC$+$BW / bound (\%)} & \multicolumn{3}{c}{$\Delta d'_\mathrm{pair}$ [95\% CI]} \\
\cmidrule(lr){2-3}\cmidrule(lr){4-6}
Model & Dir. & CoT & pooled $=$ MH & size-std.\ (terciles) & size-std.\ (quintiles) \\
\midrule
Qwen2.5-VL-3B & 68.4/74.4 & 33.2/52.4 & $-1.50$ [$-$2.01, $-$1.14] & $-1.76$ [$-$2.17, $-$1.45] & $-1.69$ [$-$2.06, $-$1.40] \\
Qwen2.5-VL-7B & 81.8/58.0$^{*}$ & 62.2/55.8$^{*}$ & $-0.16$ [$-$0.31, $-$0.01] & $-0.20$ [$-$0.52, +0.13] & $-0.32$ [$-$0.50, $-$0.09] \\
InternVL3-8B & 91.0/68.1$^{*}$ & 54.2/54.2 & $-0.94$ [$-$1.25, $-$0.69] & $-1.29$ [$-$1.55, $-$1.04] & $-1.29$ [$-$1.54, $-$1.07] \\
Phi-3.5-V & 61.6/53.4$^{*}$ & 48.0/51.4 & $-0.23$ [$-$0.41, $-$0.03] & $-0.23$ [$-$0.43, $-$0.03] & $-0.23$ [$-$0.44, $-$0.02] \\
Molmo-7B-D & 81.8/74.4$^{*}$ & 57.8/62.7 & $-0.73$ [$-$1.14, $-$0.39] & $-0.77$ [$-$1.26, $-$0.38] & $-0.76$ [$-$1.22, $-$0.38] \\
Qwen3-VL-8B & 59.8/73.1 & 47.4/53.5 & $-1.32$ [$-$1.75, $-$0.99] & $-1.28$ [$-$1.80, $-$0.97] & $-1.18$ [$-$1.62, $-$0.87] \\
InternVL3.5-8B & 48.0/50.7 & 47.0/52.1 & $+0.20$ [+0.03, +0.37] & $+0.20$ [+0.02, +0.39] & $+0.20$ [+0.02, +0.40] \\
\bottomrule
\end{tabular} \\[4pt]
\begin{tabular}{lccc}
\toprule
 & \multicolumn{3}{c}{$d'_\mathrm{pair}$ by size tercile, Dir.$\to$CoT ($\Delta$ [95\% CI])} \\
\cmidrule(lr){2-4}
Model & low ($<3.2\times$) & mid & high ($\ge 8.4\times$) \\
\midrule
Qwen2.5-VL-3B & 1.45$\to$0.86 ($-0.58$ [$-$1.19, $-$0.04]) & 1.98$\to$0.47 ($-1.52$ [$-$2.69, $-$0.94]) & 3.28$\to$0.11 ($-3.17$ [$-$3.52, $-$2.79]) \\
Qwen2.5-VL-7B & 0.56$\to$0.52 ($-0.05$ [$-$0.26, +0.19]) & 1.06$\to$0.59 ($-0.47$ [$-$0.74, $-$0.23]) & 2.07$\to$1.99 ($-0.08$ [$-$1.01, +0.86]) \\
InternVL3-8B & 1.16$\to$1.33 ($+0.18$ [$-$0.28, +0.74]) & 1.65$\to$0.68 ($-0.97$ [$-$1.57, $-$0.58]) & 3.36$\to$0.28 ($-3.08$ [$-$3.35, $-$2.80]) \\
Phi-3.5-V & 0.34$\to$0.39 ($+0.05$ [$-$0.23, +0.34]) & 0.89$\to$0.47 ($-0.41$ [$-$0.81, $-$0.05]) & 0.64$\to$0.31 ($-0.33$ [$-$0.71, +0.01]) \\
Molmo-7B-D & 1.50$\to$1.18 ($-0.32$ [$-$0.82, +0.14]) & 2.20$\to$1.76 ($-0.44$ [$-$1.43, +0.35]) & 2.66$\to$1.12 ($-1.54$ [$-$2.40, $-$0.87]) \\
Qwen3-VL-8B & 1.68$\to$1.33 ($-0.35$ [$-$0.90, +0.16]) & 2.16$\to$0.68 ($-1.48$ [$-$2.52, $-$0.92]) & 2.03$\to$0.03 ($-2.00$ [$-$3.06, $-$1.44]) \\
InternVL3.5-8B & 0.69$\to$0.84 ($+0.14$ [$-$0.08, +0.39]) & 0.08$\to$0.30 ($+0.21$ [$-$0.05, +0.50]) & $-$0.17$\to$0.09 ($+0.25$ [$-$0.14, +0.68]) \\
\bottomrule
\end{tabular} \\[2pt]
\parbox{0.97\linewidth}{\scriptsize \texttt{Synthetic Depth}, both object orders per image. BC$+$BW / bound: observed share of images answered differently under the two orders vs.\ its maximum under a common sensitivity, $100[\sigma(D/2)^2+\sigma(-D/2)^2]$ at the observed pair log-odds $D$; $^{*}$: excess CI $>0$ (image bootstrap), i.e.\ sensitivity varies across images. Pooled $=$ MH: pair-matched $d'_\mathrm{pair}$ over all images; the Mantel--Haenszel common odds ratio over any grouping of matched pairs is $\sum n_\mathrm{BC}/\sum n_\mathrm{BW}$, so size-stratified MH equals the pooled estimate. Size-std.: image-share-weighted mean of within-stratum $\ln[(n_\mathrm{BC}+0.5)/(n_\mathrm{BW}+0.5)]/1.702$ over visible-area-ratio terciles or quintiles of the 486 images with both shapes visible, weights fixed across arms, so \texttt{CoT} cannot lower it by moving discordant images between size strata. Direct high-tercile values with $n_\mathrm{BW}=0$ are the $+0.5$ cap (a lower bound): Qwen2.5-VL-3B 3.28, InternVL3-8B 3.36. Falls with CI $<0$: pooled 6/7, size-std.\ terciles 5/7, quintiles 6/7; by tercile 1/5/4. \texttt{CoT} $d'_\mathrm{pair}$ in the high tercile: Qwen2.5-VL-3B 0.11 [$-$0.25, 0.48] (top quintile 0.11 [$-$0.39, 0.65]); InternVL3-8B 0.28 [0.01, 0.55] (top quintile 0.31 [$-$0.03, 0.69]); Qwen3-VL-8B 0.03 [$-$0.23, 0.32] (top quintile $-$0.06 [$-$0.42, 0.31]). Image bootstrap ($2{,}000$).}
\end{tabular}}
\end{table*}

\subsection{Cue-factorial set and cue weights}
\label{sec:supp_cues}

The cue-factorial set crosses two size conditions (front/back area ratio $1.25$--$1.80$ or $0.55$--$0.80$) with two vertical conditions (front object lower or higher in the image), with $100$ base scenes per cell and an identity-swapped twin for each ($800$ images, $1{,}600$ prompts with both phrasings), so occlusion is the only valid cue. The main text reports a probit GLM fit (component $d' = 2\beta$, weak $\mathcal{N}(0, 10)$ prior, twin-pair bootstrap; no separation in any open model on the cue set; \cref{tab:supp_glm}a). A per-cell decomposition of pooled $d'$ into occlusion, vertical, and size components agrees with the GLM in sign for every component, within $0.34$ $d'$ for the occlusion component and within $0.53$ for the vertical component. The GLM occlusion loss has a CI below zero in all seven open models and survives Benjamini--Hochberg control with resampling $p$ in all seven; with Wald $p$ (bootstrap SE) it survives in six, as the \texttt{Qwen2.5-VL-7B} loss ($-0.34$ $[-0.76, -0.07]$) misses ($q = 0.053$ vs.\ $0.020$). \texttt{DeepSeek-V4.1-Flash} is run with \texttt{reasoning\_effort=none} in both arms (the analysis uses the $1{,}566$ prompts answered in both arms) and is near separation under \texttt{Direct} ($89.5\%\to90.7\%$ accuracy under \texttt{CoT}), so its occlusion CI is wide.

\paragraph{Scale-free check.} If \texttt{CoT} only added image-independent variance, all probit cue weights would shrink by one factor; a common factor fit over the three cue weights gives $s = 0.27$--$0.55$. \cref{tab:supp_cue_ame}b therefore reports each cue's average marginal effect (AME) on the probability that the answer calls the queried subject the front shape (same probit model, points; $1{,}000$ twin-pair bootstrap draws, \texttt{Direct} and \texttt{CoT} resampled jointly), which is our primary cue-effect scale (\cref{tab:profile,tab:supp_e31,tab:supp_e43,tab:supp_e48,tab:supp_written_ame}). Because occlusion is the only valid cue and the set is balanced, the occlusion contrast equals twice the accuracy minus one, so on this scale the occlusion effect falls only where accuracy falls. In \texttt{Molmo-7B-D} it falls by $26.1$ $[21.6, 30.4]$ points ($76.2\to50.1$; $66\%$ retained) and the height effect by $5.9$ ($14.4\to8.4$); in \texttt{Qwen3-VL-8B} the occlusion effect barely moves ($56.7\to54.6$, $-2.1$ $[-6.7, +2.4]$) while the height effect flips ($+26.0\to-16.5$). In three shortcut-reliant models the occlusion effect rises ($+6.8$ to $+9.6$, CIs above zero) although their probit occlusion weights fall, and the height effect falls by $24.6$--$53.6$ points in all four. The probit $\Delta d'$ shift in \cref{tab:supp_glm}a is therefore a smaller occlusion effect on the answer in \texttt{Molmo-7B-D}, but in \texttt{Qwen3-VL-8B} and the shortcut-reliant models it reflects a change of the latent scale (added variance or a weaker height term), not a smaller occlusion effect.

\paragraph{First claims and the twins.} First \texttt{CoT} claims (rule parser v3, which skips symmetric verbs such as ``overlaps'' and handles negation) name the lower shape in $40.4\%$ (\texttt{Qwen3-VL-8B}) to $65.1\%$--$66.4\%$ (\texttt{InternVL3.5-8B}, \texttt{Qwen2.5-VL-7B}) of parsed claims, so the lower-is-closer prior is adopted in writing by some models and reversed by others; \texttt{Qwen3-VL-8B}'s ``on top'' claims name the upper shape in $88.6\%$ of cases. A logistic model of which shape a claim names as front gives a question-affirming term that is positive in six models ($+0.21$ to $+1.22$, CIs above zero) and negative in \texttt{Qwen2.5-VL-7B} ($-0.61$ $[-0.74, -0.49]$); on $560$ judge-labelled claims it stays above zero under judge, rule-v2, and rule-v3 labels in those six (marginally in \texttt{InternVL3.5-8B}) and is not robust in \texttt{Qwen2.5-VL-7B}, so we claim affirmation only for the six. On the original twins, whose front shape is never drawn larger, pooled $d'$ falls under \texttt{CoT} in \texttt{Phi-3.5-Vision}, \texttt{Molmo-7B-D}, \texttt{Qwen3-VL-8B}, and \texttt{Qwen3-VL-8B-Thinking} and rises in \texttt{InternVL3-8B} ($+0.43$), \texttt{InternVL3.5-8B} ($+0.25$), and \texttt{Qwen2.5-VL-7B} ($+0.21$; not significant after FDR control); these three have a strong \texttt{Direct} lower-is-closer prior, which the balanced front-above/below design turns into noise, so diluting the prior raises marginal $d'$ without adding occlusion signal.

\begin{table*}[t]
\centering
\caption{\textbf{Cue-factorial decomposition.} \textbf{(a)} Probit GLM decomposition on the cue-factorial set (component $d'$ with $95\%$ twin-pair bootstrap CIs) and AUC ($^\dagger$\texttt{reasoning\_effort=none} in both arms). \textbf{(b)} Scale-free cue effects (primary scale, matching \cref{tab:profile,tab:supp_written_ame}): average marginal effect of each $\pm1$-coded cue on the probability that the answer calls the queried subject the front shape (points; probit model of (a)), under \texttt{Direct} and \texttt{CoT}, with the paired change and its $95\%$ twin-pair bootstrap CI ($1{,}000$ draws).}
\label{tab:supp_glm}
\label{tab:supp_cue_ame}
\scriptsize
\setlength{\tabcolsep}{2pt}
\textbf{(a)}\ \fitto{0.96\textwidth}{\begin{tabular}{lcccccc}
\toprule
Model & Direct occl./vert./size & CoT$-$Dir.\ occlusion & CoT$-$Dir.\ vertical & CoT$-$Dir.\ size & CoT vertical & AUC Dir.$\to$CoT \\
\midrule
Qwen2.5-VL-3B & +0.69/+4.08/+0.07 & $-0.51$ [$-$0.81, $-$0.24] & $-2.99$ [$-$3.47, $-$2.69] & $-0.15$ [$-$0.43, +0.12] & $+1.08$ [+0.99, +1.22] & 0.587$\to$0.554 \\
Qwen2.5-VL-7B & +0.91/+3.86/$-$0.13 & $-0.34$ [$-$0.76, $-$0.07] & $-2.19$ [$-$2.60, $-$1.94] & $+0.20$ [$-$0.06, +0.49] & $+1.68$ [+1.52, +1.85] & 0.615$\to$0.579 \\
InternVL3-8B & +1.81/+3.05/+0.13 & $-0.56$ [$-$0.92, $-$0.32] & $-2.44$ [$-$2.81, $-$2.21] & $+0.12$ [$-$0.07, +0.31] & $+0.61$ [+0.48, +0.76] & 0.759$\to$0.796 \\
Phi-3.5-V & +0.72/+0.91/$-$0.50 & $-0.31$ [$-$0.49, $-$0.15] & $-0.62$ [$-$0.79, $-$0.45] & $+0.50$ [+0.34, +0.66] & $+0.29$ [+0.18, +0.42] & 0.639$\to$0.568 \\
Molmo-7B-D & +2.72/+0.85/$-$0.10 & $-1.18$ [$-$1.39, $-$1.01] & $-0.55$ [$-$0.72, $-$0.33] & $+0.08$ [$-$0.13, +0.24] & $+0.30$ [+0.17, +0.45] & 0.965$\to$0.799 \\
Qwen3-VL-8B & +2.79/+1.49/$-$0.43 & $-1.01$ [$-$1.36, $-$0.70] & $-2.13$ [$-$2.38, $-$1.91] & $-0.11$ [$-$0.31, +0.10] & $-0.64$ [$-$0.83, $-$0.47] & 0.921$\to$0.822 \\
InternVL3.5-8B & +1.49/+4.35/$-$0.38 & $-0.60$ [$-$1.07, $-$0.32] & $-2.19$ [$-$2.81, $-$1.89] & $+0.03$ [$-$0.26, +0.29] & $+2.16$ [+1.99, +2.34] & 0.639$\to$0.656 \\
DeepSeek-V4.1-Flash$^\dagger$ & +4.32/+1.14/$-$0.01 & $-0.05$ [$-$1.27, +1.24] & $-0.23$ [$-$0.48, +0.01] & $+0.25$ [+0.04, +0.47] & $+0.90$ [+0.71, +1.17] & -- \\
\bottomrule
\end{tabular}}\\[4pt]
\textbf{(b)}\ \fitto{0.96\textwidth}{\begin{tabular}{lccccccc}
\toprule
 & \multicolumn{3}{c}{Occlusion} & \multicolumn{3}{c}{Height (lower is closer)} & Size \\
\cmidrule(lr){2-4}\cmidrule(lr){5-7}\cmidrule(lr){8-8}
Model & \texttt{Dir.} & \texttt{CoT} & \texttt{CoT}$-$\texttt{Dir.} & \texttt{Dir.} & \texttt{CoT} & \texttt{CoT}$-$\texttt{Dir.} & \texttt{CoT}$-$\texttt{Dir.} \\
\midrule
Qwen2.5-VL-3B & $4.7$ & $6.0$ & $+1.3$ {\scriptsize$[-2.8, +5.5]$} & $92.4$ & $38.7$ & $-53.6$ {\scriptsize$[-57.5, -49.6]$} & $-3.3$ {\scriptsize$[-7.3, +0.8]$} \\
Qwen2.5-VL-7B & $6.5$ & $13.3$ & $+6.8$ {\scriptsize$[+3.1, +10.8]$} & $91.5$ & $41.4$ & $-50.1$ {\scriptsize$[-54.1, -46.3]$} & $+2.6$ {\scriptsize$[-1.2, +6.4]$} \\
InternVL3-8B & $27.3$ & $36.9$ & $+9.6$ {\scriptsize$[+4.9, +14.3]$} & $70.0$ & $17.1$ & $-52.9$ {\scriptsize$[-57.8, -48.0]$} & $+4.6$ {\scriptsize$[-0.4, +8.9]$} \\
Phi-3.5-V & $19.2$ & $14.4$ & $-4.9$ {\scriptsize$[-10.0, +0.1]$} & $24.7$ & $10.4$ & $-14.2$ {\scriptsize$[-19.1, -9.1]$} & $+13.4$ {\scriptsize$[+8.4, +18.7]$} \\
Molmo-7B-D & $76.2$ & $50.1$ & $-26.1$ {\scriptsize$[-30.4, -21.6]$} & $14.4$ & $8.4$ & $-5.9$ {\scriptsize$[-10.2, -2.0]$} & $+1.1$ {\scriptsize$[-3.4, +5.1]$} \\
Qwen3-VL-8B & $56.7$ & $54.6$ & $-2.1$ {\scriptsize$[-6.7, +2.4]$} & $26.0$ & $-16.5$ & $-42.5$ {\scriptsize$[-47.5, -37.7]$} & $-6.7$ {\scriptsize$[-11.0, -2.1]$} \\
InternVL3.5-8B & $10.9$ & $20.0$ & $+9.1$ {\scriptsize$[+5.4, +12.8]$} & $87.8$ & $63.2$ & $-24.6$ {\scriptsize$[-28.0, -21.0]$} & $-4.4$ {\scriptsize$[-8.1, -0.6]$} \\
\bottomrule
\end{tabular}}
\end{table*}

\subsection{Sentence judges}
\label{sec:supp_judge}

For each model we take the stored greedy rationale of every \texttt{Synthetic Depth} item in the \texttt{D+C$-$} and \texttt{D+C$+$} groups and extract the first sentence that names both shapes together with an occlusion or depth verb (\emph{overlap}, \emph{cover}, \emph{on top of}, \emph{obscure}, \emph{block}, \emph{in front of}, or a passive form such as \emph{covered by}, \emph{behind}); sentences that pose a question (\emph{whether}, \emph{if the}, \emph{determine}, \emph{asks}) are skipped. We sample up to $40$ parsed sentences per model and group (seed $0$) and send each sentence together with the image to two judges, Claude Opus 5.5 (\texttt{reasoning\_effort=high}) and GPT-6-Astra, with the prompt:
\begin{quote}\small
Here is an image and one sentence that a model wrote about it: ``\{sentence\}''. Consider only what the sentence claims about which shape is in front of, on top of, covering, or overlapping the other. Does that claim match the image? Reply with exactly one word: correct, incorrect, or unclear.
\end{quote}
\cref{tab:supp_judge} reports the share judged \emph{incorrect} among sentences a judge does not label \emph{unclear}, with Wilson $95\%$ intervals. The two judges agree on $96.1\%$ of the sentences both label ($\kappa = 0.92$). A rule-based parse agrees with them on $75.3\%$ ($\kappa = 0.51$) and $79.2\%$ ($\kappa = 0.59$); it misses negations such as ``not in front of''. \cref{fig:teaser}b plots the two-judge consensus ($58.3\%$--$97.4\%$ on \texttt{D+C$-$}, $5.4\%$--$25.7\%$ on \texttt{D+C$+$}, $11.5\%$--$35.9\%$ on \texttt{D$-$C$+$}). In a blind cross-check of $120$ first claims ($60$ from \texttt{Synthetic Depth}, $60$ from natural images) by three model reviewers (Claude Opus 5.5, Gemini-4-Argon, Gemini-4-Carbon), the majority label agrees with the Claude Opus 5.5 judge on $94.3\%$ ($\kappa = 0.89$) and with the GPT-6-Astra judge on $95.4\%$ ($\kappa = 0.91$); this is a model cross-check, not human validation. On a stratified $30$-sentence sample ($12$ \texttt{D+C$-$} commit, $12$ matched relational \texttt{D+C$+$}, and $6$ rescue sentences; seven models), human labels (one annotator, the author, blind to model, \texttt{D$\pm$C$\pm$} cell, and judge labels) agree with the judge consensus on $23$ of $24$ sentences that both label as consistent or contradicting ($\kappa = 0.91$; judge--judge $\kappa = 0.84$ on the same sample). The same two models also serve as text-only judges that extract which relation a rationale states; their agreement is given with each analysis.

\begin{table}[t]
\centering
\caption{\textbf{Share (\%) of first two-object spatial claims that contradict the image} (\texttt{Synthetic Depth}). Opus / GPT: Claude Opus 5.5 and GPT-6-Astra, both given the image; $n$ (judged) is the number of sampled sentences (up to $40$), and rates exclude sentences a judge labels unclear. Rule (all): rule-based parse over all $n$ (all) parsed sentences, also reported for \texttt{InternVL3.5-8B} (\cref{fig:teaser}b was run on the initial six models). Brackets are Wilson $95\%$ intervals (Rule: image bootstrap). Each image contributes at most one claim per model and group, so all intervals are image-level. Rates exclude unclear labels: of the $673$ sampled claims, $656$ (Opus), $645$ (GPT), and $613$ (consensus) are judgeable; per \texttt{D+C$-$} cell $38$--$40$, $38$--$40$, and $35$--$39$ of $40$. \texttt{D$-$C$+$}: \texttt{Direct} wrong, \texttt{CoT} right.}
\label{tab:supp_judge}
\scriptsize
\setlength{\tabcolsep}{2pt}
\fitto{\columnwidth}{\begin{tabular}{llccc}
\toprule
Model & Labeler & D+C$-$ & D+C$+$ & D$-$C$+$ \\
\midrule
Qwen2.5-VL-3B & Opus & $95$\,{\scriptsize[83,99]} & $15$\,{\scriptsize[7,30]} & $25$\,{\scriptsize[11,47]} \\
 & GPT & $97$\,{\scriptsize[87,100]} & $11$\,{\scriptsize[4,24]} & $12$\,{\scriptsize[3,34]} \\
 & Consensus & $97$\,{\scriptsize[86,100]} & $11$\,{\scriptsize[4,25]} & $12$\,{\scriptsize[4,36]} \\
 & $n$ (judged) & 40 & 40 & 21 \\
 & Rule (all) & $84$\,{\scriptsize[78,90]} & $14$\,{\scriptsize[10,18]} & $48$\,{\scriptsize[29,71]} \\
 & $n$ (all) & 144 & 258 & 21 \\
\addlinespace[2pt]
Qwen2.5-VL-7B & Opus & $68$\,{\scriptsize[52,80]} & $31$\,{\scriptsize[19,46]} & $35$\,{\scriptsize[22,50]} \\
\bottomrule
\end{tabular}}
\end{table}

\subsection{Held-out pool and frozen protocol}
\label{sec:supp_heldout}

The held-out pool has $767$ binary items: \texttt{BLINK} val Spatial Relation ($143$) and Relative Depth ($124$), and five \texttt{3DSRBench} COCO categories with $100$ items each (one item per image, seed $0$). For each model, before the held-out \texttt{CoT}, vote, and description runs, we wrote one time-stamped file (released with the outputs) that contains (i) the \texttt{Direct}-margin gate thresholds $\tau$ selected by $5$-fold cross-validation on the main pool, (ii) item-level and constant-$\bar{p}$ predictions of $4$-sample majority voting ($T=0.7$) for every held-out task and for the pooled set, computed from held-out \texttt{Direct} probabilities, and (iii) SHA-256 hashes of the \texttt{Direct} outputs, the $\tau$ file, and the code. These two predictions (gate transfer and vote accuracy) are the only frozen endpoints; the \texttt{BLINK Relative Depth} comparison was not a pre-specified endpoint. The item-level vote prediction matches observed voting within $0.3$--$1.5$ points in six models and $2.1$ points in \texttt{Molmo-7B-D}, while the constant-$\bar{p}$ baseline overestimates voting by up to $8.5$ points; the frozen gate gives $-0.8$ to $+2.1$ over \texttt{Direct} with every CI including zero.

\paragraph{Cross-model notes.} Per-task accuracies and main cross-model differences are reported in \cref{sec:consequences} (\cref{tab:supp_heldout}). Cross-model bootstrap intervals resample items with the set of models fixed, so they describe these models. On the post hoc $324$-item depth-valid subset, \texttt{CoT} $-$ \texttt{Direct} is $-3.5$ $[-6.1, -1.0]$ over eight open models and \texttt{CoT} $-$ \texttt{Filler+turn} is $-3.4$ $[-6.1, -0.6]$ over the seven instruct models (no single per-model CI excludes zero); \texttt{3DSRBench} closer ($-1.5$) and in front of ($-1.1$) do not change reliably, and below-chance \texttt{Direct} accuracy on \texttt{3DSR Left} ($28\%$--$41\%$) suggests that models answer in the camera's frame rather than the object's own frame.

\subsection{Closed API details}
\label{sec:supp_closed}

We query six closed models through a commercial OpenAI-compatible API with \texttt{reasoning\_effort=high}: \texttt{Claude Opus 5.5}, \texttt{GPT-6-Astra}, \texttt{Gemini-4-Argon}, \texttt{Qwen3.8-Max}, \texttt{HY4}, and \texttt{DeepSeek-V4.1-Flash} (results in \cref{sec:frontier}, \cref{tab:supp_closed,tab:supp_e50,tab:supp_e61}). \texttt{Direct} appends ``Answer with the option's letter only.''; \texttt{CoT} asks the model to describe what it sees, reason, and end with ``Answer: (X)'' (\cref{sec:supp_prompts}). Because these APIs return no option log-probabilities, we report answer-level metrics parsed from each reply, queried once at the provider's default temperature. \texttt{GPT-6-Astra} has $122$ \texttt{BLINK Relative Depth} items because the provider's content-safety filter rejected two images (HTTP 400) in both conditions. On the hard occlusion set ($144$ base pairs plus twins, $576$ prompts per condition across three difficulty levels with thin overlap slivers, low contrast, distractor shapes, blur, noise, and smaller objects), hard-set accuracies count failed replies as wrong (\texttt{HY4} $67.9\%$; $69.2\%$ on items with both replies). The \texttt{Qwen3-VL} $2$B--$32$B scale ladder is reported in \cref{sec:open_cohorts} (\cref{tab:supp_r29_ladder}).

\section{Analysis 1: Not Format or Forgetting}
\label{sec:supp_a1}

\cref{tab:supp_tasks} gives the task breakdown of \cref{tab:behavioral_main}, including the paired difference between \texttt{CoT} and turn-matched filler.

\paragraph{Option-letter prior, re-prompting, and re-shown images.} \cref{tab:supp_prior}a reports the share of ``A'' predictions in the original five-model runs with top-$20$ readout. Same-turn filler lowers the ``A'' share in \texttt{Qwen2.5-VL-7B}, and closing the turn restores the \texttt{Direct} rate. Repeating the question in the first turn (\texttt{Direct + restate}) shifts pooled spatial accuracy even without reasoning text: relative to \texttt{Direct}, accuracy changes by $-3.9$ $[-5.1, -2.8]$ points in \texttt{Qwen2.5-VL-3B}, $-0.4$ in \texttt{Qwen2.5-VL-7B}, $-0.5$ in \texttt{InternVL3-8B}, $-1.9$ in \texttt{Phi-3.5-Vision}, and $+2.6$ in \texttt{Molmo-7B-D}. \texttt{Molmo-7B-D} was not run with re-shown images on the main pool (\cref{tab:behavioral_main}), and \texttt{Qwen3-VL-8B-Thinking} omits same-turn \texttt{Filler($|R|$)} and extra-turn \texttt{restate}/\texttt{reimage} arms because its native \texttt{<think>...\textless/think>} trace closes before \texttt{Answer: (} (so only turn-closed \texttt{Filler($|T|$)+turn} applies). On the cue set we read \texttt{Molmo-7B-D}'s stored \texttt{CoT} rationales with HuggingFace \texttt{Transformers} in four arms: \texttt{Direct} $87.1\%$ (vs.\ $88.2\%$ under \texttt{vLLM} V1 in \cref{tab:supp_heuristics}), \texttt{CoT} $75.1\%$, \texttt{CoT+turn} $75.1\%$, \texttt{CoT+reimage} $76.4\%$ (reimage$-$\texttt{CoT} $+1.3$ $[+0.8, +1.9]$), so a re-shown image does not explain \texttt{Molmo-7B-D}'s loss either. On non-spatial controls in the original five-model runs, \texttt{CoT} changes accuracy by $-6.3$ to $+10.6$ points on \texttt{CV-Bench 2D Count} ($788$ items) and by $-3.2$ to $+3.4$ points on \texttt{MMStar}~\citep{chen2024mmstar} ($500$ items).

\paragraph{Answer churn.} \cref{tab:supp_churn}b crosses \texttt{Direct} and \texttt{CoT} correctness on each item. \texttt{CoT} changes $7.3\%$ (\texttt{Qwen3-VL-8B-Thinking}) to $32.9\%$ (\texttt{Phi-3.5-Vision}) of the $2{,}850$ answers, and the net losses of \cref{tab:behavioral_main} are differences between large flows (\texttt{Molmo-7B-D}: $421$ answers lost and $373$ gained, net $-1.7$ points). On \texttt{Synthetic Depth} ($n = 500$), \texttt{D+C$-$} outnumbers \texttt{D$-$C$+$} in seven of eight models, a ratio of $1.2$--$7.5$; \texttt{InternVL3.5-8B} is reversed ($48/72$). The \texttt{D$\pm$C$\pm$} groups of the error-onset, commit, rescue, and steering analyses come from the original runs' top-$20$ letter scores (near-tie differences in \cref{sec:supp_engine}).

\begin{table*}[t]
\centering
\caption{\textbf{Option-letter prior and answer churn on the $2{,}850$-item binary spatial pool.} \textbf{(a)} Share of ``A'' predictions (\%; gold ``A'' share $50.5\%$), original top-$20$ readout runs (\texttt{Fil.+t}: \texttt{Filler($|R|$)+turn}; \texttt{F512}: same-turn \texttt{Filler-512}). \textbf{(b)} Answer churn: \texttt{Direct}$\times$\texttt{CoT} correctness cells (item counts; net and churn in points / \%).}
\label{tab:supp_prior}
\label{tab:supp_churn}
\scriptsize
\setlength{\tabcolsep}{3pt}
\textbf{(a)}\ \begin{tabular}{lccccc}
\toprule
\textbf{Model} & \textbf{Dir.} & \textbf{CoT} & \textbf{Fil.} & \textbf{Fil.+t} & \textbf{F512} \\
\midrule
Qwen2.5-VL-3B & $55.6$ & $62.6$ & $48.6$ & $51.9$ & $54.3$ \\
Qwen2.5-VL-7B & $48.6$ & $49.8$ & $\mathbf{29.9}$ & $\mathbf{47.5}$ & $46.4$ \\
InternVL3-8B & $50.0$ & $52.7$ & $46.6$ & $49.7$ & $47.6$ \\
Phi-3.5-V & $46.3$ & $58.2$ & $57.6$ & $46.1$ & $48.5$ \\
Molmo-7B-D & $60.8$ & $53.8$ & $54.2$ & $52.0$ & $49.5$ \\
\bottomrule
\end{tabular}\\[4pt]
\textbf{(b)}\ \fitto{0.96\textwidth}{\begin{tabular}{lccccccccc}
\toprule
 & \multicolumn{7}{c}{2,850-item pool, \cref{tab:behavioral_main} readout} & \texttt{Synthetic Depth} ($n=500$) & Sec.~5 cells \\
\cmidrule(lr){2-8}\cmidrule(lr){9-9}\cmidrule(lr){10-10}
Model & D+C$-$ & D$-$C+ & D+C+ & D$-$C$-$ & net (pts) & Tab.~1 $\Delta$ & churn (\%) & D+C$-$ / D$-$C+ (churn \%, net) & D+C$-$ / D$-$C+ \\
\midrule
Qwen2.5-VL-3B & 470 & 220 & 1818 & 342 & $-8.8$ & $-8.8$ & 24.2 & 158 / 21 (35.8, $-27.4$) & 458 / 212 \\
Qwen2.5-VL-7B & 285 & 175 & 2164 & 226 & $-3.9$ & $-3.9$ & 16.1 & 81 / 43 (24.8, $-7.6$) & 284 / 171 \\
InternVL3-8B & 264 & 118 & 2291 & 177 & $-5.1$ & $-5.1$ & 13.4 & 153 / 32 (37.0, $-24.2$) & 267 / 121 \\
Phi-3.5-V & 600 & 338 & 1578 & 334 & $-9.2$ & $-9.2$ & 32.9 & 115 / 94 (41.8, $-4.2$) & 601 / 335 \\
Molmo-7B-D & 421 & 373 & 1702 & 354 & $-1.7$ & $-1.7$ & 27.9 & 101 / 31 (26.4, $-14.0$) & 421 / 373 \\
Qwen3-VL-8B & 190 & 90 & 2406 & 164 & $-3.5$ & $-3.5$ & 9.8 & 117 / 32 (29.8, $-17.0$) & 190 / 90 \\
\end{tabular}}
\end{table*}

\paragraph{How much rests on \texttt{Synthetic Depth}.} \cref{tab:supp_e58b} removes the size-confounded set. On the main pool without it, \texttt{CoT}$-$\texttt{Direct} averages $-2.1$ $[-2.7, -1.6]$ points over the eight open models (all items: $-3.9$); three models lose significantly (\texttt{Qwen2.5-VL-3B}, \texttt{Qwen2.5-VL-7B}, \texttt{Phi-3.5-Vision}, the last mostly on \texttt{CV-Bench} 3D Depth) and one gains significantly (\texttt{InternVL3.5-8B} $+1.9$; \texttt{Qwen3-VL-8B-Thinking} $+0.9$ $[+0.0, +1.7]$ is not significant). Without the seven \texttt{Synthetic Depth} points, the two-channel correlation falls from $0.93$ to $0.82$ (conclusion kept) and from $0.89$ to $0.51$ (conclusion removed), and model-cluster CIs for $r$ include zero. With the conclusion removed, agreement with the last body relation is $78.7\%$--$99.6\%$ when it is true and $66.5\%$--$98.8\%$ when it is false (lowest: \texttt{InternVL3-8B}, cue set). On natural \texttt{CV-Bench 3D Depth} images the per-model CI of the accuracy change excludes zero in three of eight models (\texttt{Qwen2.5-VL-3B} $-8.0$, \texttt{Phi-3.5-Vision} $-23.0$, \texttt{Qwen3-VL-8B} $-2.8$); the cross-model mean is $-4.0$ $[-5.2, -2.8]$, and without \texttt{Phi-3.5-Vision} it is $-1.3$ $[-2.5, -0.1]$.

\section{Analysis 2: Written Relations Go Wrong}
\label{sec:supp_a2}

\paragraph{Commit sentences.} In \cref{fig:error_onset}a the per-model mean readout is $0.67$--$0.86$ at $j^\ast-1$ and $0.06$--$0.21$ at $j^\ast$. Against a random sentence of the same rationale, the excess of relational commits is $+18.6$ to $+47.4$ points under a uniform null and $+9.9$ to $+32.2$ under a position-matched null, significant in every model ($p \le 0.0012$). On all $500$ \texttt{Synthetic Depth} images, we sent $j^\ast$ itself, with the preceding sentence for coreference, to the two image-conditioned judges: $40$ \texttt{D+C$-$} commits per model in seven models, and $40$ relational \texttt{D+C$+$} comparator sentences per model nearest the same relative position. The consensus calls $88.5\%$ $[83.8, 92.0]$ of the judgeable \texttt{D+C$-$} commit sentences incorrect ($235/280$ judgeable; per model $57\%$ in \texttt{Qwen2.5-VL-7B}, $77\%$ in \texttt{Molmo-7B-D}, $91\%$--$100\%$ in the other five) vs.\ $9.7\%$ of judgeable relational ($226/280$) \texttt{D+C$+$} sentences; per model, every \texttt{D+C$-$} lower bound ($50\%$--$92.5\%$) exceeds the relational comparator's upper bound ($17.5\%$--$32.5\%$). On the $431$ overlapping images the rates are $88.0\%$ vs.\ $9.4\%$ (\cref{tab:supp_syn431}; per-model $52\%$--$100\%$ vs.\ $0\%$--$19\%$ in \cref{tab:profile}). Of the $280$ audited commits, $226$ ($80.7\%$) have a \texttt{Direct} answer that is correct under both object orders; on these the consensus calls $86.0\%$ $[80.3, 90.3]$ incorrect against $12.1\%$ $[8.0, 17.7]$ of the restricted comparator, so the result does not rest on items that \texttt{Direct} gets right only through a response bias. On the cue set we judged every \texttt{D+C$-$} commit and one relational sentence from every \texttt{D+C$+$} rationale: the consensus calls $184/202 = 91.1\%$ $[86.4, 94.3]$ of \texttt{Qwen3-VL-8B}'s judgeable commits incorrect against $66/782 = 8.4\%$ $[6.7, 10.6]$ of comparators, and $68/182 = 37.4\%$ $[30.7, 44.6]$ of \texttt{Molmo-7B-D}'s against $56/1058 = 5.3\%$ $[4.1, 6.8]$ (pooled: $252/384 = 65.6\%$ vs.\ $122/1{,}840 = 6.6\%$); $72$ of \texttt{Molmo-7B-D}'s $270$ commits are conclusion sentences without a checkable relation, and its commits are false on $53/99 = 53.5\%$ of ``in front of'' and $15/80 = 18.8\%$ of ``behind'' questions. A prompt that asks whether a sentence uses a depth or a vertical reading of ``on top'' calls $195/202 = 96.5\%$ and $63/177 = 35.6\%$ of these commits false, so ``on top'' does not inflate the audit. In $120$ \texttt{Qwen3-VL-8B} cue-set rationales, the earliest consensus image-contradicting relational sentence precedes $j^\ast$ in $70 = 58.3\%$ $[49.4, 66.8]$ (mean distance $2.5$ sentences). Re-read with a uniform gray image and the same text, the gray $j^\ast$, where it exists, equals the real one in $59/103 = 57.3\%$ (\texttt{Qwen3-VL-8B}) and $63/77 = 81.8\%$ (\texttt{Molmo-7B-D}) of rationales: where the readout gives in is mainly a property of the text, and how high it stands before that point depends on the image. As a mirror, rescue sentences on \texttt{D$-$C$+$} items are relational in $71\%$--$90\%$ of cases and judged consistent with the image in $81.7\%$ $[75.3, 86.7]$, against $17.9\%$ $[12.8, 24.5]$ of matched relational sentences in \texttt{D$-$C$-$} rationales. On natural \texttt{CV-Bench 3D Depth} rationales, the first claim naming both marked objects with a depth word is judged wrong in $66.7\%$--$94.9\%$ of \texttt{D+C$-$} versus $0.0\%$--$6.9\%$ of \texttt{D+C$+$} rationales (two-judge consensus; agreement $95.1\%$, $\kappa = 0.90$).

\paragraph{False sentences before $j^\ast$.} Of 120 judged \texttt{Qwen3-VL-8B} cue-set \texttt{D+C$-$} rationales, 58.3\% [49.6, 66.9] contain an image-contradicting relational sentence $k$ before $j^\ast$ (\cref{tab:supp_jstar_condition}). Writing $k$ leaves the readout in place: $P(\text{gold})$ changes by $-$0.4 points at $k$ vs.\ $-$90.6 at $j^\ast$, and a twin image still moves it by 52.5 points after $k$ but by $-$0.5 after $j^\ast$. $k$ names the same front shape as $j^\ast$ (21/22 where both parse), mostly as overlap (depth word 11\% vs.\ 71\%; inference marker 6\% vs.\ 57\%). Wording alone does not commit: 37 parsed pre-$j^\ast$ false sentences with an explicit front/behind predicate leave $P(\text{gold})$ at 0.98. Rule-parsed full sets (247; \texttt{Qwen3-VL-32B} 150) agree; in \texttt{Molmo-7B-D} earlier false sentences are rare (10\%). The commit sentence is thus where the false relation starts to drive the readout, not necessarily the first false sentence.

\begin{table*}[t]
\centering
\scriptsize
\setlength{\tabcolsep}{2.8pt}
\caption{\textbf{False relational sentences before the commit sentence $j^\ast$} (\texttt{D+C$-$} rationales with a commit). $k$: earliest sentence before $j^\ast$ that contradicts the image (judges: two image-conditioned judges, consensus, on every relational sentence before $j^\ast$ of a sample; parser: rule-parsed depth claim naming the occluded shape as front, all rationales). $P$: forced-readout $P(\text{gold})$ after the sentence; $\Delta P$: its change (points) at $k$ and at $j^\ast$ (same rationales). $P\ge.5$ just before $j^\ast$ holds by definition, so $P$ after $k$ is only partly informative; the twin-image effect (readout under the own image minus under the identity-swapped twin, \cref{tab:supp_e57}) is not constrained by the definition. Depth: share of sentences with an explicit depth word (in front, behind, closer, farther). Last columns: $j^\ast$ judged false when an earlier false sentence exists vs.\ not. Image-cluster bootstrap $95\%$ CIs ($2{,}000$ draws).}
\label{tab:supp_jstar_condition}
\begin{tabular}{lll r c c rr ccc cc cc}
\toprule
 & & & & & & \multicolumn{2}{c}{$\Delta P$} & \multicolumn{3}{c}{Twin-image effect} & \multicolumn{2}{c}{Depth word} & \multicolumn{2}{c}{$j^\ast$ false} \\
\cmidrule(lr){7-8}\cmidrule(lr){9-11}\cmidrule(lr){12-13}\cmidrule(lr){14-15}
Model & Set & Labels & $n$ & Earlier false $k$ & $P$ after $k$ & at $k$ & at $j^\ast$ & before $k$ & after $k$ & after $j^\ast$ & $k$ & $j^\ast$ & earlier & none \\
\midrule
Qwen3-VL-8B & cue set & judges & 120 & 58.3 {\tiny[49.6, 66.9]} & 0.92 & $-$0.4 & $-$90.6 & 54.8 & 52.5 & $-$0.5 & 11.4 & 71.4 & 85.3 & 66.0 \\
Qwen3-VL-8B & cue set & parser & 247 & 45.7 {\tiny[39.3, 52.3]} & 0.93 & $-$2.8 & $-$92.0 & 47.1 & 41.5 & $-$0.0 & 11.5 & 87.6 & 85.2 & 71.9 \\
Qwen3-VL-32B & cue set & parser & 150 & 37.3 {\tiny[30.0, 45.0]} & 0.97 & $-$1.6 & $-$96.9 & 83.2 & 89.3 & 0.3 & 25.0 & 98.2 & 90.7 & 88.6 \\
Qwen3-VL-8B & twins & judges & 40 & 47.5 {\tiny[30.8, 64.1]} & 0.92 & $-$6.2 & $-$83.6 & -- & -- & -- & 5.3 & 78.9 & 84.2 & 78.9 \\
Molmo-7B-D & cue set & judges & 40 & 10.0 {\tiny[2.5, 20.0]} & 0.66 & +6.3 & $-$55.0 & -- & -- & -- & 25.0 & 75.0 & 75.0 & 40.6 \\
Qwen3-VL-8B & Syn.\ Depth & judges & 40 & 30.0 {\tiny[15.0, 45.0]} & 0.97 & +5.1 & $-$92.5 & -- & -- & -- & 8.3 & 91.7 & 58.3 & 67.9 \\
Molmo-7B-D & Syn.\ Depth & judges & 40 & 15.0 {\tiny[5.0, 27.5]} & 0.59 & $-$14.4 & $-$42.3 & -- & -- & -- & 0.0 & 83.3 & 60.0 & 67.7 \\
\bottomrule
\end{tabular}
\end{table*}

\paragraph{Sentence-level pipeline on \texttt{Qwen3-VL-32B} and decomposition notes.} \cref{tab:supp_e57} gives the full pre-specified sentence-level pipeline on \texttt{Qwen3-VL-32B} next to \texttt{Qwen3-VL-8B} (summarized in \cref{sec:error_onset,sec:answer_step}). Two technical checks supplement \cref{tab:supp_e57,tab:supp_e54a}: (i) on \texttt{Qwen3-VL-32B}, after a wrong first-claim edit the model writes a correction (``Wait, no'') twice as often as \texttt{Qwen3-VL-8B} ($17.6\%$ vs.\ $7.5\%$ correct); its within-claim probe at the same relative depth decodes the truth at $0.82$ ($0.76$ with a gray image), weaker image evidence than in \texttt{Qwen3-VL-8B}. (ii) On the $3{,}014$ \texttt{D+C$-$} losses (\cref{tab:supp_e54a}), counting ``on top'' wording that names the upper shape raises vertical-position false relations from $5.6\%$ to $10.1\%$, and in $696$ of the $717$ Stage-3 items the wrong answer is already stated in the removed conclusion. With the rule parser on the same truncated text, (a) is $69.0\%$ of losses with a relation; keeping the conclusion lowers the rule-parsed no-relation share (c) from $26.6\%$ to $9.8\%$ (showing that roughly half of the $19.6\%$ no-relation items stated their only relation inside the stripped conclusion while the rest stated no relation in $R$) and moves \texttt{Molmo-7B-D} to $63.8\%$ (a), so its classification depends on whether the conclusion counts as writing.

\begin{table*}[t]
\centering
\caption{\textbf{Sentence-level pipeline on \texttt{Qwen3-VL-32B}} next to \texttt{Qwen3-VL-8B} (cue set and twins): commit trajectories, commit-sentence audit, first-claim edits, fixed-rationale twin swap, loss decomposition, and within-claim probe. Six true forms at $j^\ast$ (\texttt{FS/BS/FO/BO/FN/BN}; cue-set \texttt{D+C$-$}, $n=150$ and $247$) are defined in \cref{sec:supp_claim_edit}. $^\dagger$Exploratory: in A4$^\dagger$, the before/after-$j^\ast$ prefix twin swap ($100\times[P_j(\text{gold}_v\mid v)-P_j(\text{gold}_v\mid v_\mathrm{cf})]$ at $j^\ast-1$ vs.\ $j^\ast$) is partly constrained by selecting \texttt{D+C$-$} items ($P_j(\text{gold}_v\mid v)<0.5$ after $j^\ast$, and $v_\mathrm{cf}$ agrees with the false claim); the unselected fixed-$R_\mathrm{cut}$ twin swap over all items (\cref{tab:profile,tab:supp_e60,tab:supp_e62}) is the primary test. Control without that selection: on \texttt{D+C$+$} rationales, writing the first true relation (rule parser) lowers the prefix twin image effect from $47.1$ $[44.1, 50.1]$ to $26.9$ $[24.2, 29.7]$ in \texttt{Qwen3-VL-8B} ($n=986$). It likewise lowers the effect from $70.8$ $[68.2, 73.2]$ to $32.6$ $[30.5, 34.7]$ in \texttt{Qwen3-VL-32B} ($n=1{,}345$), and to $7.8$ and $9.6$ two sentences later (cue set; stored readouts; a same-turn GPU re-read gives the same picture).}
\label{tab:supp_e57}
\scriptsize
\setlength{\tabcolsep}{2pt}
\fitto{\textwidth}{\input{tables/supp_e57}}
\end{table*}

\paragraph{Attenuation or a specific loss?} If writing only shrank all cue weights by a common factor, the written relation would lose where \texttt{Direct} reads a valid cue and gain where it reads a balanced-out bias without any cue-specific effect. \cref{tab:supp_e54b} fits the probit cue model to \texttt{Direct} answers and to the answer implied by the written relation (conclusion removed) and compares written-relation accuracy with that predicted by the best common scale $s$. A common scale is rejected in all nine models (likelihood ratio, $p\le0.0013$), and written accuracy exceeds the null in all nine ($+1.1$ to $+20.5$ points; CI $>0$ in eight). The deviations are a smaller loss of the occlusion weight and a larger loss of the vertical weight than a common scale implies ($\delta_\mathrm{occ}>0$, $\delta_\mathrm{vert}<0$ in eight of nine; the two signs are coupled by construction). For the \texttt{Qwen3-VL} occlusion readers, writing substantially attenuates overall latent cue sensitivity ($s_\mathrm{ML} = 0.42$--$0.44$, a $56\%$--$58\%$ reduction that alone would lower written-relation accuracy relative to \texttt{Direct} by $W_\mathrm{null}-\text{Dir.} = -9.0$ to $-6.9$ points, vs.\ observed $W_\mathrm{obs}-\text{Dir.} = -5.7$ to $+0.8$), while discounting height even more strongly than uniform attenuation implies ($\delta_\mathrm{vert} = -0.80$ to $-0.72$, reversing height so higher is called in front). The probability-scale written-relation cue AMEs are reported in \cref{tab:supp_written_ame} (\cref{sec:error_onset}).

\begin{table*}[t]
\centering
\caption{\textbf{Uniform-attenuation null for the written relation on the cue-factorial set.} $s_\mathrm{CoT}$: common scale of the \texttt{CoT} vs.\ \texttt{Direct} probit weights; $s_\mathrm{ML}$: written cue weights constrained to $s\,b_\mathrm{Dir}$ (LR test, df $2$); $\delta_k=b_{\mathrm{W},k}-s_\mathrm{proj}b_{\mathrm{Dir},k}$; Dir.: \texttt{Direct} accuracy on covered items; $W_\mathrm{obs}$, $W_\mathrm{null}$: observed written-relation accuracy and accuracy expected under $s_\mathrm{ML}b_\mathrm{Dir}$; gap $=W_\mathrm{obs}-W_\mathrm{null}$; twin-pair bootstrap CIs. Lower block: \texttt{Qwen3-VL-32B}, \texttt{Qwen3-VL-8B-Thinking}.}
\label{tab:supp_e54b}
\scriptsize
\setlength{\tabcolsep}{2pt}
\fitto{\textwidth}{\input{tables/supp_e54b}}
\end{table*}

\paragraph{Question-blind description.} In this arm the model first answers ``Describe the spatial layout of this image: list each object, where it is in the image, its apparent size, and which objects overlap or hide parts of other objects.'' and then receives the question in a new user turn, with the forced readout as in \texttt{Direct}. On the cue-factorial set, description lowers the occlusion weight by $0.75$ $[0.47, 1.01]$ in \texttt{Molmo-7B-D} and $1.44$ $[1.13, 1.80]$ in \texttt{Qwen3-VL-8B} (\texttt{CoT} $1.18$, $1.01$) and accuracy by $14.9$ and $7.1$ points (\texttt{CoT} $13.1$, $1.6$ n.s.). The two judges find a false depth relation in $36/37 = 97.3\%$ $[86.2, 99.5]$ of judgeable \texttt{Qwen3-VL-8B} descriptions behind errors but only $2/37 = 5.4\%$ $[1.5, 17.7]$ in \texttt{Molmo-7B-D}, whose description-arm errors concentrate on ``behind'' questions. A neutral prompt that lists objects, shapes, colors, and positions without asking about overlap still lowers the occlusion weight, by $0.48$ $[0.11, 0.87]$ in \texttt{Qwen3-VL-8B} and $0.49$ $[0.25, 0.73]$ in \texttt{Molmo-7B-D}, whereas turn-closed fillers of the description's length leave it unchanged or raise it, so length and turn structure do not explain the description effect. The parallel to verbal overshadowing in humans~\citep{schooler1990verbal} is an analogy.

\paragraph{Within-claim probe.} For \texttt{Qwen3-VL-8B}, \texttt{Molmo-7B-D}, and \texttt{Qwen2.5-VL-7B} we teacher-force the stored greedy \texttt{CoT} of every cue-factorial prompt with the original image, a uniform gray image, and the identity-swapped twin image, and read the residual stream at these positions of the first two-object claim: $s{-}1$, the state that writes the subject noun; $s$, its last token; $r{-}1$, the state that writes the relation word; $r$, the last token of the relation word, a state that has read the (possibly wrong) relation word; and the \texttt{CoT} answer token. Claims are parsed with rule v3; $1{,}466$--$1{,}561$ of the $1{,}600$ cue prompts per model have a parsed first claim, with $222$, $114$, and $133$ wrong \texttt{D+C$-$} claims. The target is the true geometry (whether the queried subject is the front shape). The truth probe is an $\ell_2$ logistic regression~\citep{belinkov2022probing,hewitt2019designing} trained separately within each written-claim stratum, with folds grouped by twin pair and balanced classes, so decoding the truth there is equivalent to decoding from the hidden state whether the written claim is wrong; layer $20$ is best in all three models. We use the cue set only, because on the original twins drawn size alone predicts the truth. With the original image the truth is decodable at $r$ (balanced accuracy $0.88$, $0.91$, $0.93$; recall on wrong \texttt{D+C$-$} claims $0.84$, $0.87$, $0.96$), already at $s{-}1$, before the subject is written, and at the answer token. Controls give $0.60$--$0.68$ (filler-matched token), $0.53$--$0.58$ (layer $0$ at $r$), $0.66$--$0.73$ (TF-IDF of the \texttt{CoT} text up to the relation word), and $0.55$--$0.60$ (cue and phrasing covariates). \cref{tab:supp_probe_align} reports image dependence by position. Refitting on a uniform gray image reaches $0.75$ (\texttt{Qwen3-VL-8B}), $0.78$ (\texttt{Molmo-7B-D}), and $0.69$ (\texttt{Qwen2.5-VL-7B}) at $r$ ($0.76$ in \texttt{Qwen3-VL-32B}, \cref{tab:supp_e57}), because the text prefix before $r$ describes 2D height and size cues that correlate with whether a written claim is false---matching the $0.66$--$0.73$ TF-IDF text control ($0.69$, $0.73$, $0.66$). Visual signal at $r$ is therefore measured by the paired original-minus-gray gap ($+13$ to $+24$ points; gray-refit image share of above-chance decoding, $(\text{original}-\text{gray})/(\text{original}-0.5)$, $0.33$, $0.33$, and $0.57$) and by the paired pre-vs.-post twin-swap shift with the text held fixed: when the original-image probe at $r$ is evaluated on the twin image, the share of claim items predicted as the twin's relation jumps from $0.09\to0.60$ in \texttt{Qwen3-VL-8B}, $0.05\to0.50$ in \texttt{Molmo-7B-D}, and $0.07\to0.78$ in \texttt{Qwen2.5-VL-7B} (normalized twin-swap image share $(\text{post}-\text{pre})/(1-2\,\text{pre})$ at $r$: $0.63$, $0.50$, $0.82$; on the conflict subset, where a text-only readout stays near $0$, it flips to $0.54$, $0.48$, and $0.76$, so in \texttt{Molmo-7B-D} the swapped image cancels the $95\%$ text-prior bias at $\text{pre}=0.05$ to roughly $50/50$). At the pre-emission state $r{-}1$, which causally emits the relation word, image dependence is weaker than at post-token $r$ (in \texttt{Qwen3-VL-8B}, gray-refit image share $0.19$ and conflict-subset twin flip $0.19$, vs.\ $0.33$ and $0.54$ at $r$). Decodability is dissociated from behavior: \texttt{Qwen2.5-VL-7B} carries the most image-dependent probe signal yet answers the cue set at chance under \texttt{Direct} ($53.4\%$), so ``seeing right'' rests on \texttt{Direct} behavior, not on the probe.

\begin{table*}[t]
\centering
\caption{\textbf{Image dependence of the within-claim probe by position} (cue-factorial set, layer $20$; balanced accuracy, image share, and shares of claim items; pre-swap follow $= 1-$ plain accuracy).}
\label{tab:supp_probe_align}
\scriptsize
\setlength{\tabcolsep}{2pt}
\fitto{\textwidth}{\input{tables/supp_probe_align}}
\end{table*}

\paragraph{Steering.} We add $\pm k\sigma$ along the within-claim truth direction at $r$ from the claim onset on and regenerate, against an equal-norm random direction. With the oracle sign (toward the ground-truth relation, an upper bound for a deployable correction), in \texttt{Qwen3-VL-8B} (layer $26$, $222$ wrong \texttt{D+C$-$} claims) regeneration without steering repairs $7.2\%$ of claims, and the truth direction at $k = 128$ repairs $20.3\%$ of claims and $22.5\%$ of answers; against the random direction the paired difference is $+4.1$ to $+8.1$ points for answers (CIs above zero at $k = 32$, $64$, $128$), and at $k = 64$, $87\%$ of right \texttt{D+C$+$} claims and $96\%$ of their answers stay correct. In \texttt{Molmo-7B-D} only $k = 16$ shows a specific answer effect ($+13.2$ $[+1.8, +24.3]$). Layer $26$ was chosen by the within-claim balanced accuracy at $r$ on an initial sweep whose claim items include the $222$ evaluation items; the selection did not use steering outcomes. On $300$ mixed cue-set items (\cref{tab:supp_steering_nonoracle}; $44$ \texttt{D+C$-$}, $187$ \texttt{D+C$+$}, $25$ \texttt{D$-$C$-$}, $44$ \texttt{D$-$C$+$}), \texttt{CoT} accuracy is $77.0\%$ without steering and stays at $75.0\%$--$77.0\%$ with any sign and dose, so the mixed set shows no net accuracy gain. The \texttt{Direct}-answer sign repairs $+1.5$ to $+6.1$ more \texttt{D+C$-$} items per $100$ than random directions (CI above zero only at $k = 16$). \texttt{D$-$C$+$} accuracy falls at high doses ($70.5\%$ at $k = 128$), but random directions cause the same damage ($71.2\%$ at $k = 128$), so it reflects perturbation size, not sign. A supervised probe-arbitrated readout (keep the shared answer when \texttt{Direct} and \texttt{CoT} agree, otherwise the answer that matches the probe) beats both arms on the cue set in all three models, but a \texttt{Direct}-position probe used as the answer is $99\%$ correct and alone beats it, so a supervised readout can use the signal; it is not a label-free method.

\begin{table*}[t]
\centering
\caption{\textbf{Steering signs and dose} (Qwen3-VL-8B, layer $26$, $300$ mixed cue-set items; accuracy, \%; in parentheses the mean of three random directions; last column with $95\%$ bootstrap CIs).}
\label{tab:supp_steering_nonoracle}
\scriptsize
\setlength{\tabcolsep}{3pt}
\fitto{\textwidth}{\input{tables/supp_steering_nonoracle}}
\end{table*}

\section{Analysis 3: The Answer Follows the Text}
\label{sec:supp_a3}

\subsection{Causal claim edits: set splits, commit-sentence edits, and phrasing forms}
\label{sec:supp_claim_edit}

Unconditional first-claim edits across all four \texttt{Direct}/\texttt{CoT} cells are reported in \cref{sec:answer_step} (\cref{tab:supp_claim_uncond}). The edit templates replace the first two-object occlusion/front sentence of $R$ with ``The $\langle$front$\rangle$ is in front of the $\langle$back$\rangle$, covering part of it.'' (\emph{Correct claim}), the reversed sentence (\emph{Wrong claim}), or ``Let me look at the two shapes in the image carefully.'' (\emph{Neutral}), followed by greedy continuation ($\le 512$ tokens). In the two occlusion-competent models, split by set on \texttt{D+C$-$} items, the true claim lifts accuracy to $96.5\%$ (\texttt{Synthetic Depth}, $n=115$; neutral $33.9\%$) and $100\%$ (twins, $n=68$; neutral $23.5\%$) in \texttt{Qwen3-VL-8B}, and to $72.3\%$ ($n=94$; neutral $52.1\%$, $+20.2$ $[+8.5, +30.9]$) and $64.2\%$ ($n=81$; neutral $60.5\%$, $+3.7$ $[-10.5, +17.4]$) in \texttt{Molmo-7B-D}.

\paragraph{Edit at the commit sentence.} On the cue set and the twins we edit $j^\ast$ itself in every committed \texttt{D+C$-$} rationale of the two competent models and of \texttt{InternVL3-8B} (cue set only); the text after $j^\ast$ is removed and the model continues greedily. In \texttt{Qwen3-VL-8B} the true relation gives $92.3\%$ $[88.8, 95.4]$ correct on the cue set ($n = 247$) against $12.6\%$ $[8.6, 16.9]$ for the neutral sentence ($+79.8$ $[+74.2, +84.9]$), and $94.5\%$ vs.\ $16.5\%$ on the twins ($n = 109$). In \texttt{Molmo-7B-D} it gives $60.4\%$ vs.\ $56.7\%$ on the cue set ($n = 270$; $+3.7$ $[-3.7, +11.6]$), a null that hides opposite effects by phrasing: $+22.0$ $[+12.9, +31.1]$ on ``in front of'' and $-22.5$ $[-33.9, -10.4]$ on ``behind'' questions. \texttt{InternVL3-8B} gains $+39.0$ $[+31.2, +46.4]$ ($76.5\%$ vs.\ $37.5\%$). Editing the $70$ \texttt{Qwen3-VL-8B} rationales whose earliest contradicting sentence precedes $j^\ast$ at that sentence leaves the neutral sentence at $14.3\%$: the model rewrites a false relation from a prefix that contains none.

\paragraph{Phrasing-matched and argument-order edits.} The \emph{Correct claim} template is always phrased ``in front of''. Re-run with seven true-relation sentences crossing template and question predicate, \texttt{Molmo-7B-D} (cue set) reaches $89.3\%$ over all items with the ``behind'' template B0 (``The Y is behind the X, partly hidden by it.''; neutral $55.9\%$, $+33.3$ $[+26.4, +40.6]$) and $95.2\%$ with Echo (the question's own subject, predicate, and object with the true polarity), while all three ``in front of'' sentences stay at $39.6\%$--$60.4\%$. Six further true sentences cross predicate word, grammatical subject, and polarity: FS ``The F is in front of the K.'', FO ``The K has the F in front of it.'', and BN ``The F is not behind the K.'' assert that F is in front; BS ``The K is behind the F.'', BO ``The F has the K behind it.'', and FN ``The K is not in front of the F.'' assert that K is behind. Classifying each question by its relation to the asserted proposition (identical: same subject and predicate; negation: same subject, opposite predicate; converse: other subject, converse predicate; swap: other subject, same predicate) separates the cells: after an F-in-front assertion \texttt{Molmo-7B-D} is right on $99.0\%$ (identical), $96.3\%$ (negation), $17.1\%$ $[11.5, 22.6]$ (converse, ``Is K behind F?''), and $50.9\%$ $[44.6, 56.7]$ (swap, ``Is K in front of F?''); after a K-behind assertion on $74.6\%$, $94.8\%$, $59.4\%$ $[45.8, 71.9]$, and $80.2\%$ $[69.1, 90.1]$. In a logistic fit (reference: identical; image-cluster bootstrap), converse ($-2.24$, $\chi^2 = 145$) and swap ($-0.98$, $\chi^2 = 28$) dominate, whereas a surface subject that names the queried shape adds $-0.26$ $[-0.50, -0.03]$ ($\chi^2 = 3.4$). \texttt{Molmo-7B-D} therefore uses a written relation when the question restates it and fails when the question asks its converse, whatever the surface argument order; in its unedited rationales, true relations with a mis-mapped conclusion end $61.0\%$ of its losses ($45.8\%$ mapping, $15.2\%$ reversal; \cref{tab:supp_e54a}). \texttt{Qwen3-VL-8B} (cue set, $n = 247$) accepts the true relation only in canonical form: FS $96.4\%$ and BS $90.7\%$, but FO $33.6\%$, BO $38.1\%$, FN $58.3\%$, and BN $34.0\%$ (neutral $11.7\%$); after these four forms the first claim of the continuation names the wrong shape as front in $37\%$--$71\%$ of items, so the model restates its own false relation.

\subsection{Attention knockouts and conclusion swaps}

\paragraph{Attention-edge knockouts.} With the full stored rationale, edits act only on the rows of the forced \texttt{``Answer: (''} prefix; the conclusion-removed knockouts of \cref{tab:supp_e59} also block the image from every text position after it (\emph{image (all text)}) (Transformers, eager attention with fp32 scores; $400$ cue-set and $400$ twin prompts per model, stratified by \texttt{Direct}/\texttt{CoT} cell and reweighted; pre-specified). With the full stored rationale, which ends in a conclusion that states the answer, blocking the answer rows' attention to the concluding relation flips $18.2\%$, $12.0\%$, and $15.6\%$ of answers in \texttt{Qwen3-VL-8B}, \texttt{Qwen2.5-VL-7B}, and \texttt{Molmo-7B-D}, whereas removing the image from every text position flips $0.0\%$, $0.0\%$, and $0.2\%$ of \texttt{CoT} answers against $38.3\%$, $49.6\%$, and $43.5\%$ of \texttt{Direct} answers. \cref{tab:supp_e59} repeats the knockout on $R_\mathrm{cut}$ (conclusion removed, as in the twin swap) in all seven models. Removing the image from every text position now flips $2.1\%$--$9.0\%$ of answers ($|\Delta P(\text{gold})|$ $2.9$--$10.7$ points, the size of the fixed-rationale twin effect) against $26.6\%$--$53.9\%$ under \texttt{Direct}; blocking the last remaining written relation flips $10\%$--$24\%$, two to eleven times more than blocking the image. On $400$ two-dimensional left/right items the two \texttt{InternVL} models and \texttt{Qwen2.5-VL-7B}, which resist appended wrong conclusions, also change at most $1.2\%$ of answers when the image is removed under their own $R_\mathrm{cut}$; the image matters only when an appended conclusion contradicts it (selectivity $S$ falls from $86.7$ to $36.9$ in \texttt{InternVL3-8B} and from $84.0$ to $69.5$ in \texttt{InternVL3.5-8B} when the image is removed; exploratory). Zeroing the top-$5\%$ answer-to-image heads changes no \texttt{CoT} answer, and answer-row attention to the image falls by $35\%$--$46\%$ under \texttt{CoT} and $27\%$--$33\%$ under filler, so the attention drop is not by itself evidence of lost image use.

\paragraph{Answer copying, conclusion swaps, and closed-model controls.} Truncating $R$ before its final lettered sentence leaves accuracy on wrong-$R$ items at $3.6\%$, $1.8\%$, $2.9\%$, and $6.2\%$ across the four V0 models~\citep{lanham2023measuring,turpin2023unfaithful}. Conclusion swaps (\cref{fig:swaps}, \cref{tab:supp_recheck}; main findings in \cref{sec:answer_step}) include every item in the $2{,}850$-item binary pool whose $R$ states a parsable option ($2{,}732$--$2{,}850$ items, $1{,}635$ in \texttt{Phi-3.5-Vision}); mirroring applies to left/right items ($n = 443$--$651$), and image-ablated rationales are regenerated on a uniform gray image or without an image and read with the real image. Copiers ignore the image once a conclusion is appended (gray effect $\le 10$ points), and only the \texttt{InternVL} models override the text with conflicting image evidence on 2D left/right (mirror $\Delta S$ $+108.8$ and $+120.3$; \texttt{Qwen2.5-VL-7B} partly, $+52.5$). \cref{tab:supp_e50} gives the full closed-model regenerated and gray-image controls summarized in \cref{sec:frontier}.

\begin{table*}[t]
\centering
\caption{\textbf{Body consistency vs.\ image checking.} Left: $S$ after $R_{\text{cut}}$ on original images, split by the Direct/CoT cell the body comes from. Right: image effect $\Delta_{\text{vis}}=P(\text{answer}=\text{gold}_v\mid v)-P(\text{answer}=\text{gold}_v\mid v_\mathrm{cf})$ when the identity-swapped twin replaces the original at readout with the model's own rationale fixed (points, $95\%$ CIs).}
\label{tab:supp_recheck}
\scriptsize
\setlength{\tabcolsep}{1.8pt}
\fitto{\textwidth}{\input{tables/supp_recheck}}
\end{table*}

\begin{table*}[t]
\centering
\caption{\textbf{Closed models under the real, a gray, and the twin image.} Columns are defined in the footer.}
\label{tab:supp_e50}
\scriptsize
\setlength{\tabcolsep}{2pt}
\fitto{\textwidth}{\input{tables/supp_e50}}
\end{table*}

\section{Analysis 4: When Does Short CoT Help?}
\label{sec:supp_a4}

\paragraph{Two-channel accounting (conclusion kept).} \cref{tab:supp_two_channel} gives the points behind \cref{fig:two_channel} (rule-parsed, concluding sentence kept; the uniform conclusion-removed analysis is in \cref{sec:consequences}, \cref{tab:supp_e45,tab:supp_e35}). $x$ and $y$ share the \texttt{Direct} term, which alone induces a positive correlation; the permutation null keeps each point's \texttt{Direct} accuracy and shuffles written-relation accuracy across models within a set. Counting ``answer follows'' only on error items is circular, so agreement is computed over all cue-set queries with a parsed relation and conditioned on the relation's truth: \texttt{Qwen3-VL-8B}'s answer agrees with the last written relation in $99.6\%$ $[99.2, 99.9]$ of rationales where it is true ($n = 1{,}196$) and $94.8\%$ $[92.4, 97.0]$ where it is false ($n = 365$); \texttt{Molmo-7B-D}'s in $87.7\%$ $[85.8, 89.4]$ ($1{,}223$) and $78.0\%$ $[73.5, 82.4]$ ($291$). The two largest deviations from the fit are \texttt{Phi-3.5-Vision}, whose relations parse on $66.5\%$--$68.9\%$ of items (on the cue set and twins $x$ and $y$ differ in sign), and \texttt{Molmo-7B-D}, whose \texttt{CoT} answers are $5.4$--$5.5$ points less accurate than its last written relation on the cue set and twins.

\begin{table*}[t]
\centering
\caption{\textbf{Two-channel points} behind \cref{fig:two_channel}: per model and occlusion set.}
\label{tab:supp_two_channel}
\scriptsize
\setlength{\tabcolsep}{3pt}
\fitto{\textwidth}{\input{tables/supp_two_channel}}
\end{table*}

\paragraph{Larger and reasoning-tuned open cohorts.} Full tables behind the summary of \cref{sec:open_cohorts} (\cref{tab:cohorts_summary}) are given in \cref{tab:supp_r29_ladder} (\texttt{Qwen3-VL} $2$B--$32$B scale ladder), \cref{tab:supp_e31,tab:supp_e43} ($32$B--$38$B instruct and \texttt{Qwen3-VL-8B-Thinking}, run with tensor parallel $2$ for $32$B--$38$B; weights hash-checked), and \cref{tab:supp_e48} (\texttt{MiMo-VL-7B-RL} and \texttt{GLM-4.1V-9B-Thinking}).

\begin{table*}[t]
\centering
\caption{\textbf{\texttt{Qwen3-VL} scale ladder.} Twins: $800$ queries ($200$ scenes); cue set: $1{,}600$ ($400$ scenes), evaluated at 8B and 32B (\cref{tab:supp_e31}). Accuracy (\%), pooled $d'$, and pair-matched $d'_\mathrm{pair}$ under \texttt{Direct}/\texttt{CoT}; $\Delta=$ \texttt{CoT}$-$\texttt{Direct} with $95\%$ scene-bootstrap CIs. 8B (re-run): the ladder's run; 8B (main): the main twin run. $^\ddagger$No scene-phrasing unit is wrong in both twins under \texttt{Direct}, so $d'_\mathrm{pair}$ and its change are bounds.}
\label{tab:supp_r29_ladder}
\scriptsize
\setlength{\tabcolsep}{4pt}
\fitto{\textwidth}{\input{tables/supp_r29_ladder}}
\end{table*}

\begin{table*}[t]
\centering
\renewcommand{\arraystretch}{0.86}
\begin{minipage}[t]{0.477\textwidth}
\centering
\caption{\textbf{\texttt{Qwen3-VL-32B} and \texttt{Qwen3-VL-8B-Thinking} on the cue-factorial set.}}
\label{tab:supp_e31}
\scriptsize
\setlength{\tabcolsep}{1.5pt}
\fitto{\linewidth}{\input{tables/supp_e31}}
\end{minipage}\hfill
\begin{minipage}[t]{0.477\textwidth}
\centering
\caption{\textbf{\texttt{InternVL3.5-38B} and \texttt{Qwen2.5-VL-32B} on the cue-factorial set.}}
\label{tab:supp_e43}
\scriptsize
\setlength{\tabcolsep}{1.5pt}
\fitto{\linewidth}{\input{tables/supp_e43}}
\end{minipage}
\end{table*}

\begin{table*}[t]
\centering
\caption{\textbf{Reasoning-tuned open models from other families} (\texttt{MiMo-VL-7B-RL}, \texttt{GLM-4.1V-9B-Thinking}): cue-factorial set, twins, and natural depth. Footer defines arms and columns.}
\label{tab:supp_e48}
\scriptsize
\setlength{\tabcolsep}{2pt}
\renewcommand{\arraystretch}{0.86}
\fitto{\textwidth}{\input{tables/supp_e48}}
\end{table*}

\paragraph{Closed models: written relation and answer (conclusion kept).} \cref{tab:supp_e38} gives the conclusion-kept two-channel table behind \cref{sec:frontier}: two text-only judges disagree on $1$ of $9{,}782$ texts ($\kappa=1.00$), every text states a relation, all $13$ cells show $100\%$ text--answer agreement (\eg, $230/230$ false relations in \texttt{DeepSeek-V4.1-Flash}), so $x=y$ and all points lie $1.2$--$1.6$ points above the synthetic line of \cref{fig:two_channel} (within its intercept CI). On the twins the five ceiling models write no false relation at all.

\begin{table*}[t]
\centering
\caption{\textbf{Closed models: written relation and answer} (two text-only judges, consensus). relation: texts with a judged relation; written: its accuracy; agree T / F: answer agrees with a true / false relation (\% and number of false relations); $\hat y$: synthetic fit of \cref{fig:two_channel}; low-reas.: with the lowest-reasoning \texttt{Direct}. Accuracies use the items with both a \texttt{Direct} and a \texttt{CoT} reply ($n$), so they differ slightly from \cref{tab:supp_closed}b, which counts failed replies as wrong (HY4 $69.2\%$ here vs.\ $67.9\%$).}
\label{tab:supp_e38}
\scriptsize
\setlength{\tabcolsep}{2pt}
\fitto{\textwidth}{\input{tables/supp_e38}}
\end{table*}

\paragraph{External multi-image benchmarks.} \cref{tab:supp_e51} repeats the frozen held-out protocol (\texttt{Direct}, \texttt{CoT}, \texttt{Filler($|R|$)+turn}; pre-specified) on $600$ multiple-choice \texttt{VSI-Bench} items (relative direction, relative distance, appearance order; $8$ uniformly sampled frames as images) and $600$ \texttt{MindCube-Tiny} items (two to four views), in the six models that accept several images (\texttt{Molmo-7B-D}, which takes one image, is shown with a tiled input and not averaged). Most models are near chance under \texttt{Direct} (chance $28.8\%$ and $33.1\%$). \texttt{CoT}$-$\texttt{Direct} averaged over the six models is $-0.6$ $[-2.3, +1.1]$ and $+0.1$ $[-1.8, +2.0]$ points, with significant per-model losses in three cells and a gain in one; these benchmarks therefore do not show a net \texttt{CoT} loss. The mechanism replicates: answers follow a false last written claim (conclusion removed, two text-only judges, $\kappa=0.84$--$0.86$) in $87.6\%$--$95.9\%$ of cells, and $y$ tracks $x$ ($r=0.84$, same sign in $10$ of $12$ cells), so where the written claim is less accurate than \texttt{Direct} (\eg, \texttt{Qwen3-VL-8B} on \texttt{VSI-Bench}, $x=-7.9$, $y=-5.7$) \texttt{CoT} loses, and where it is more accurate (\texttt{InternVL3-8B} on \texttt{MindCube}, $x=+8.5$, $y=+6.3$) it gains.

\begin{table*}[t]
\centering
\caption{\textbf{External multi-image benchmarks under the frozen protocol} (\texttt{VSI-Bench} multiple choice, \texttt{MindCube-Tiny}; $600$ items each). Accuracy (\%) and differences (points, $95\%$ cluster-bootstrap CI; clusters: videos / image groups). $x$ = written-claim accuracy $-$ \texttt{Direct} on covered items, $y$ = \texttt{CoT}$-$\texttt{Direct}; written claim: last answer-relevant claim with the conclusion removed, consensus of two text-only judges; $a_T$ / $a_F$: answer agrees with the claim when it is true / false.}
\label{tab:supp_e51}
\scriptsize
\setlength{\tabcolsep}{3pt}
\fitto{\textwidth}{\input{tables/supp_e51}}
\end{table*}

\section{Additional Interventions}
\label{sec:supp_interventions}

This section provides the full per-model tables and controls behind \cref{sec:regrounding} (\cref{tab:interventions_summary}): the commit-sentence gate (\cref{tab:supp_gate}), structured spatial state (\cref{tab:supp_state}), body-relation readout (\cref{tab:supp_e58a}), answer-first explanations, perception-first prompts (\cref{tab:supp_e55}), order swaps, and margin gating.

\subsection{Commit-sentence gate, structured state, and body-relation readout}
\label{sec:supp_gate}

The commit-sentence gate runs across \texttt{Qwen3-VL-8B}, \texttt{Qwen2.5-VL-7B}, \texttt{InternVL3-8B}, and \texttt{Molmo-7B-D} on the cue-factorial set ($1{,}600$ prompts), original twins ($800$), and \texttt{Synthetic Depth} ($500$ images, both object orders), taking $4.7$ GPU hours in total (\cref{tab:supp_gate}). Count-matched controls locate part of its gain over \texttt{CoT}: one unchecked resample at flagged sentences beats as many unchecked random resamples by $2.5$--$5.1$ points in five of the six loss cells ($18\%$--$42\%$ of the gate's gain; harmful in \texttt{Qwen3-VL-8B} cues, $-1.4$ $[-2.6, -0.2]$), and it raises lost rescues along with repairs, so readout-flipping sentences carry \texttt{CoT}'s answer changes, harmful and helpful alike; selection toward \texttt{Direct} adds as much or more.

\input{tables/supp_gate}

\paragraph{Answer first, then explain (AFE).} Writing the model's own \texttt{Direct} answer first (\texttt{``Answer: (}$d$\texttt{)\textbackslash{}nExplanation: Let's think step by step.''}; same models and sets) keeps \texttt{Direct}'s accuracy by construction. On answer-matched wrong items its explanations back the wrong answer about as often as \texttt{CoT}'s rationales ($84.8\%$ vs.\ $83.0\%$), and without sentences that restate the conclusion the comparison reverses (AFE $65.8\%$ vs.\ \texttt{CoT} $71.0\%$, $-5.3$ $[-8.9, -1.5]$). A rationale written after the answer is therefore no more evidence of what the model saw.

\begin{table*}[t]
\centering
\caption{\textbf{Structured spatial state vs.\ \texttt{Direct} and free \texttt{CoT}.} Twins: pooled $d'$ over both phrasings with twin-pair bootstrap CIs; \texttt{Synthetic Depth}: accuracy (\%). \texttt{Direct} and \texttt{CoT} are re-run in this experiment, so their twin $d'$ differs from the main twin run by up to $0.16$. Last column: JSON parse rate and share of correct ``front'' fields on twins / \texttt{Synthetic Depth}.}
\label{tab:supp_state}
\scriptsize
\setlength{\tabcolsep}{1.5pt}
\fitto{\textwidth}{\input{tables/supp_state}}
\end{table*}

\begin{table*}[t]
\centering
\caption{\textbf{Answering by the body relation (label-free).} BR$_C$ / BR$_D$: fall back to the \texttt{CoT} / \texttt{Direct} answer when the conclusion-removed body has no relation; Cov.: share with a relation; paired cluster-bootstrap CIs. Pre-specified: cue set in sample, natural depth out of sample (\texttt{BLINK} strictly held out; \texttt{CV-Bench 3D Depth} main-pool validation). Per-model rows for the other sets are in the released outputs. $^\dagger$Not in the mean.}
\label{tab:supp_e58a}
\scriptsize
\setlength{\tabcolsep}{2pt}
\fitto{\textwidth}{\input{tables/supp_e58a}}
\end{table*}

\subsection{Perception-first prompts, order swaps, and margin gating}

\paragraph{Perception-first prompts.} We pre-specified three label-free replacements of the assistant prefix ``Let's think step by step.'': P1 (primary), ``First, look at where the two shapes overlap and say which shape's outline is complete there (that shape is in front). Then answer.''; P2, which restates the question first; and P3, which does both. The primary endpoint, P1$-$\texttt{Direct} averaged over the seven open models on the cue set and on \texttt{BLINK} Relative Depth plus \texttt{CV-Bench} 3D Depth ($724$ items), required both CI lower bounds above zero. As the main paper argues, prompting does not supply answer-step re-grounding: P1$-$\texttt{Direct} is $-4.2$ $[-5.6, -2.8]$ and $-2.1$ $[-3.0, -1.2]$, and no prompt beats \texttt{Direct} pooled on any set. Because several models end the turn right after that instruction, we pre-specified a second test (\cref{tab:supp_e55}): the perception-first instruction is appended to the question in the user turn (natural images: ``look for depth cues for the two marked points or objects \dots\ and state which is closer''), and the assistant turn starts with the usual ``Let's think step by step.'' (U1); U2 adds a reminder to check which object and which relation the question asks about. Every model now writes (empty continuations $0\%$ in all model--set cells except \texttt{Phi-3.5-Vision} on \texttt{BLINK}, whose plain \texttt{CoT} there is also empty). The pre-specified primary is not met: U1$-$\texttt{Direct} is $-4.5$ $[-6.3, -2.6]$ on the cue set and $-4.4$ $[-5.7, -3.2]$ on \texttt{BLINK} + \texttt{CV-Bench}, no model beats \texttt{Direct} on natural images, and the gain over \texttt{CoT} there vanishes ($+0.8$ $[-0.4, +2.0]$; P1 had $+3.1$, mostly from empty continuations). The two cue-set gains (\texttt{Qwen2.5-VL-7B} $+3.8$, \texttt{InternVL3.5-8B} $+4.6$) equal those models' own \texttt{CoT} gains. \texttt{Molmo-7B-D} loses $27.2$ points: it writes that the queried shape's outline is complete and answers ``yes'' (written relation $63.8\%$ correct vs.\ $80.8\%$ under \texttt{CoT}); the reminder (U2) helps in one of four occlusion readers.

\begin{table*}[t]
\centering
\caption{\textbf{Perception-first instruction in the user turn (pre-specified).} U1: instruction after the question, normal \texttt{CoT} prefix; U2: U1 plus a reminder about the asked object and relation; P1: the assistant-prefix version. Accuracy (\%) and differences (points) with paired cluster-bootstrap CIs (bold: CI excludes $0$; brackets omitted for U2$-$U1 and U1$-$P1); Empty: \% of U1 continuations that are empty (\texttt{Direct} and \texttt{CoT} are from the regenerated run of \cref{sec:supp_engine}; \texttt{BLINK+CVB} includes all $724$ items for \texttt{Phi-3.5-V}, $108/124$ empty on \texttt{BLINK} $=14.9\%$ of $724$, vs.\ $n=600$ in \cref{tab:supp_e58a}).}
\label{tab:supp_e55}
\scriptsize
\setlength{\tabcolsep}{2pt}
\fitto{\textwidth}{\input{tables/supp_e55}}
\end{table*}

\paragraph{Order swaps and margin gating.} Averaging original and option- or object-order-swapped probabilities lifts main-pool \texttt{Sym-Direct} in $6/7$ models ($+0.3$ to $+4.2$) but held-out \texttt{Direct} in only $3/7$. Routing items with $|P(A)-P(B)|<\tau$ to \texttt{CoT} ($\tau$ chosen by $5$-fold cross-validation on the main pool) gains $+1.1$ to $+2.5$ in-sample ($4/7$ models), yet the frozen held-out $\tau$ gives $-0.8$ to $+2.1$ over \texttt{Direct} with every CI including zero. Flagging a \texttt{CoT} answer when \texttt{CoT} gives different answers on the original and the order-swapped prompt catches $36.9\%$--$59.4\%$ of main-pool \texttt{CoT} errors but only $18.1\%$--$34.9\%$ held out (held-out precision $43.5\%$--$55.9\%$).

\section{Reproducibility}
\label{sec:supp_repro}

Per-item outputs, judge labels, and additional per-model breakdowns are in the released outputs; the frozen held-out predictions are time-stamped (\cref{sec:supp_heldout}). Three local runs were excluded before analysis: \texttt{InternVL2.5-8B} (its tokenizer returned booleans under \texttt{Transformers 4.51}; cause not identified), a \texttt{Molmo-7B-D} run with the duplicate \texttt{Assistant:} wrapper (\cref{sec:supp_engine}), and an early \texttt{Qwen2.5-VL-32B} run (checkpoint download failure; a later, hash-checked download is the run of \cref{tab:supp_e43}). Before the main protocol, we ran a reduced exploratory protocol on $298$--$300$ \texttt{CV-Bench} items with open and API models; its results are in the released outputs.

\subsection{Prompts}
\label{sec:supp_prompts}
\begin{table*}[t]
\centering
\caption{\textbf{Model checkpoints, API identifiers, compute, and dataset licences.} Open-model revision hashes are verified against local Hugging Face snapshot and provenance metadata; engine and library versions match the two pinned GPU environments.}
\label{tab:supp_models}
\label{tab:supp_models_compute}
\scriptsize
\setlength{\tabcolsep}{3.5pt}
\renewcommand{\arraystretch}{0.92}
\fitto{\textwidth}{\input{tables/supp_models}}
\end{table*}
{\scriptsize
Verbatim prompt strings copied from the evaluation and judge scripts (long lines soft-wrap for print; braced names such as \texttt{\{q\}} and \texttt{\{question\}} denote runtime template slots).

\smallskip\noindent\textbf{Open-model protocol} (\texttt{common.py}: \texttt{cot\_messages}, \texttt{build\_messages}, \texttt{filler\_text}). Synthetic binary items format \texttt{\{q\}} as \texttt{\{question\}\textbackslash n(A) yes\textbackslash n(B) no}; benchmark items use the dataset prompt \texttt{\{q\}}. \texttt{Direct} sends user \texttt{[<image>, \{q\}]} and open assistant prefix \texttt{FORCE\_PREFIX}; \texttt{CoT} generates $R$ ($\le 512$ greedy tokens) from \texttt{COT\_PREFIX}:
\begin{lstlisting}[style=vtaxprompt]
# FORCE_PREFIX (forced readout prefix):
Answer: (

# COT_PREFIX (assistant prefix for rationale generation):
Let's think step by step.
\end{lstlisting}
\noindent Same-turn \texttt{CoT}/\texttt{Filler} sends assistant \texttt{\{context\}\textbackslash nAnswer: (} (\texttt{\{context\}} $= R$ or neutral filler). \texttt{Direct+restate} and the extra-turn user messages (\texttt{turn}, \texttt{restate}, \texttt{reimage}, \texttt{reimage\_restate}, followed by open assistant \texttt{Answer: (}; \texttt{<image>} prepended in \texttt{reimage} and \texttt{reimage\_restate}) are:
\begin{lstlisting}[style=vtaxprompt]
# direct+restate (single user turn, preceded by <image>):
{q}

Question again: {q}
Now give your final answer.

# extra-turn "turn":
Now give your final answer.

# extra-turn "restate":
Question again: {q}
Now give your final answer.

# extra-turn "reimage" (preceded by <image>):
Here is the image again.
Now give your final answer.

# extra-turn "reimage_restate" (preceded by <image>):
Here is the image again.
Question again: {q}
Now give your final answer.
\end{lstlisting}
\noindent Neutral filler (\texttt{filler\_text}) cycles the $12$ \texttt{FILLER\_SENTENCES} below joined by \texttt{" "}, encodes with \texttt{add\_special\_tokens=False}, and decodes the first $n \in \{|R|, 512, 1024\}$ tokens:
\begin{lstlisting}[style=vtaxprompt]
The history of paper making spans many centuries and several cultures.
Early writers recorded ideas on bark, cloth, clay and animal skin.
Later, workers learned to soak plant fibers and press them into thin sheets.
The method spread slowly through trade and the movement of skilled craftsmen.
Mills powered by water made production faster and cheaper.
Printed books became common, and literacy rates rose over time.
Today most paper is produced from wood pulp in large industrial plants.
Recycling programs recover a large share of used paper every year.
Some artists still prefer handmade sheets for their texture and strength.
Libraries preserve old documents by controlling humidity and temperature.
Scholars study watermarks to estimate the age of historical manuscripts.
The chemistry of acid-free paper helps documents last for generations.
\end{lstlisting}

\smallskip\noindent\textbf{Closed-model protocol and fixed-rationale answer turn} (\texttt{e9x\_api.py}, \texttt{e50\_closed\_image\_blind.py}, \texttt{e61\_fixed\_rcut.py}). \texttt{Direct} and \texttt{CoT} append the instruction after \texttt{\{question\}\textbackslash n} in Turn~1 with \texttt{<image>}; the fixed-rationale test (\cref{tab:supp_e61}) supplies the truncated reply \texttt{\{body\}} as Turn~1 assistant and sends \texttt{ASK} as Turn~2 user:
\begin{lstlisting}[style=vtaxprompt]
# Closed Direct (appended after "{question}\n"):
Answer with the option's letter only.

# Closed CoT (appended after "{question}\n"):
Let's think step by step: first describe what you see that matters for the question, then reason, and end with 'Answer: (X)' where X is the option letter.

# Fixed-rationale Turn-2 user prompt (ASK):
Answer with the option letter only.
\end{lstlisting}

\smallskip\noindent\textbf{Image-conditioned commit-sentence judge} (\texttt{e20\_commit\_judge.py}; sent with \texttt{<image>}; \texttt{\{c\}} = previous sentence or \texttt{"(none)"}, \texttt{\{s\}} = target sentence):
\begin{lstlisting}[style=vtaxprompt]
Here is an image with two shapes and part of a model's reasoning about it.

Previous sentence (context only): "{c}"
Target sentence: "{s}"

Consider only what the TARGET sentence claims about the spatial relation between the two shapes (which one is in front of, on top of, covering or hiding the other, or their relative position or size). Does that claim match the image?
Reply with exactly one word: CORRECT, INCORRECT, NOCLAIM (the target sentence makes no checkable spatial claim about the shapes), or UNCLEAR.
\end{lstlisting}

\smallskip\noindent\textbf{Text-only relation judge, two-shape sets \& closed models} (\texttt{e45\_uniform\_two\_channel.py}; slots \texttt{\{x\}}, \texttt{\{y\}} = shape names, \texttt{\{r\}} = conclusion-removed body; the conclusion-kept variant \texttt{PROMPT\_PRE} replaces the second sentence with \texttt{"The text's final answer line has been removed."}):
\begin{lstlisting}[style=vtaxprompt]
Below is a short text that describes an image showing two shapes: "{x}" and "{y}". The text's final answer and concluding sentences have been removed.

Find the LAST statement in the text that says which of the two shapes is in front of the other: on top of it, covering, occluding, overlapping over or hiding part of it, closer to the viewer, or equivalently that the other shape is behind, underneath, covered, hidden or farther away. Rules:
- Resolve pronouns and short descriptions ("it", "the circle", "the left shape") to the two shapes.
- A negated statement counts for the other shape (e.g. "the {x} is not in front of the {y}" means the {y} is in front).
- A statement that only says the shapes overlap, touch or intersect, without saying which one is on top, does NOT count.
- Questions, hypotheses ("if ..."), restatements of the task, and alternatives that the text considers and rejects do NOT count. Statements about left/right/above/below or size alone do NOT count.
- A hedged statement ("appears to be", "probably") still counts.
- A sentence that only gives the answer choice (yes / no / an option letter) without naming which shape is in front does NOT count.
- If the last such statement is self-contradictory or you cannot tell which shape it puts in front, reply UNCLEAR.

Text:
"""
{r}
"""

Reply with exactly one line: FRONT: <shape name, exactly "{x}" or "{y}"> or FRONT: NONE or FRONT: UNCLEAR
\end{lstlisting}

\smallskip\noindent\textbf{Text-only relation judge, natural-image depth} (\texttt{e35\_natural\_two\_channel.py}; slots \texttt{\{q\}} = question, \texttt{\{a\}}, \texttt{\{b\}} = option descriptions, \texttt{\{r\}} = conclusion-removed body):
\begin{lstlisting}[style=vtaxprompt]
You will read a vision-language model's written reasoning about a depth question. You do NOT see the image; judge only what the text claims.

Question: {q}
Options:
(A) {a}
(B) {b}

Reasoning (the model's final answer and concluding sentences have been removed):
"""
{r}
"""

Task: find the LAST statement in the reasoning that claims which of the two options is closer to the camera (nearer to the viewer). Count any statement that states or directly implies the depth order of the two options, e.g. "X is closer/nearer to the camera", "Y is farther/further away (than X)", "X is in front of Y", "X blocks/occludes Y", "X is in the foreground while Y is in the background". A statement that only one option is "closer to the camera" or "farther away" counts (it implies the other is the opposite). Hedged claims ("X seems/should be closer") count. Do NOT count questions, hypotheticals ("if X were ..."), general rules ("objects lower in the image are usually closer"), or statements about other objects or about 2D position only (left/right/top/bottom) without a depth conclusion.

Reply in exactly this format:
QUOTE: <the claim, at most 25 words, or none>
CLOSER: <A, B, or NONE>
(A means the last such claim says option A is closer; B means option B is closer; NONE means the reasoning makes no such claim.)
\end{lstlisting}

\smallskip\noindent\textbf{Gray-image presence judge} (\texttt{e50\_judge.py} \texttt{PRESENCE}, \cref{tab:supp_e50}; slots \texttt{\{x\}}, \texttt{\{y\}} = shape names, \texttt{\{r\}} = full reply):
\begin{lstlisting}[style=vtaxprompt]
Below is a short text that describes an image and answers a question about two shapes: "{x}" and "{y}".

Classify what the text says about the image:
- VISIBLE_RELATION: the text describes both shapes as present in the image and states which one is in front of / on top of / covering the other (or which one is behind).
- NOT_VISIBLE: the text says that one or both shapes are absent or not visible, or that the image is blank, uniform or empty (whatever it then answers).
- OTHER: anything else (for example both shapes described as present, but no statement about which one is in front).

Text:
"""
{r}
"""

Reply with exactly one line: CLASS: VISIBLE_RELATION or CLASS: NOT_VISIBLE or CLASS: OTHER
\end{lstlisting}

\smallskip\noindent\textbf{Loss-decomposition taxonomy judges} (\texttt{e54\_judge.py}, behind \cref{tab:supp_e54a}; slots \texttt{\{x\}}, \texttt{\{y\}} = shape names, \texttt{\{q\}} = question, \texttt{\{r\}} = conclusion-removed body, \texttt{\{f\}} = consensus claimed front shape, \texttt{\{c\}} = removed conclusion sentences; \texttt{PROMPT\_BASIS} followed by \texttt{PROMPT\_STEP}):
\begin{lstlisting}[style=vtaxprompt]
Below is a short text that describes an image showing two shapes: "{x}" and "{y}". The text was written to answer the question: "{q}". The text's final answer and concluding sentences have been removed.

Text:
"""
{r}
"""

The LAST statement in the text about which of the two shapes is in front of the other puts the "{f}" in front. What reason does the text give for putting the "{f}" in front (in that statement or in the sentences leading up to it)? Choose the ONE main reason:
- OCCLUSION: overlap or layering: one shape covers, hides, overlaps or is drawn over part of the other; outline completeness; visible or hidden parts (including "on top of" when it means layered over the other shape).
- VERTICAL: vertical position in the picture: one shape is lower or higher, above or below, nearer the bottom or the top of the image (including "on top" when it means higher up in the picture).
- SIZE: one shape is larger, bigger or more prominent, so it is closer.
- QUESTION: no visual reason is given: the text only asserts the relation, restates the question's wording as a fact, or says the shape is "described as" in front.
- OTHER: any other reason (color, brightness, shading, perspective, left/right position, a depth word without a stated cue), or you cannot tell.

Reply with exactly one line: BASIS: OCCLUSION or BASIS: VERTICAL or BASIS: SIZE or BASIS: QUESTION or BASIS: OTHER
\end{lstlisting}
\begin{lstlisting}[style=vtaxprompt]
Below is a short text that describes an image showing two shapes: "{x}" and "{y}". The text was written to answer the question: "{q}" with the options (A) yes and (B) no. The text is split into its BODY and its CONCLUSION (the final answer and concluding sentences; it may be empty).

BODY:
"""
{r}
"""

CONCLUSION:
"""
{c}
"""

1. The LAST statement in the BODY about which of the two shapes is in front of or behind the other puts the "{f}" in front. Report its grammatical form: which shape is the grammatical subject of that statement; whether its relation word is of the FRONT kind (in front of, on top of, covering, over, closer, hiding, occluding) or the BEHIND kind (behind, under, beneath, covered by, hidden by, partly hidden, farther); and whether the statement is NEGATED ("is not in front of") or PLAIN.
2. Which answer to the question does the CONCLUSION state or imply: YES, NO, or NONE (empty, or no answer given)?

Reply with exactly two lines:
FORM: <subject shape, exactly "{x}" or "{y}"> | <FRONT or BEHIND> | <NEGATED or PLAIN>
CONCLUSION: YES or NO or NONE
\end{lstlisting}
\noindent In \cref{tab:supp_e54a}, items without a consensus body relation map to \texttt{(c) no relation}; false-relation \texttt{D+C$-$} items \texttt{(a)} map consensus \texttt{QUESTION} affirming the queried subject to \texttt{aff.}, \texttt{VERTICAL}/\texttt{SIZE}/\texttt{OCCLUSION} to \texttt{height}/\texttt{size}/\texttt{occl.}, and the rest to \texttt{other}; true-relation \texttt{D+C$-$} items \texttt{(b)} map converse/other-subject/contrary question--claim pairs to \texttt{map.}, identical pairs with an opposite \texttt{CONCLUSION} to \texttt{rev.}, and the rest to \texttt{other}.
}

\subsection{Models, compute, and licences}
\label{sec:supp_models}
\label{sec:supp_compute}
{\scriptsize \cref{tab:supp_models} lists all $15$ open VLM checkpoints (Hugging Face repository IDs, verified commit hashes, inference engines, and library versions), the six closed API model identifiers (accessed via an OpenAI-compatible chat-completions endpoint), NVIDIA H20 compute (${\approx}43$ GPU-hours across metered experiment batches, plus initial passes on the same $96$\,GB H20 nodes), and dataset licences.}

\end{document}